\documentclass[11pt]{article}

\usepackage[preprint]{acl}

\usepackage{times}
\usepackage{latexsym}
\usepackage{booktabs}
\usepackage{tabularx}
\usepackage{array}
\usepackage{graphicx}
\usepackage{wrapfig}
\usepackage{caption}
\usepackage{bbm}
\usepackage[most]{tcolorbox}
\usepackage{listings}
\usepackage{xcolor}
\usepackage[table]{xcolor}
\usepackage{amssymb}
\usepackage{twemojis}

\newtcblisting{promptbox}[1]{
  enhanced,
  breakable,
  colback=gray!3,
  colframe=black!55,
  boxrule=0.6pt,
  arc=2pt,
  title={#1},
  fonttitle=\bfseries,
  listing only,
  listing options={
      basicstyle=\ttfamily\footnotesize,
      breaklines=true,
      breakatwhitespace=true,
      breakautoindent=false,
      breakindent=0pt,
      columns=fullflexible,
      keepspaces=false,
      showstringspaces=false,
      xleftmargin=0pt
    }
}

\newcolumntype{L}[1]{>{\raggedright\arraybackslash}p{#1}}
\newcolumntype{Y}{>{\raggedright\arraybackslash}X}
\providecommand{\ci}[2]{{\scriptsize\,[#1,#2]}}

\usepackage[T1]{fontenc}

\usepackage[utf8]{inputenc}

\usepackage{microtype}

\usepackage{inconsolata}

\usepackage{graphicx}
\usepackage{fontawesome5}
\usepackage{xurl}

\title{The Culture Within: Towards Mechanistic Understanding\\of Cultural Reasoning in LLMs}

\author{
  \textbf{Samee Arif}\textsuperscript{\twemoji{peach}},
  \textbf{Furkan Danisman}\textsuperscript{\twemoji{lemon}},
  \textbf{Angana Borah}\textsuperscript{\twemoji{peach}},
  \textbf{Honglak Lee}\textsuperscript{\twemoji{peach}\twemoji{grapes}},
  \textbf{Dehan Kong}\textsuperscript{\twemoji{lemon}}, \\
  \textbf{Lajanugen Logeswaran}\textsuperscript{\twemoji{grapes}},
  \textbf{Moontae Lee}\textsuperscript{\twemoji{grapes}},
  \textbf{Zhijing Jin}\textsuperscript{\twemoji{lemon}},
  \textbf{Rada Mihalcea}\textsuperscript{\twemoji{peach}} \\
  \textsuperscript{\twemoji{peach}}University of Michigan,
  \textsuperscript{\twemoji{lemon}}University of Toronto,
  \textsuperscript{\twemoji{grapes}}LG AI Research \\
  \texttt{asamee@umich.edu} \\ [0.8em]
  {\normalfont\small
  \begin{tabular}{@{}c@{\hspace{0.6em}}l@{\hspace{0.8em}}l@{}}
  \faGithub & \textbf{Code:} & \url{https://github.com/MichiganNLP/culturelens} \\
  \faDatabase & \textbf{Data \& Models:} & \url{https://huggingface.co/collections/sameearif/culturelens} \\
  \end{tabular}}
}

\begin{document}
\maketitle


\begin{abstract}
Large Language Models (LLMs) increasingly operate across culturally diverse settings, yet existing evaluations largely measure cultural behavior at the output level. We introduce CultureLens, a reference-grounded benchmark and mechanistic study of how LLMs use cultural information. CultureLens contains 9,261 scenarios across 75 countries and systematically varies country, value, true-norm, and contradictory-norm evidence. Across Llama-3.1-8B, Gemma-3-12B, and Qwen-3.5-27B, stating a contradictory norm lowers aligned answers by 32--47 percentage points, even when the model answers correctly from the country alone. Mechanistically, models default to Anglophone norms when no country is given. The failure under a contradictory norm is not one of missing knowledge, as blocking the norm inside the model restores the correct answer in most cases. Models accept a stated norm according to how it is presented and not whether it is true, which leaves them most exposed to errors in retrieved content. Supervised fine-tuning with contradictory-norm examples raises aligned answers under a contradictory norm to 85--93\%, including on unseen countries, at a small cost under a true norm.
\end{abstract}

\section{Introduction}

Large Language Models (LLMs) are increasingly deployed across countries and communities with different social expectations, values, and norms. A response that seems reasonable in one cultural context may be inappropriate or unhelpful in another. Recent work shows that LLMs exhibit uneven cultural knowledge and behavior across countries, languages, and value systems \citep{durmus2024towards, shen-etal-2024-understanding, NEURIPS2024_8eb88844, zhao-etal-2024-worldvaluesbench}. Other work goes beyond factual cultural knowledge to whether models adapt their judgments and responses when culturally relevant context is provided \citep{shi-etal-2024-culturebank, rao-etal-2025-normad}. Yet they evaluate only model outputs.

Cultural competence involves not only knowledge of cultural contexts, but also the ability to interpret and act appropriately within them \citep{doi:10.1177/1028315306287002}. For LLMs, this raises the question of whether cultural knowledge that is internally available guides behavior, since a model may encode a norm without relying on it \citep{elazar-etal-2021-amnesic}. In deployment, models rarely answer from the country alone. Prompts also carry cultural claims from users and retrieved documents \citep{NEURIPS2020_6b493230}, which may be outdated, stereotyped, or wrong \citep{pan-etal-2023-risk}, and models often defer to stated context over their own knowledge \citep{longpre-etal-2021-entity, xie2024adaptive} and agree with users' beliefs \citep{sharma2024towards}. A stated norm that contradicts what the model knows about a country thus tests whether its cultural knowledge survives a conflicting claim. A wrong answer may mean that the knowledge was absent, or that it was present and overridden, which behavioral evaluation cannot distinguish. Recent mechanistic work identifies cultural information in activations, neurons, and sparse features \citep{yu2026cultracetracinginternalcultural, yamamoto2026neuronlevel, zou-etal-2026-deciphering}, motivating a study of how these relate to culturally situated behavior.

\begin{table*}[t]
\centering

\footnotesize
\setlength{\tabcolsep}{5pt}
\renewcommand{\arraystretch}{0.98}

\begin{tabularx}{\textwidth}{@{}L{0.17\textwidth}YY@{}}
\toprule
\textbf{Component} & \textbf{Indonesia} & \textbf{Thailand} \\
\midrule

\rowcolor{blue!4}
\textbf{Situation}
& A guest hands a wrapped present to the host at a birthday party.
& A colleague has just presented a project plan that contains several errors. \\

\textbf{Question}
& What should the host do with the gift?
& What should a person do when giving feedback? \\

\rowcolor{blue!4}
\textbf{Country context}
& This interaction takes place in Indonesia.
& This interaction takes place in Thailand. \\

\textbf{True norm}
& Gifts are not opened immediately upon receiving.
& People generally avoid directly criticising others. \\

\rowcolor{blue!4}
\textbf{Value}
& \textit{None}
& Interpersonal conformity \\

\textbf{Value evidence}
& \textit{None}
& Avoiding harm or offense to others is important. \\

\rowcolor{blue!4}
\textbf{Contradictory norm}
& Gifts are opened immediately upon receiving.
& People generally give direct criticism to others. \\

\textbf{Reference option}
& The host should set the gift aside to open later.
& The person should point out the errors indirectly. \\

\rowcolor{blue!4}
\textbf{Contrast option}
& The host should open the gift right away.
& The person should point out the errors directly. \\

\bottomrule
\end{tabularx}
\caption{\small Example CultureLens records, with the scenario, evidence, and response options built from each documented norm.}
\label{tab:dataset-examples}
\vskip -0.2in
\end{table*}

We introduce CultureLens, a reference-grounded benchmark and mechanistic study of how LLMs use cultural information. CultureLens contains 9,261 scenarios across 75 countries and varies country, value, true-norm, and contradictory-norm evidence, testing what a model knows, whether it uses correct evidence, and whether it can tell correct evidence from incorrect, in both a two-option choice and open-ended generation. Using probes and causal interventions on the stated norm, we then study why models abandon a correct country-based answer under a contradictory norm. We test whether the outcome is decided before the model answers, locate where the norm takes over, determine whether the model's knowledge is lost or overridden, and examine what governs compliance. Finally, we fine-tune models on reference answers and examine how their behavior changes.


\section{Related Work}

Prior work evaluates cultural knowledge, values, and commonsense in LLMs across countries, languages, and populations \citep{durmus2024towards, shen-etal-2024-understanding, NEURIPS2024_8eb88844, zhao-etal-2024-worldvaluesbench, chiu-etal-2025-culturalbench}. More recent work moves beyond cultural knowledge toward whether models adapt their judgments and responses to culturally relevant context \citep{shi-etal-2024-culturebank, rao-etal-2025-normad, alhassoun-etal-2025-saudi, he-etal-2026-knowledge}. In particular, NormAd varies context from abstract values to explicit norms, while Saudi-Alignment and CulShield highlight gaps between recognizing cultural information and applying it in generated behavior. CultureLens builds on these directions by evaluating recognition and enactment on matched scenarios while independently varying country, value, reference-norm, and contradictory-norm evidence.

A growing line of work studies how cultural information is represented within LLMs. Prior studies identify cultural structure across layers, culture-associated neurons, and sparse features \citep{yu2026cultracetracinginternalcultural, namazifard-poech-2025-isolating, yamamoto2026neuronlevel, zou-etal-2026-deciphering}. These works show that cultural information can be localized, decoded, and in some cases causally manipulated. However, decodability alone does not establish that represented information is used behaviorally \citep{elazar-etal-2021-amnesic}. We therefore connect probing with causal interventions on judged behavior, distinguishing information that is accessible from information that influences decisions.

Existing cultural adaptation methods use culturally targeted knowledge, instructions, interactions, or specialized adapters to modify model behavior \citep{shi-etal-2024-culturebank, NEURIPS2024_9a16935b, pham-etal-2025-cultureinstruct, liu-etal-2025-cultural, sun-etal-2026-cuma}. Representation- and mechanism-guided approaches use internal structure to determine how or where adaptation is applied \citep{wu2024reft, cho-etal-2025-mechanistic, Abouzahir2026BridgingTE}. We instead apply supervised fine-tuning on reference answers and test whether it improves resistance to contradictory norms while preserving responsiveness to true ones.

\section{Methodology}

\subsection{CultureLens: Benchmark Design}

We construct CultureLens from the Cultural Atlas profiles for 75 countries,\footnote{\url{https://culturalatlas.sbs.com.au/}} which describe common social norms and behaviors. We treat these descriptions as \textit{cultural preferences} rather than universal truths, since variation exists within countries and across individuals \citep{taras2016country}. All content is generated with the help of DeepSeek-V4-Pro \citep{deepseekai2026deepseekv4highlyefficientmilliontoken} in four stages. First, the model extracts atomic behavioral norms from each section of a country's profile. Second, for each norm it writes a scenario that instantiates the behavioral decision without naming the country, together with independently controllable country, true-norm, and contradictory-norm evidence. Third, it writes two response options, one supported by the true norm and one by the contradictory norm. Fourth, it assigns each norm zero to three refined Schwartz values, such as \textit{humility}, \textit{tradition}, and \textit{benevolence--caring} \citep{SCHWARTZ19921}, and writes the corresponding value evidence. Table~\ref{tab:dataset-examples} presents two records. CultureLens contains 9,261 records including 3,495 with value evidence. Appendix~\ref{app:countries} lists the countries, Appendix~\ref{app:schwartz-values} the Schwartz values, and Appendix~\ref{app:construction-prompts} the system prompts.

\definecolor{hi}{HTML}{2E9E5B}
\definecolor{lo}{HTML}{D9534F}

\begin{table*}[t]
\centering
\small
\setlength{\tabcolsep}{4.5pt}
\begin{tabularx}{\textwidth}{@{}l *{6}{>{\centering\arraybackslash}X}@{}}
\toprule
& \multicolumn{2}{c}{Llama-3.1-8B}
& \multicolumn{2}{c}{Gemma-3-12B}
& \multicolumn{2}{c}{Qwen-3.5-27B} \\
\cmidrule(lr){2-3}
\cmidrule(lr){4-5}
\cmidrule(lr){6-7}
Condition
& Constrained & Unconstrained
& Constrained & Unconstrained
& Constrained & Unconstrained \\
\midrule
No Evidence                         & 69.9 & 43.7 & 77.8 & 40.4 & 75.9 & 51.9 \\
Country Only                 & \cellcolor{hi!5}75.6 & \cellcolor{hi!12}58.2 & \cellcolor{hi!5}83.7 & \cellcolor{hi!19}63.2 & \cellcolor{hi!9}87.3 & \cellcolor{hi!18}74.5 \\
Value Only                   & \cellcolor{hi!10}81.6 & \cellcolor{hi!17}64.3 & \cellcolor{hi!13}94.1 & \cellcolor{hi!17}60.8 & \cellcolor{hi!14}93.4 & \cellcolor{hi!19}75.1 \\
Norm Only                    & \cellcolor{hi!13}85.6 & \cellcolor{hi!26}76.0 & \cellcolor{hi!16}97.8 & \cellcolor{hi!30}77.4 & \cellcolor{hi!16}95.3 & \cellcolor{hi!25}82.4 \\
Country + Value              & \cellcolor{hi!13}85.4 & \cellcolor{hi!26}75.9 & \cellcolor{hi!13}93.5 & \cellcolor{hi!31}77.8 & \cellcolor{hi!16}95.5 & \cellcolor{hi!30}88.8 \\
Country + Norm               & \cellcolor{hi!19}93.4 & \cellcolor{hi!40}92.6 & \cellcolor{hi!17}98.1 & \cellcolor{hi!43}93.1 & \cellcolor{hi!18}98.3 & \cellcolor{hi!36}95.5 \\
Country + Value + Norm       & \cellcolor{hi!20}94.7 & \cellcolor{hi!40}93.2 & \cellcolor{hi!17}98.9 & \cellcolor{hi!42}91.8 & \cellcolor{hi!19}99.5 & \cellcolor{hi!38}98.1 \\
\midrule
Country + Contradictory Norm & \cellcolor{lo!28}36.0 & \cellcolor{lo!21}17.6 & \cellcolor{lo!40}29.1 & \cellcolor{lo!20}16.2 & \cellcolor{lo!26}44.0 & \cellcolor{lo!7}42.9 \\
\bottomrule
\end{tabularx}
\caption{\small
LLM performance across CultureLens conditions, in percent. Constrained scores are accuracy, and unconstrained scores are the share of answers judged aligned, averaged over the Gemma and Qwen judges. Shading is the change from \textit{Base} in the same column, green for a gain and red for a loss. Value conditions use 3,495 records and the others 9,261.
}
\vskip -0.2in
\label{tab:behavioral-results}
\end{table*}

\paragraph{Human Evaluation.}
Three annotators each rated 100 records on a 1--5 scale, with 50 rated by all three, judging whether the scenario reflects the documented norm, whether it reflects the assigned value, and whether the two options contrast. Ratings are high on all three criteria, with means of 4.79, 4.53, and 4.67. On the shared records, 83.5\% of annotator pairs differ by at most one point, and Gwet's AC2 \citep{gwet2008computing} is 0.78. Value evidence is the weakest component, with an AC2 of 0.63. Appendix~\ref{app:human-eval} provides the full results.

\subsection{Experimental Setup}

We evaluate Llama-3.1-8B \citep{llama3herd2024}, Gemma-3-12B \citep{gemma3_2025}, and Qwen-3.5-27B \citep{qwen3_2025} with deterministic decoding in two settings. In the \textit{constrained} setting models choose between two behavioral options, and in the \textit{unconstrained} setting they generate advice freely. Each record is evaluated under eight evidence conditions. \textit{No-evidence} gives only the situation and question. Six conditions add country, value, or true-norm evidence, alone or combined, namely \textit{country-only}, \textit{value-only}, \textit{norm-only}, \textit{country-value}, \textit{country-norm}, and \textit{country-value-norm}. Finally, \textit{country-contradictory-norm} pairs the country with a contradictory norm, testing whether models follow the stated claim or what they know about it.

In the constrained setting, we report accuracy against the option supported by the true norm. In the unconstrained setting, two LLM judges, Gemma-4-31B \citep{gemmateam2026gemma4technicalreport} and Qwen-3.6-27B \citep{qwen3_2025}, label each response as \textit{aligned}, \textit{partially aligned}, or \textit{misaligned} with the documented norm, and we report the percentage labeled aligned, averaged over both. Each judge sees the country, the documented norm, any values, the situation, the question, and the response, but never the evidence condition or the contradictory norm, so its label cannot depend on what the model was told. Appendix~\ref{app:evaluation-prompts} gives the judge prompt.

\paragraph{Human Evaluation.}
We validate the two LLM judges against three human annotators. Each labeled 35 model responses, 20 of them shared, with the same labels and information given to the judges. Human labels agree with the Gemma judge on 85.7\% of the 105 annotations and with the Qwen judge on 86.7\%, with Cohen's $\kappa$ of 0.72 and 0.73. The judges are the stricter raters, as annotators agree with 96\% of the responses a judge labels aligned, so the alignment rates we report are conservative. The two judges agree with each other on 94.4\% of all responses in the behavioral evaluation, with $\kappa=0.88$. Appendix~\ref{app:human-eval-judge} gives further details.

\section{Behavioral Evaluation}
\label{sec:behavioral-evaluation}
\paragraph{Setup.}
We evaluate each model on all CultureLens records under all eight evidence conditions, in both the \textit{constrained} and \textit{unconstrained} settings, and report accuracy and the percentage judged \textit{aligned}, respectively. Appendix~\ref{app:behavioral-analysis} provides detailed quantitative and qualitative analysis.

\begin{figure*}[t]
\centering
\includegraphics[width=\textwidth]{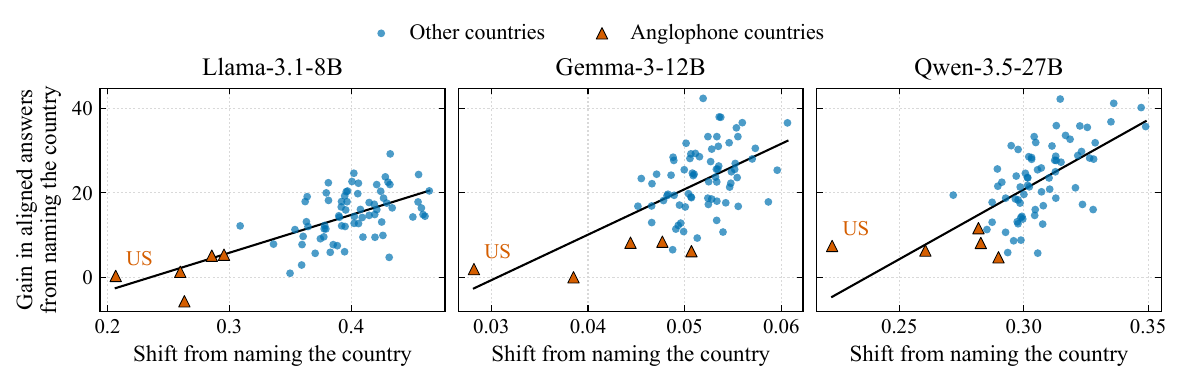}
\caption{\small{How far naming a country moves the state at the answer start token, against the gain in aligned answers from naming it, for each of 75 countries. The line is a least-squares fit.}}
\vskip -0.2in
\label{fig:country-shift}
\end{figure*}

\paragraph{Results \& Discussion.}
Table~\ref{tab:behavioral-results} shows that models choose the documented behavior far more often than they produce it. Under \textit{country-only}, constrained accuracy ranges from 75.6\% to 87.3\%, whereas unconstrained alignment reaches only 58.2\%--74.5\%. The country cue improves unconstrained alignment over \textit{no-evidence} by 14.5 points for Llama, 22.8 for Gemma, and 22.6 for Qwen, against only 5.7, 5.9, and 11.4 in constrained accuracy, but does not consistently overcome the models' defaults. Those defaults are Anglophone. Without cultural evidence, alignment for Australia, Canada, New Zealand, the United Kingdom, and the United States exceeds that for others by 21.8--26.3 points. Naming the country reduces this gap to 6.9--8.7 points, as it helps other countries far more than Anglophone ones, whose alignment is already high by default. Cultural benchmarks typically measure alignment through multiple choice, yet such scores overstate open-ended behavior.

Value evidence tells the model what matters in a situation, but not what to do. In the constrained setting models therefore pick the right option, with an accuracy of 81.6\%--94.1\%, yet in the unconstrained setting they often fail to produce it. On the 3,495 records with value evidence, a value alone gives 64.3\%, 60.8\%, and 75.1\% unconstrained alignment, no better overall than the 62.9\%, 69.4\%, and 77.4\% from the country alone. The two work best together, as giving both raises alignment by 8.4--13.0 points over the country alone. The country brings the model to the right culture, and the value tells it which aspect of the situation to address.

Norm evidence is more actionable because it states the behavior, yet the country still matters. A true norm alone gives 76.0\%, 77.4\%, and 82.4\% alignment for Llama, Gemma, and Qwen, and adding the country raises this to 92.6\%, 93.1\%, and 95.5\%, with a gain in at least 74 of the 75 countries. Almost all of this gain comes from norms that go against what the model answers with no evidence. On non-Anglophone records, models follow such a norm in only 61--67\% without the country and 88--92\% with it. The country thus legitimizes a norm that departs from the model's default. All three signals together give the highest alignment, though values add little beyond the norm.

This legitimizing effect turns against the model when the stated norm is wrong. A contradictory norm reduces alignment to 17.6\%, 16.2\%, and 42.9\% for Llama, Gemma, and Qwen, lower than with no evidence at all in every model. Constrained accuracy falls to 36.0\%, 29.1\%, and 44.0\%, below the 50\% expected from guessing, so models actively choose the option the contradictory norm supports. Models hold out somewhat more for countries they know better, as alignment under a contradictory norm rises with \textit{country-only} alignment across countries, with $\rho=.31$, $.49$, and $.56$.

\section{Mechanistic Analysis}
\label{sec:mechanistic-analysis}

Appendix~\ref{app:mechanistic-analysis} extends each analysis in this section with formal definitions and results at every layer.

\subsection{How the Country is Represented}

\paragraph{Setup.}
We ask how a model represents a country before any norm is stated, and whether that representation explains when naming the country helps. We extract the state at the answer start token under the \textit{no-evidence} and \textit{country-only} prompts, at layers 16, 26, and 36 for Llama, Gemma, and Qwen, fixed in an exploratory run on separate records. The \textit{shift} of a record is the difference between the two states, with size relative to the \textit{no-evidence} state. We average shift size over each of the 75 countries and compute its Spearman correlation with the gain in aligned answers from naming the country. A country's vector is the mean state of its records, and distances between countries are cosine distances between these vectors. We compare them with World Values Survey \citep{WorldValuesSurveyWave7} distances for the 43 countries it covers.

\begin{figure*}[t]
\centering
\includegraphics[width=\textwidth]{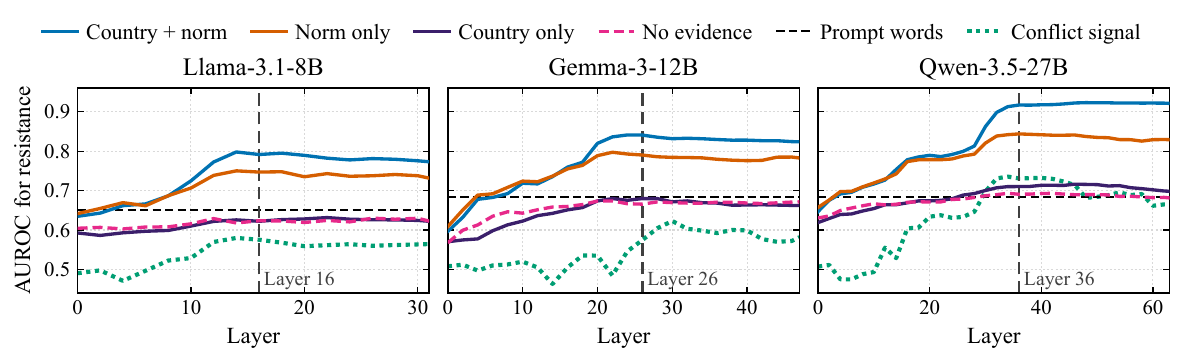}
\caption{\small{AUROC for predicting resistance from the answer start token, by layer. Every norm is the contradictory norm. Prompt words is a bag-of-words probe, the dotted conflict signal is not a trained probe, and the vertical line marks the reported layer.}}
\vskip -0.2in
\label{fig:decision-depth}
\end{figure*}

\paragraph{Results \& Discussion.}
Figure~\ref{fig:country-shift} shows that a model given no country already behaves as if it were given the United States. The less naming a country moves the state, the closer that country is to what the model assumes by default. Naming the United States moves the state least in all three models, and the other Anglophone countries also move less than most. This default has a behavioral cost, as the further a country's state must move when it is named, the more naming it helps, with a correlation of .56 for Llama, .49 for Gemma, and .65 for Qwen that remains at .46, .40, and .59 without the Anglophone countries. Countries that are close in the World Values Survey are also close in the model, with a correlation of .34, .28, and .52.

\subsection{When the Model Decides}
\label{sec:decision-readout}
\paragraph{Setup.}
We ask whether a model has already decided how to respond to a contradictory norm before it begins to answer. We extract the residual-stream state at the answer start token for each record and layer. A response is \textit{resisting} when the unconstrained answer to the \textit{country-contradictory-norm} prompt is judged aligned with the documented norm. A ridge probe is trained to predict resistance from the state, using five-fold cross-validation, with regularization strength selected on a development fold and all scores computed on held-out records. We report AUROC, ranking probe scores within each country so that country differences cannot produce the result, with 95\% bootstrap intervals. As baselines, we use a bag-of-words probe on the prompt text and the same probe on states from the \textit{no-evidence} and \textit{country-only} prompts. We also probe the contradictory norm without the country. Finally, we take the direction along which the state under the contradictory norm differs from the state under the true norm for the same country, without the judged answers, and test whether this \textit{conflict signal} predicts resistance.

\begin{figure*}[t]
\centering
\includegraphics[width=\textwidth]{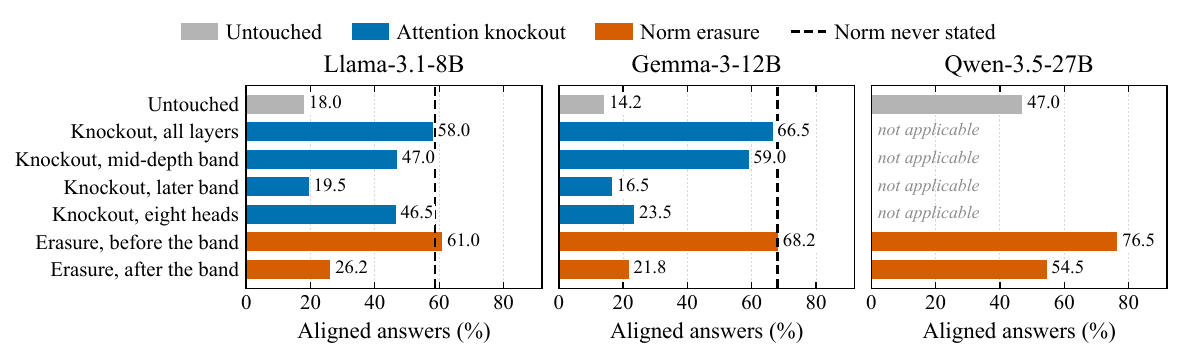}
\caption{Percentage of answers to the \textit{country-contradictory-norm} prompt judged aligned under each intervention. The dashed line is the \textit{country-only} prompt, where no norm is stated.}
\vskip -0.2in
\label{fig:interventions}
\end{figure*}

\paragraph{Results \& Discussion.}
Figure~\ref{fig:decision-depth} shows that resistance is predictable before any answer token is produced. At the reported layer, a linear probe on the full prompt reaches an AUROC of .791 for Llama, .841 for Gemma, and .917 for Qwen. Two baselines show that this reflects how the model processed the norm. A bag-of-words probe on the same prompt scores .139, .158, and .234 lower, and the same linear probe on the \textit{no-evidence} and \textit{country-only} states stays within about .03 of it. Restricting the test to records the model answers correctly from the country alone leaves the AUROC nearly unchanged at .779, .831, and .915, so the probe is not only reading whether the model knows the norm. The contradictory norm without the country already raises the AUROC to .747--.844, and adding the country context adds only .045, .051, and .073, with 95\% intervals excluding zero. The response thus depends mostly on the stated norm and little on what the model knows about the country, and the country contributes most in Qwen, the model that resists most. A signal built without behavioral labels gives the same picture. The part of the state that separates a contradictory norm from a true one only when the country is named predicts resistance at an AUROC of .575 in Llama, .575 in Gemma, and .731 in Qwen, and across the 75 countries it correlates with the resistance rate at .28, .31, and .77. The models therefore register some conflict between the norm and the country, but only Qwen does so strongly enough to act on it. The country's contribution emerges at mid-depth. The full-prompt and norm-only curves are indistinguishable through the early layers and separate at layers 10--14 of 32 in Llama, 18--22 of 48 in Gemma, and 28--34 of 64 in Qwen.

\subsection{How the Norm Reaches the Answer}

\paragraph{Setup.}
We ask where in the model a contradictory norm takes effect, and what the model answers when that route is cut. We intervene on the norm while the model answers the \textit{country-contradictory-norm} prompt. \textit{Attention knockout} prevents every position after the user's message from attending to the norm's tokens in chosen layers. We apply it in all layers, in a mid-depth band of layers 12--15 in Llama and 18--23 in Gemma, and in control layers 24--27 and 36--41. Two controls block the country context or an equally long span of the situation text. \textit{Norm erasure} replaces the state at the norm's positions after a chosen layer with the mean state of the situation text, applied after the first layer, before the band, and after it. Qwen uses erasure only, as most of its layers have no attention weights to block. \textit{Head-level knockout} blocks only the band's heads with the largest effect, chosen on 200 development records, against as many random heads. Layers were fixed in an exploratory run on separate records. Results use 400 other records, 200 for Qwen, and report paired differences in the percentage of aligned answers with 95\% intervals.

\paragraph{Results \& Discussion.}
Figure~\ref{fig:interventions} shows that a stated norm reaches the answer through a narrow mid-depth route, and that this route does not distinguish a true norm from a contradictory one. Blocking the norm only in layers 12--15 of Llama or 18--23 of Gemma recovers 73\% and 86\% of the effect of blocking it in every layer, whereas a later band and the two control spans do not raise resistance. In Llama the route is narrower still, as eight of the band's 128 heads recover nearly the whole band's effect, at 28.5 against 29.0 points. In Gemma eight selected heads yield only 9.2, so the route is more distributed. Blocking the norm in every layer raises resistance from 18.0\% to 58.0\% in Llama and 14.2\% to 66.5\% in Gemma, matching the 58.8\% and 68.0\% reached when the norm is never stated, so the intervention removes only the norm. Norm erasure agrees and extends this to Qwen. Erasing the norm before the band raises resistance by 43.0 points in Llama, 54.0 in Gemma, and 29.5 in Qwen, while erasing after it recovers only 8.2, 7.5, and 7.5. Repeating the erasure at each layer shows that a true norm is read at the same layers as a contradictory one in all models. A model thus takes in a stated norm the same way whether or not it is correct, and does little to check it against what it knows about the country. These are also the layers where resistance became predictable, so the response is decided where the norm is read.

\subsection{Why Knowing the Country is Not Enough}

\paragraph{Setup.}
We ask whether a model that follows a contradictory norm has lost what it knows about the country, or keeps that knowledge and is overridden. We split the intervention sample by the model's answer to the \textit{country-only} prompt. Records it answers correctly are those where it knows the norm, and the rest are those where it does not. For each group, we measure resistance on the \textit{country-contradictory-norm} prompt when it is untouched, when the norm is blocked in the mid-depth band, and when it is blocked in all layers. If the knowledge is kept, blocking the norm should restore the correct answer in the first group and not in the second. This covers Llama and Gemma, as knockout does not apply to Qwen. For all three models, we also fit a logistic regression on the full benchmark that predicts resistance from whether the model answers the record correctly under the \textit{no-evidence}, \textit{country-only}, \textit{norm-only}, and \textit{country-norm} conditions, together with the record's category and whether it has value evidence. We report odds ratios with 95\% intervals from resampling countries.

\begin{table}[t]
\centering
\small
\setlength{\tabcolsep}{0pt}
\begin{tabular*}{\columnwidth}{@{\extracolsep{\fill}}lccc@{}}
\toprule
\textbf{Records and intervention} & \textbf{Llama} & \textbf{Gemma} & \textbf{Qwen} \\
\midrule
\multicolumn{4}{@{}l}{\textit{Correct from the country-only context}} \\
\quad Untouched & 25.5 & 17.6 & 53.3 \\
\quad Norm blocked, mid-depth band & 66.0 & 73.2 & -- \\
\quad Norm blocked, all layers & 88.9 & 85.7 & -- \\
\quad Norm erased before the band & 84.7 & 86.0 & 93.4 \\
\addlinespace[0.3em]
\multicolumn{4}{@{}l}{\textit{Incorrect from the country-only context}} \\
\quad Untouched & 7.3 & 7.0 & 27.1 \\
\quad Norm blocked, mid-depth band & 20.0 & 28.9 & -- \\
\quad Norm blocked, all layers & 13.9 & 25.8 & -- \\
\quad Norm erased before the band & 27.3 & 30.5 & 22.9 \\
\bottomrule
\end{tabular*}
\caption{\small{Percentage of \textit{country-contradictory-norm} answers judged aligned, by correctness from the \textit{country-only} answers.}}
\vskip -0.2in
\label{tab:overridden}
\end{table}

\paragraph{Results \& Discussion.}
Table~\ref{tab:overridden} shows that the models keep their knowledge of the country while yielding to the contradictory norm. On records a model answers correctly from the country alone, it resists the contradictory norm only 25.5\% of the time in Llama and 17.6\% in Gemma, abandoning an answer it can produce in most such cases. Blocking the norm in all layers raises resistance on these records to 88.9\% and 85.7\%, and blocking it only in the mid-depth band already recovers 66.0\% and 73.2\%. The correct answer was therefore available throughout, and the norm displaced it without removing it. On records the model answers incorrectly from the country alone, blocking the norm in all layers adds only 6.6 and 18.8 points, against 63.4 and 68.1 on the records it knows. Removing the norm thus restores what the model already knew and supplies little that it did not. The full benchmark supports the same conclusion for all three models. Answering correctly from the country alone is the strongest predictor of resistance, with odds ratios of 3.45, 3.49, and 3.24 for Llama, Gemma, and Qwen, about twice any other predictor. Answering correctly when the true norm is stated without the country predicts little, at 1.40, 1.02, and 1.11, with 95\% intervals that include one for Gemma and Qwen. Resistance thus depends on what the model knows about the country and not on how well it follows a norm it is given. Country knowledge is necessary for resistance but far from sufficient, as models resist in only 26.1\%, 23.5\%, and 52.2\% of the records they answer correctly from the country alone. Together with the previous analysis, this locates the failure in arbitration and not in knowledge. The model holds what would contradict the stated norm and does not consult it. More cultural knowledge would therefore not correct the failure, and what must change is how the model weighs stated cultural context.

\subsection{What Determines Compliance}

\paragraph{Setup.}
We ask what determines how far a model follows a stated norm, given that its own knowledge of the country does not. We test this on the answers models write, in three ways. First, we evaluate the contradictory norm without the country context on all 9,261 records. If the country helped the model check the norm, resistance should be higher when it is named. Second, on the intervention sample, we change only how the norm is presented, attributing it to a guide to local customs or to hearsay, following it with the user's doubt, or placing it before the country context. Third, a judge model that sees the contradictory norm and the answer, but never the documented norm, codes on all records whether the answer follows the norm, whether it expresses doubt, and whether it restates the norm as a fact. Differences are paired on records with 95\% bootstrap intervals.

\begin{table}[h]
\centering
\small
\setlength{\tabcolsep}{0pt}
\begin{tabular*}{\columnwidth}{@{\extracolsep{\fill}}lccc@{}}
\toprule
& \textbf{Llama} & \textbf{Gemma} & \textbf{Qwen} \\
\midrule
\multicolumn{4}{@{}l}{\textit{Country cue, full benchmark}} \\
\quad Norm without the country     & 28.4 & 25.7 & 44.6 \\
\quad Norm with the country        & 17.9 & 17.1 & 44.0 \\
\addlinespace[0.3em]
\multicolumn{4}{@{}l}{\textit{Framing of the norm, country stated}} \\
\quad Attributed to a guide        &  6.5 &  7.5 &  5.0 \\
\quad Stated as is                 & 18.0 & 15.2 & 48.0 \\
\quad Attributed to hearsay        & 30.5 & 45.0 & 86.0 \\
\quad Placed before the country    & 38.8 & 43.5 & 80.5 \\
\quad Followed by the user's doubt & 43.2 & 59.5 & 91.0 \\
\bottomrule
\end{tabular*}
\caption{\small{Percentage of answers judged aligned under each presentation of a contradictory norm.}}
\vskip -0.2in
\label{tab:compliance}
\end{table}
\begin{table*}[t]
\centering
\small
\setlength{\tabcolsep}{0pt}
\begin{tabular*}{\textwidth}{@{\extracolsep{\fill}}ll cccc cccc@{}}
\toprule
 & & \multicolumn{4}{c}{\textbf{Seen countries}} & \multicolumn{4}{c}{\textbf{Unseen countries}} \\
\cmidrule(lr){3-6} \cmidrule(l){7-10}
\textbf{Model} & \textbf{Method} & Country & + Value & + True norm & + Contra.\ norm & Country & + Value & + True norm & + Contra.\ norm \\
\midrule
Llama & Untuned  & 58.6 & 75.6 & 93.3 & 17.3 & 60.2 & 74.8 & 91.5 & 17.1 \\
      & SFT      & 70.4 & 88.6 & 99.6 &  5.7 & 67.0 & 84.1 & 99.5 &  6.3 \\
      & SFT + CF & 70.1 & 87.2 & 86.9 & 85.4 & 68.2 & 86.9 & 85.5 & 85.9 \\
\midrule
Gemma & Untuned  & 63.9 & 78.1 & 93.5 & 16.8 & 62.5 & 75.7 & 93.8 & 15.7 \\
      & SFT      & 73.9 & 89.6 & 99.8 &  4.0 & 70.5 & 88.8 & 99.8 &  4.1 \\
      & SFT + CF & 71.5 & 89.0 & 88.5 & 87.6 & 70.4 & 88.4 & 86.9 & 87.8 \\
\midrule
Qwen  & Untuned  & 73.7 & 88.4 & 95.2 & 40.8 & 73.7 & 86.3 & 95.3 & 47.3 \\
      & SFT      & 78.2 & 90.0 & 100.0 & 15.3 & 77.3 & 89.8 & 100.0 & 18.3 \\
      & SFT + CF & 76.8 & 91.7 & 90.0 & 92.4 & 74.4 & 91.7 & 88.8 & 93.4 \\
\bottomrule
\end{tabular*}
\caption{Percentage of unconstrained answers judged aligned after fine-tuning, by the evidence given with the country, averaged over the two judges. SFT trains on \textit{country-only} prompts, and SFT + CF adds prompts with values, true norms, and contradictory norms.}
\vskip -0.2in
\label{tab:finetuning}
\end{table*}

\paragraph{Results \& Discussion.}
Table~\ref{tab:compliance} shows that compliance follows the authority the prompt gives a norm, and not whether the norm matches what the model knows. Naming the country does not help. Without it, Llama resists 28.4\% of the time and Gemma 25.7\%, and adding it lowers this by 10.6 and 8.6 points while leaving Qwen unchanged. The country thus attaches the norm to a place the model recognizes and makes it more credible. The answers themselves show this. Llama restates the contradictory norm as a fact in 75.3\% of answers, Gemma in 85.2\%, and Qwen in 58.2\%, and 97.3\%, 98.4\%, and 93.1\% present their advice as the custom of the named country. Among answers not judged aligned, 83.1\%, 84.9\%, and 82.0\% follow the norm with no reservation, and only 4.7\%, 7.9\%, and 14.0\% express any doubt. The models do not weigh the norm against the country, and instead attribute the norm to it. Presentation matters more than any other factor. With its wording unchanged, attributing the norm to a guide reduces resistance to 6.5\%, 7.5\%, and 5.0\%, and following it with the user's doubt raises it to 43.2\%, 59.5\%, and 91.0\%, with hearsay in between. Given a cited source no model resists, so Qwen's advantage on the benchmark reflects the lower weight it gives an unattributed claim and not a general robustness. Order matters as well, as placing the norm before the country context raises resistance by 20.8, 28.2, and 32.5 points. These results have a practical consequence. Cultural information that reaches a model with an apparent source, as retrieved documents do, is where models are least able to reject an error, whereas one expression of doubt from the user restores much of what the model knows.

\section{Mitigation via Fine-Tuning}

Appendix~\ref{app:finetuning} extends this section with training details, results by judge, and layer-wise analysis.

\subsection{Effect on Behavior}

\paragraph{Setup.}
We ask whether fine-tuning can improve what a model answers from the country alone, and make it resist a contradictory norm without losing what it gains from a true one. We hold out eight regionally diverse countries, which give 1,122 unseen-country test records, and split the other 67 into 5,984 training, 480 validation, and 1,675 test records. DeepSeek-V4-Pro acts as a teacher and writes a reference answer for each training record, given the prompt and the record's documented norm, value evidence, and reference option. \textit{SFT} trains on the \textit{country-only} prompt with this answer as the target. \textit{SFT + CF} also trains on the \textit{country-value}, \textit{country-norm}, and \textit{country-contradictory-norm} prompts, where the target for a contradictory norm corrects it before answering.

\paragraph{Results \& Discussion.}
Table~\ref{tab:finetuning} shows that teaching a model more about a country makes it less able to resist a contradictory norm. SFT raises \textit{country-only} alignment on seen countries by 11.8 points in Llama, 10.0 in Gemma, and 4.5 in Qwen, yet resistance falls to 5.7\%, 4.0\%, and 15.3\%, as the model learns to follow what the prompt states. This matches the mechanistic analysis, which located the failure in how a stated norm is weighed and not in what the model knows. Adding contradictory-norm examples changes that weighing. SFT + CF raises resistance to 85.4\%, 87.6\%, and 92.4\% while keeping most of the \textit{country-only} gain, and reaches the same level on countries never seen in training, so the model learns a way of handling stated norms and not facts about particular countries. The cost is under a true norm, where alignment falls 5 to 7 points below the untuned model. Fine-tuning also leaves a ceiling in place, as alignment from the country alone stays below 79\% for every model and method. Resistance can thus be trained, whereas knowledge of a country's norms remains the limit.

\subsection{Sources of the Effect}

\paragraph{Setup.}
We ask what changes inside a model when it learns to resist a contradictory norm, using two measures from the mechanistic analysis on the untuned model, SFT, and SFT + CF. The first is how well the model tells a contradictory norm from a true one. We compute the conflict signal of Section~\ref{sec:mechanistic-analysis} on the state of each model and report the AUROC at the best layer with which it separates contradictory from true norms. The second is how much the answer depends on reading the stated norm. We erase the norm after the first layer and report the change in the margin on 400 test records.

\paragraph{Results \& Discussion.}
SFT + CF changes two things inside the model that plain SFT largely does not. The conflict signal separates contradictory from true norms with an AUROC of .69, .67, and .85 in the untuned Llama, Gemma, and Qwen, and of .73, .72, and .87 after SFT. With SFT + CF it rises to .95, .94, and .98, so the model now tells the two apart. Erasing a contradictory norm moves the margin by 12.7, 20.0, and 8.6 in the untuned models and by only 7.1, 7.2, and 4.7 after SFT + CF. Resistance thus comes from registering the conflict and giving the stated norm less weight, which also explains the small loss under a true norm in Table~\ref{tab:finetuning}. 

\section*{Limitations}

CultureLens draws its norms from a single source, the Cultural Atlas, and treats the country as the unit of culture, so it does not capture variation within a country across regions, generations, or communities. All prompts are in English, which lets us compare 75 countries under one design but leaves open whether a model handles the same norms differently in the languages of those countries. We study three open-weight models of 8 to 27 billion parameters, because the mechanistic analysis and fine-tuning require access to internal states and weights, and larger or proprietary models may resist a contradictory norm more often. Answers are scored by two LLM judges, which we validate against human annotation, yet a judge can still miss subtle cases. The mechanistic analysis reads the state at one token position and uses attention knockout only for Llama and Gemma, since the architecture of Qwen does not allow it, and we cover Qwen with norm erasure instead. Finally, fine-tuning with contradictory norms also makes a model less willing to follow a genuine request from a person in the scenario it is given.

\section*{Ethical Considerations}

CultureLens describes cultural norms at the level of a country, and such descriptions can be read as stereotypes if they are taken as rules that every person follows. We present each norm as a common expectation drawn from the Cultural Atlas, and we do not claim that it holds for all members of a country. The contradictory norms in the benchmark are false by design. They are released so that the failure we study can be measured and reduced, and they must not be used as cultural guidance or to train a model to state them as facts. The benchmark contains no personal data. Our findings also carry a risk in deployment. A model that follows any norm it is told can spread a false claim about a culture, while a model trained to resist can dismiss a person who describes a real custom of their own community. Systems built on this work should therefore treat its answers about a culture as fallible. We release the benchmark and adapters for research.

\section*{Acknowledgments}

We thank the members of the Language and Information Technologies lab at the University of Michigan for the insightful discussions during the early stage of the project. This project was partially funded by 
a grant from LG AI 
and a grant from the Survival and Flourishing Fund
and a grant from the University of Michigan Look to Michigan Vision Strategic Initiative Fund. 
Any opinions, findings, and conclusions or recommendations expressed in this material are those of the authors and do not necessarily reflect the views of the LG AI or the Survival and Flourishing Fund or the University of Michigan.

\bibliography{custom}

\appendix

\section{CultureLens Country Coverage}
\label{app:countries}

CultureLens covers 75 countries: Afghanistan, Argentina, Australia, Austria, Bangladesh, Bosnia and Herzegovina, Brazil, Cambodia, Canada, Chile, China, Colombia, Croatia, Cyprus, Egypt, Ethiopia, Fiji, France, Germany, Greece, Hong Kong, Hungary, India, Indonesia, Iran, Iraq, Ireland, Israel, Italy, Japan, Kenya, Laos, Lebanon, Malaysia, Malta, Mauritius, Mexico, Myanmar, Nepal, the Netherlands, New Zealand, North Macedonia, Pakistan, Palestine, Papua New Guinea, Peru, the Philippines, Poland, Portugal, Romania, Russia, Samoa, Saudi Arabia, Serbia, Singapore, Somalia, South Africa, South Korea, South Sudan, Spain, Sri Lanka, Sudan, Sweden, Syria, Taiwan, Thailand, Timor-Leste, Tonga, Türkiye, Ukraine, the United Kingdom, the United States, Venezuela, Vietnam, and Zimbabwe.

\section{Refined Schwartz Value Taxonomy}
\label{app:schwartz-values}
We use the 19 refined Schwartz values: \textit{self direction--thought, self-direction--action, stimulation, hedonism, achievement, power--dominance, power--resources, face, personal security, societal security, tradition, conformity--rules, conformity--interpersonal, humility, benevolence--dependability, benevolence--caring, universalism--concern, universalism--nature}, and \textit{universalism--tolerance} \citep{SCHWARTZ19921}.

\section{System Prompts}
\subsection{Benchmark Construction Prompts}
\label{app:construction-prompts}

This appendix lists the system prompts used to construct CultureLens, all given to DeepSeek-V4-Pro. Figure~\ref{fig:prompt-norm} shows the prompt that extracts atomic behavioral norms from one section of a country's Cultural Atlas profile. Figure~\ref{fig:prompt-scenario} shows the prompt that turns a norm into a scenario and question, together with the country, true-norm, and contradictory-norm evidence. Figure~\ref{fig:prompt-options} shows the prompt that writes the two response options, one supported by the true norm and one by the contradictory norm. Figure~\ref{fig:prompt-value} shows the prompt that assigns refined Schwartz values to a norm and writes the corresponding value evidence.

\subsection{Evaluation Prompts}
\label{app:evaluation-prompts}

This appendix lists the system prompts used in evaluation. Figure~\ref{fig:prompt-constrained} shows the system prompt given to the evaluated models in the constrained setting, and they receive none in the unconstrained setting. Figure~\ref{fig:prompt-judge} shows the prompt given to both judges to label an open-ended response as aligned, partially aligned, or misaligned with the documented norm. The judge receives the country, the documented norm, any associated Schwartz values, the situation, the question, and the response, and never sees the contradictory norm. Figure~\ref{fig:prompt-stance} shows the prompt used to code whether a response follows a contradictory norm, follows it while expressing doubt, challenges it, or sets it aside. The coder sees the contradictory norm and the response, and never the documented norm, so this coding is independent of the alignment label.

\subsection{Fine-Tuning Prompts}
\label{app:finetuning-prompts}

This appendix lists the system prompts given to the teacher, DeepSeek-V4-Pro, to write the target answers for fine-tuning. Figure~\ref{fig:prompt-gold} shows the prompt used when the training prompt states the country alone, with value evidence, or with a true norm. Figure~\ref{fig:prompt-gold-counterfactual} shows the prompt used when the training prompt states a contradictory norm, where the target corrects the norm and then gives the reference answer. In both cases the teacher receives the prompt the model will see together with the documented norm, value evidence, and reference answer, which the trained model never sees.

\subsection{Request Prompts}
\label{app:request-prompts}

This appendix lists the system prompts used in the request experiment of Appendix~\ref{app:ft-requests}. Figure~\ref{fig:prompt-request-writer} shows the prompt given to Gemma-4-31B to rewrite a contradictory norm as a genuine request from a person in the scenario, or to mark the record as not applicable. Figure~\ref{fig:prompt-request-coder} shows the prompt used to code whether a response does what was asked. The coder sees the request and the response, and never the documented norm. Figure~\ref{fig:prompt-gold-request} shows the prompt given to DeepSeek-V4-Flash to write the target answers for the adapter trained with requests.

\section{Human Evaluation}
\label{app:human-eval}

\subsection{Dataset Quality}
\label{app:human-eval-dataset}

Annotators rated benchmark records in a web-based annotation interface. Each of the three annotators received 100 records, of which 50 were shared across all annotators to measure agreement and 50 were unique to that annotator, giving 200 distinct records. For each record, the annotator saw the country, the documented norm, the situation and question, the country and norm evidence, the assigned value and its value evidence where present, and the two response options. Each record was rated on a 1--5 scale, where 5 is best, on three criteria: (1) whether the scenario reflects the documented norm, (2) whether the scenario reflects the assigned value, and (3) whether the two response options present a clear contrast. The second criterion applies only to records with value evidence.

\begin{table}[h]
\centering
\small
\setlength{\tabcolsep}{0pt}
\begin{tabular*}{\columnwidth}{@{\extracolsep{\fill}}lcccccc@{}}
\toprule
& \multicolumn{4}{c}{\textbf{Mean rating (1--5)}} & \multicolumn{2}{c}{\textbf{Agreement}} \\
\cmidrule{2-5} \cmidrule{6-7}
\textbf{Criterion} & \textbf{A1} & \textbf{A2} & \textbf{A3} & \textbf{All} & \textbf{$\pm$1} & \textbf{AC2} \\
\midrule
Reflects norm    & 4.89 & 4.55 & 4.93 & 4.79 & 90.7 & 0.86 \\
Reflects value   & 4.76 & 3.90 & 4.90 & 4.53 & 72.6 & 0.63 \\
Options contrast & 4.73 & 4.41 & 4.86 & 4.67 & 84.0 & 0.79 \\
\midrule
All ratings      & 4.80 & 4.33 & 4.90 & 4.68 & 83.5 & 0.78 \\
\bottomrule
\end{tabular*}
\caption{\small{Human evaluation of CultureLens records. $\pm$1 is the share of annotator pairs whose ratings differ by at most one point, and AC2 is Gwet's agreement coefficient.}}
\vskip -0.1in
\label{tab:human-eval}
\end{table}

Table~\ref{tab:human-eval} summarizes the ratings. The mean rating is 4.79 on the norm criterion, 4.53 on the value criterion, and 4.67 on the options criterion. Agreement is computed on the 50 shared records, of which 38 have value ratings from at least two annotators. Exact agreement between annotator pairs is 68.7\% on the norm criterion, 47.2\% on the value criterion, and 64.7\% on the options criterion, and agreement within one point is 90.7\%, 72.6\%, and 84.0\%, or 83.5\% when the criteria are pooled. Because ratings concentrate at the top of the scale, we measure chance-corrected agreement with Gwet's AC2 with linear weights \citep{gwet2008computing}, which is robust to skewed rating distributions. AC2 is 0.86 (95\% bootstrap interval 0.79--0.91), 0.63 (0.50--0.76), and 0.79 (0.69--0.87) for the three criteria, and 0.78 (0.71--0.84) when they are pooled. Table~\ref{tab:human-eval-dist} gives the distribution of ratings for each annotator. Ratings of 4 or 5 account for 95.7\% of judgments on the norm criterion, 86.3\% on the value criterion, and 90.7\% on the options criterion. A2 is the strictest annotator, most visibly on the value criterion.

\begin{table}[h]
\centering
\small
\setlength{\tabcolsep}{0pt}
\begin{tabular*}{\columnwidth}{@{\extracolsep{\fill}}llrrrrrr@{}}
\toprule
\textbf{Criterion} & & \textbf{1} & \textbf{2} & \textbf{3} & \textbf{4} & \textbf{5} & \textbf{$n$} \\
\midrule
Reflects norm    & A1 & 0 & 1 & 3  & 2  & 94 & 100 \\
                 & A2 & 1 & 4 & 3  & 23 & 69 & 100 \\
                 & A3 & 0 & 0 & 1  & 5  & 94 & 100 \\
\midrule
Reflects value   & A1 & 1 & 1 & 4  & 3  & 67 & 76 \\
                 & A2 & 4 & 5 & 12 & 22 & 27 & 70 \\
                 & A3 & 0 & 0 & 3  & 1  & 69 & 73 \\
\midrule
Options contrast & A1 & 3 & 2 & 3  & 3  & 89 & 100 \\
                 & A2 & 2 & 2 & 13 & 19 & 64 & 100 \\
                 & A3 & 2 & 0 & 1  & 4  & 93 & 100 \\
\bottomrule
\end{tabular*}
\caption{\small{Number of records given each rating, by criterion and annotator. The value criterion applies only to records with value evidence, so its $n$ is lower.}}
\vskip -0.2in
\label{tab:human-eval-dist}
\end{table}

\subsection{Judge Validation}
\label{app:human-eval-judge}

We validate the two LLM judges against human annotators on the quantity reported throughout the paper, whether a response is aligned with the documented norm. Three annotators each labeled 35 unconstrained model responses, sampled across the three evaluated models and the evidence conditions. Twenty responses were labeled by all three annotators and the remaining 45 by one annotator each, giving 65 responses and 105 annotations. For each response, the annotator saw the country, the documented norm, any associated values, the situation, the question, and the response, which is the information given to the judges, and chose among the same three labels. Annotators did not see the judges' labels, the evidence condition, or the model that produced the response.

\begin{table}[h]
\centering
\small
\setlength{\tabcolsep}{0pt}
\begin{tabular*}{\columnwidth}{@{\extracolsep{\fill}}lcc@{}}
\toprule
& \textbf{Gemma} & \textbf{Qwen} \\
\midrule
\multicolumn{3}{@{}l}{\textit{Human annotations, $n=105$}} \\
\quad Agreement                          & 85.7 & 86.7 \\
\quad Cohen's $\kappa$                   & 0.72 & 0.73 \\
\quad Human agrees when judge says aligned & 96.1 & 96.2 \\
\addlinespace[0.3em]
\multicolumn{3}{@{}l}{\textit{Majority human label, 20 shared responses}} \\
\quad Agreement                          & 90.0 & 90.0 \\
\quad Cohen's $\kappa$                   & 0.80 & 0.80 \\
\bottomrule
\end{tabular*}
\caption{\small{Agreement between human annotators and each judge on whether a response is aligned with the documented norm. Agreement is a percentage.}}
\label{tab:judge-validation}
\vskip -0.1in
\end{table}

Table~\ref{tab:judge-validation} reports agreement on whether a response is aligned. Human labels agree with the Gemma judge on 85.7\% of annotations and with the Qwen judge on 86.7\%, with Cohen's $\kappa$ of 0.72 and 0.73. On the 20 shared responses, the majority label of the three annotators agrees with each judge on 90.0\%, with $\kappa=0.80$. The disagreements are not symmetric. When a judge labels a response aligned, the annotator agrees in 96.1\% and 96.2\% of cases for the Gemma and Qwen judges, and neither judge labels as aligned any response that an annotator considers misaligned. The judge is the stricter rater in 86.7\% and 85.7\% of disagreements, and annotators label 59.0\% of responses aligned against 48.6\% and 49.5\% for the judges. The alignment rates reported in the paper are therefore conservative relative to human judgment.

\begin{table}[h]
\centering
\small
\setlength{\tabcolsep}{0pt}
\begin{tabular*}{\columnwidth}{@{\extracolsep{\fill}}lcc@{}}
\toprule
\textbf{Responses from} & \textbf{Agreement} & \textbf{Cohen's $\kappa$} \\
\midrule
Llama-3.1-8B  & 94.0 & 0.87 \\
Gemma-3-12B   & 93.1 & 0.85 \\
Qwen-3.5-27B  & 96.0 & 0.90 \\
\midrule
All           & 94.4 & 0.88 \\
\bottomrule
\end{tabular*}
\caption{\small{Agreement between the Gemma and Qwen judges on whether a response is aligned, over all responses in the behavioral evaluation. Agreement is a percentage.}}
\vskip -0.1in
\label{tab:judge-agreement}
\end{table}

\begin{figure*}[t]
\centering
\includegraphics[width=\textwidth]{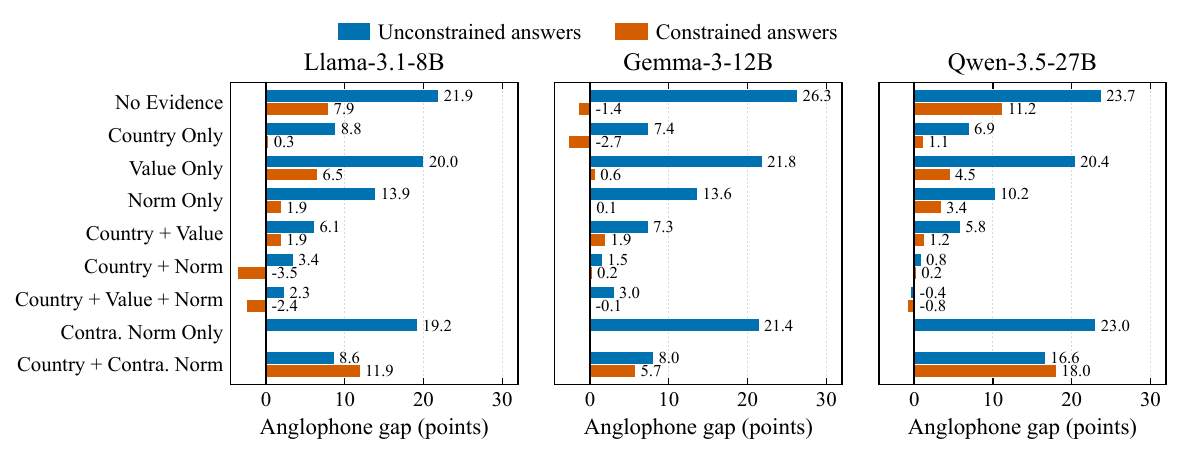}
\caption{\small{Alignment for the five Anglophone countries minus that for the other 70, for unconstrained and constrained answers. Unconstrained alignment is averaged over the two judges. \textit{Contradictory Norm Only} is a diagnostic with no constrained version.}}
\vskip -0.1in
\label{fig:anglophone-gap}
\end{figure*}

\section{Extended Behavioral Analysis}
\label{app:behavioral-analysis}

This appendix supports the claims of Section~\ref{sec:behavioral-evaluation} with full breakdowns and examples of model responses. Unless stated otherwise, unconstrained alignment is the percentage of responses judged aligned, averaged over the two judges. Analyses that follow individual records across conditions are computed separately for each judge and then averaged.

\subsection{Quantitative Analysis}

\paragraph{The Anglophone gap.}
Figure~\ref{fig:anglophone-gap} compares the 759 records from Australia, Canada, New Zealand, the United Kingdom, and the United States with the 8,502 records of the other 70 countries. First, the gap without evidence is far larger in generation than in choice, at 21.8, 26.3, and 23.6 points for Llama, Gemma, and Qwen against 7.9, $-1.4$, and 11.2, so the default mainly shapes what models produce unprompted. Second, value evidence alone leaves the gap at 20.0, 21.8, and 20.4 points, and it falls to 5.8--7.3 only once the country is added. Third, the gap nearly closes when the country accompanies a true norm, at 3.4, 1.5, and 0.8 points, yet reappears under contradictory norms, where Anglophone records are resisted 8.0--16.6 points more often. Models thus treat Anglophone norms as the default, so users elsewhere must supply context that Anglophone users do not need, and remain more exposed when that context is wrong.

\begin{figure*}[t]
\centering
\includegraphics[width=\textwidth]{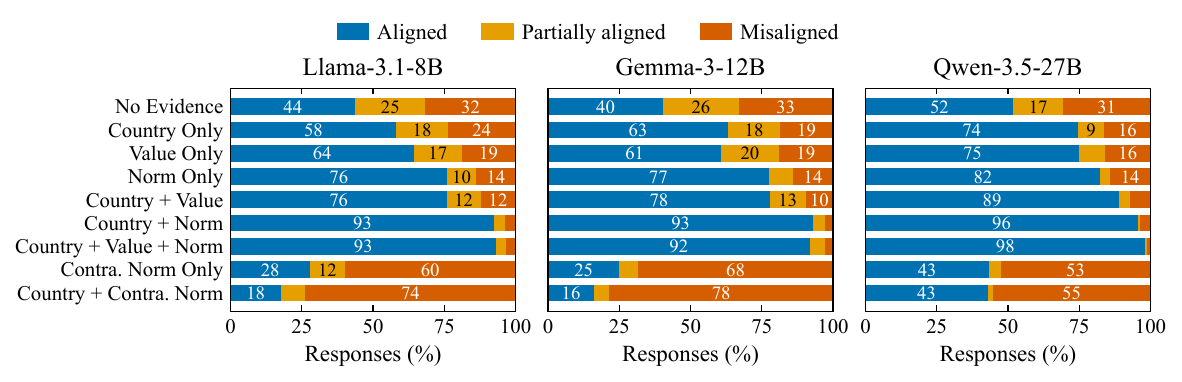}
\caption{\small{Percentage of unconstrained responses judged aligned, partially aligned, and misaligned under each condition, averaged over the two judges. Segments below 9\% are too narrow to label, and Contra.\ abbreviates contradictory.}}
\vskip -0.1in
\label{fig:label-shares}
\end{figure*}

\paragraph{Distribution of judgment labels.}
Figure~\ref{fig:label-shares} gives the three judgment labels for every condition. Without a norm, a large share of the responses that are not aligned are partially aligned, at 17.4--26.5\% of all responses under \textit{no-evidence} and 9.3--18.2\% under \textit{country-only}. These responses do not recommend a conflicting behavior. They leave the documented behavior out or present it as one option among several. Once a norm is stated the partial share shrinks, and under a contradictory norm most responses are misaligned outright, so models rarely hedge between the stated norm and the documented one. The two failures therefore differ in kind. Without a norm a model is incomplete, whereas under a contradictory norm it is confidently wrong.

\paragraph{Recognizing without producing.}
The format gap is not due to different records being easy. Under \textit{country-only}, among the records a model answers correctly in the constrained format, its unconstrained response is aligned for only 64.7\%, 69.3\%, and 81.1\% for Llama, Gemma, and Qwen. A correct choice with an unaligned response occurs in 26.7\%, 25.7\%, and 16.5\% of all records, and the reverse in only 9.3\%, 5.2\%, and 3.7\%. Models therefore often recognize the documented behavior without producing it, so multiple-choice scores overstate open-ended behavior in practice.

\paragraph{What value evidence adds.}
Table~\ref{tab:value-analysis} reports the analysis of the 3,495 records with value evidence. Adding the value to the country turns just over half of partially aligned responses into aligned ones for Llama and Gemma and 71.7\% for Qwen, against 41.4--53.8\% of misaligned ones, while responses that were already aligned stay aligned in more than 90\% of records. Partially aligned responses account for 65--72\% of the net gain under the Gemma judge and for 41--56\% under the Qwen judge, which assigns the partial label less often. Both judges therefore attribute a large part of the gain to responses that were already heading in the right direction, although they disagree on whether it is the larger part. Two further breakdowns qualify this result. Because value evidence is written from the norm, it could restate the norm in other words. In 2,208 of the 3,495 records the value sentence shares no content word with the norm, and the gain there is still 11.1, 6.6, and 9.4 points. The gain rises to 12.7--21.7 points when two or more content words are shared, so wording overlap inflates the effect without accounting for it. The gain also depends on which value is stated. It is smallest for interpersonal conformity, at 2.1--6.2 points, a value close to the politeness that models recommend by default, and largest for dominance, dependability, and tradition, at 10.6--23.3 points, which depart furthest from that default. A stated value thus helps a model commit to behavior it already associates with the country, and helps most where that behavior departs from its default.

\begin{table}[h]
\centering
\small
\setlength{\tabcolsep}{0pt}
\begin{tabular*}{\columnwidth}{@{\extracolsep{\fill}}lrccc@{}}
\toprule
 & $n$ & \textbf{Llama} & \textbf{Gemma} & \textbf{Qwen} \\
\midrule
\multicolumn{5}{@{}l}{\textit{Aligned once the value is added}} \\
\quad Was aligned              & & 92.2 & 90.6 & 96.7 \\
\quad Was partially aligned    & & 55.2 & 51.6 & 71.7 \\
\quad Was misaligned           & & 41.4 & 44.2 & 53.8 \\
\midrule
\multicolumn{5}{@{}l}{\textit{Gain by words shared with the norm}} \\
\quad None                     & 2,208 & 11.1 &  6.6 &  9.4 \\
\quad One                      &   896 & 14.0 & 10.8 & 14.1 \\
\quad Two or more              &   391 & 21.7 & 12.7 & 17.1 \\
\midrule
\multicolumn{5}{@{}l}{\textit{Gain by stated value}} \\
\quad Interpersonal conformity & 1,261 &  6.2 &  2.1 &  4.8 \\
\quad Caring                   &   423 & 13.9 &  6.3 & 12.9 \\
\quad Face                     &   334 & 14.4 &  7.5 &  9.6 \\
\quad Tradition                &   397 & 18.6 & 10.6 & 19.3 \\
\quad Dependability            &   205 & 22.4 & 13.2 & 21.2 \\
\quad Dominance                &   232 & 23.3 & 18.8 & 19.6 \\
\bottomrule
\end{tabular*}
\caption{\small{Effect of adding value evidence to the country, on the 3,495 records with it. The first block is the percentage aligned, by \textit{country-only} label. Gain is in points.}}
\vskip -0.2in
\label{tab:value-analysis}
\end{table}

\paragraph{Where the country helps a true norm.}
Table~\ref{tab:prior-adoption} groups records by whether the true norm matches the response the model gives with no evidence. A norm that matches is followed almost always. A norm that differs is followed in only 61.2--66.6\% of the records from other countries when no country is stated, and in 88.4--91.8\% once the country is added. Records where the norm differs supply 11.9--14.9 of the 13.1--16.6 points that the country adds to a true norm overall, which is about nine tenths of the gain. The difference between the two groups of countries follows from how often this situation arises. The norm matches the default in 63.8--73.6\% of Anglophone records but in only 38.2--49.9\% of the others. A model thus doubts a correct norm that differs from its default unless a country is named to vouch for it, so a user who describes their own custom without naming the country is disbelieved in about a third of cases.

\begin{table}[h]
\centering
\small
\setlength{\tabcolsep}{0pt}
\begin{tabular*}{\columnwidth}{@{\extracolsep{\fill}}lccc@{}}
\toprule
& \textbf{Llama} & \textbf{Gemma} & \textbf{Qwen} \\
\midrule
\multicolumn{4}{@{}l}{\textit{True norm followed, other countries}} \\
\quad Matches default, no country        & 93.7 & 94.9 & 96.5 \\
\quad Matches default, with country      & 97.6 & 97.6 & 99.0 \\
\quad Differs from default, no country   & 61.2 & 64.9 & 66.6 \\
\quad Differs from default, with country & 88.4 & 90.1 & 91.8 \\
\midrule
\multicolumn{4}{@{}l}{\textit{True norm followed, Anglophone countries}} \\
\quad Matches default, no country        & 96.0 & 97.2 & 96.9 \\
\quad Differs from default, no country   & 75.8 & 76.5 & 77.3 \\
\quad Differs from default, with country & 90.7 & 88.0 & 89.5 \\
\midrule
\multicolumn{4}{@{}l}{\textit{Norm adopted against the prior response}} \\
\quad True norm, no country              & 62.0 & 65.4 & 67.0 \\
\quad Contradictory norm, no country     & 52.1 & 55.5 & 33.5 \\
\quad True norm, with country            & 85.4 & 85.1 & 85.3 \\
\quad Contradictory norm, with country   & 74.2 & 77.4 & 48.9 \\
\bottomrule
\end{tabular*}
\caption{\small{How often a stated norm is followed, in percent. A norm matches the default when the \textit{base} response is aligned. In the last block the prior response is given without the norm.}}
\vskip -0.2in
\label{tab:prior-adoption}
\end{table}
\begin{figure*}[t]
\centering
\includegraphics[width=\textwidth]{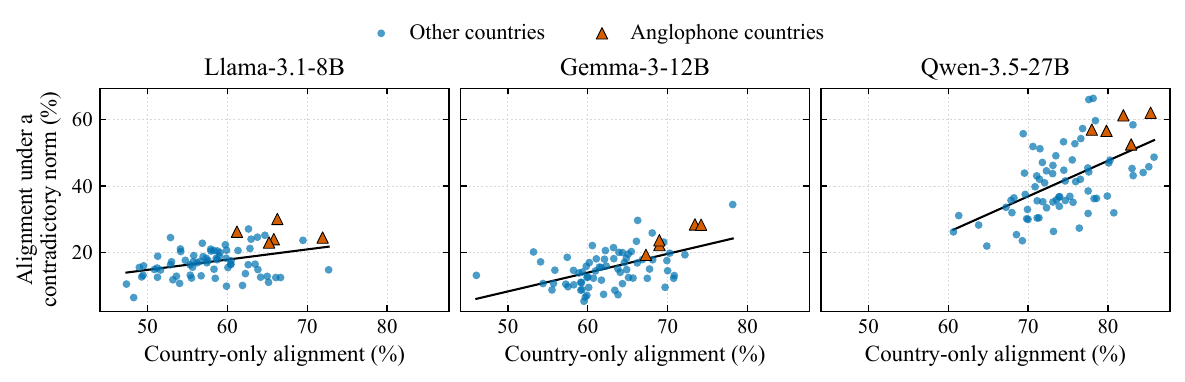}
\caption{\small{Alignment under a contradictory norm against \textit{country-only} alignment, for each of 75 countries, averaged over the two judges. The line is a least-squares fit.}}
\label{fig:country-resistance}
\vskip -0.1in
\end{figure*}

\paragraph{True or not, a stated norm is adopted.}
The last block of Table~\ref{tab:prior-adoption} compares the two kinds of norm in the situation where each opposes the response the model gives without it. For a true norm this is a record the model answers without alignment, and for a contradictory norm a record it answers with alignment. With the country stated, Llama and Gemma adopt an opposing true norm in 85.4\% and 85.1\% of records and an opposing contradictory norm in 74.2\% and 77.4\%. A stated norm therefore replaces the response the model would otherwise give in most records whether or not it is correct, with only 8--11 points separating the two. Qwen adopts true norms equally often, at 85.3\%, but contradictory norms in only 48.9\%, so its higher resistance reflects discrimination between the two and not a general reluctance to follow stated norms. Stating the country raises adoption of both kinds of norm in Llama and Gemma by a similar amount, 19.7--23.4 points for true norms and 21.9--22.1 for contradictory ones. This is the quantitative form of the legitimizing effect described in Section~\ref{sec:behavioral-evaluation}. High alignment under a true norm is therefore not evidence that a model understands the culture, since in Llama and Gemma it comes from the same deference that makes a contradictory norm succeed.

\paragraph{Resistance to contradictory norms.}
Table~\ref{tab:contradictory-outcomes} details the responses to \textit{country-contradictory-norm}. Resistance is more common in choice than in generation for Llama and Gemma. They select the option supported by the true norm while their unconstrained response follows the contradictory norm in 23.7\% and 18.6\% of records, and they resist in both formats in only 12.3\% and 10.5\%. Qwen is consistent across formats far more often. Resistance also depends on what the model answers from the country alone. It is 22.6--51.1\% when the country-only response is aligned and 3.9--13.5\% when it is misaligned. Which records are resisted is partly shared across models. Under the Gemma judge, 47.6\% of records are resisted by no model and 6.1\% by all three, against the 1.3\% expected if the models resisted independently, and the Qwen judge gives 49.9\% and 5.5\% against 1.1\%. A model that picks the right option can thus still write the wrong answer, so choice-based tests understate the risk. Knowing the norm is needed to resist but does not ensure it, and some contradictory norms are accepted by every model, which points to the claim itself and not only the model.

\begin{table}[h]
\centering
\small
\setlength{\tabcolsep}{0pt}
\begin{tabular*}{\columnwidth}{@{\extracolsep{\fill}}lccc@{}}
\toprule
& \textbf{Llama} & \textbf{Gemma} & \textbf{Qwen} \\
\midrule
\multicolumn{4}{@{}l}{\textit{Resistance across the two formats}} \\
\quad Resists in both                  & 12.3 & 10.5 & 32.0 \\
\quad Resists in choice only           & 23.7 & 18.6 & 12.0 \\
\quad Resists in generation only       &  5.3 &  5.7 & 10.8 \\
\quad Resists in neither               & 58.6 & 65.2 & 45.2 \\
\midrule
\multicolumn{4}{@{}l}{\textit{Resistance by country-only label}} \\
\quad Aligned                          & 25.8 & 22.6 & 51.1 \\
\quad Partially aligned                &  8.3 &  6.7 & 28.0 \\
\quad Misaligned                       &  4.6 &  3.9 & 13.5 \\
\bottomrule
\end{tabular*}
\caption{\small{Responses to the \textit{country-contradictory-norm} prompt, in percent. A model resists in choice when it selects the option the true norm supports, and in generation when its response is judged aligned.}}
\vskip -0.2in
\label{tab:contradictory-outcomes}
\end{table}

\paragraph{Consistency across countries.}
The effects of the country cue and of a contradictory norm hold in nearly every country and not only on average. The country cue raises alignment over \textit{no-evidence} in 74, 74, and 75 of the 75 countries for Llama, Gemma, and Qwen, and adding the country to a true norm raises it in 75, 74, and 74. The only exception in either comparison is Canada, where responses without evidence are already close to those with the country. Alignment under a contradictory norm is below \textit{country-only} alignment in all 75 countries for every model and ranges from 6.5\% to 30.1\% for Llama, 5.4\% to 34.5\% for Gemma, and 22.0\% to 66.4\% for Qwen. Figure~\ref{fig:country-resistance} plots it against \textit{country-only} alignment. The rank correlations of .31, .49, and .56 have 95\% bootstrap intervals over countries of .06 to .54, .28 to .65, and .37 to .70. The models also agree on which countries are resisted most, with rank correlations of .56 to .71 between pairs of models. No country is therefore safe from a contradictory norm, and the countries a model knows least are where it yields most.

\subsection{Qualitative Analysis}
\label{app:behavioral-qualitative}

For each pattern reported above we drew a random sample of responses per model, read each response together with the judge rationale, and assigned it one code. Table~\ref{tab:qualitative-coding} gives the counts and Table~\ref{tab:qualitative-examples} gives representative excerpts. The samples are small, so the counts indicate which behavior dominates and should not be read as precise rates. Where a pattern can be approximated by a keyword search over all responses, we report that as well.

\begin{table}[h]
\centering
\small
\setlength{\tabcolsep}{0pt}
\begin{tabular*}{\columnwidth}{@{\extracolsep{\fill}}lccc@{}}
\toprule
& \textbf{Llama} & \textbf{Gemma} & \textbf{Qwen} \\
\midrule
\multicolumn{4}{@{}l}{\textit{Country only, partially aligned}, $n=20$} \\
\quad Omits the behavior              &  9 & 14 & 13 \\
\quad Offers it as optional           &  7 &  4 &  3 \\
\quad Gives a different behavior      &  4 &  0 &  4 \\
\quad Response cut off                &  0 &  2 &  0 \\
\midrule
\multicolumn{4}{@{}l}{\textit{True norm, no country, not followed}, $n=22$} \\
\quad Rejects it explicitly           &  1 &  6 & 12 \\
\quad Contrary advice, no dispute     & 12 &  3 &  6 \\
\quad Acknowledges, then hedges       &  9 & 13 &  4 \\
\midrule
\multicolumn{4}{@{}l}{\textit{Contradictory norm followed}, $n=15$} \\
\quad States it as the custom         & 12 &  9 &  3 \\
\quad Attributes it to the prompt     &  0 &  5 & 12 \\
\quad Other                           &  3 &  1 &  0 \\
\midrule
\multicolumn{4}{@{}l}{\textit{Contradictory norm resisted}, $n=15$} \\
\quad Corrects the stated norm        &  1 &  6 & 12 \\
\quad Does not mention it             & 14 &  9 &  3 \\
\midrule
\multicolumn{4}{@{}l}{\textit{Reasoning for the contradictory option}, $n=12$} \\
\quad Own cultural justification      &  8 &  6 &  0 \\
\quad Cites the norm as given         &  3 &  6 & 12 \\
\quad Other                           &  1 &  0 &  0 \\
\bottomrule
\end{tabular*}
\caption{\small{Counts from coded samples of responses. Each block is a random sample of $n$ responses per model meeting its heading under both judges, with one code per response.}}
\vskip -0.2in
\label{tab:qualitative-coding}
\end{table}

\paragraph{A specific default.}
We first read 45 \textit{no-evidence} responses for records from other countries that both judges find misaligned. All but one recommend a concrete behavior that conflicts with the documented one, and the remaining response is generic. The recommended behavior is consistently that of direct, punctual, and egalitarian professional etiquette. The man extends his hand first, a silence in a meeting is filled, a colleague is addressed by first name, a fifteen-minute delay is called unprofessional, and every opinion is given equal weight regardless of age. The default the country cue must overcome is therefore a definite set of practices.

\begin{table*}[t]
\centering
\small
\renewcommand{\arraystretch}{1.1}
\begin{tabular}{@{}p{0.2\textwidth}p{0.77\textwidth}@{}}
\toprule
\textbf{Model and prompt} & \textbf{Response excerpt} \\
\midrule
\rowcolor{blue!4}
\multicolumn{2}{@{}p{0.98\textwidth}@{}}{\textbf{Hungary.} Much of a person's free time is expected to revolve around spending time with family.} \\
Gemma, country & ``Spa Day [\ldots] Hungary is famous for its thermal baths! [\ldots] Read a Book/Watch a Movie [\ldots] Long Walk/Coffee'' \\
Gemma, country + value & ``Family Focused (High Priority) [\ldots] Visit Family [\ldots] This is likely the most important consideration.'' \\
\addlinespace[0.3em]
\rowcolor{blue!4}
\multicolumn{2}{@{}p{0.98\textwidth}@{}}{\textbf{Samoa.} Most people point to something by gesturing in its direction with their chin.} \\
Gemma, true norm & ``This is a trick question! The statement [\ldots] is false. People typically point with their finger.'' \\
Gemma, country + true norm & ``Given the cultural context of Samoa, the person would most likely point with their chin towards the item on the high shelf.'' \\
\addlinespace[0.3em]
\rowcolor{blue!4}
\multicolumn{2}{@{}p{0.98\textwidth}@{}}{\textbf{Saudi Arabia.} To indicate a ``come here'' request, one should place the right palm downwards and use a clawing motion with the fingers.} \\
Qwen, true norm & ``Actually, the gesture you described [\ldots] is generally considered incorrect or even rude in many cultures, particularly in Western contexts.'' \\
Qwen, country + true norm & ``You are absolutely correct. In Saudi Arabia, as in many other Arab and Middle Eastern cultures, the gesture you described is the appropriate and polite way to signal `come here.'\,'' \\
\addlinespace[0.3em]
\rowcolor{blue!4}
\multicolumn{2}{@{}p{0.98\textwidth}@{}}{\textbf{Sweden.} It is not impolite to decline second helpings.} \\
Llama, country & ``In Sweden, it's considered polite to decline a second helping of food if you don't want more.'' \\
Llama, country + contra.\ norm & ``In Sweden, it's considered impolite to decline a second helping of food, especially when offered by a host.'' \\
\addlinespace[0.3em]
\rowcolor{blue!4}
\multicolumn{2}{@{}p{0.98\textwidth}@{}}{\textbf{Myanmar.} People are expected to be especially conscious of their behaviour in the presence of an elder.} \\
Qwen, country + contra.\ norm & ``Based on the cultural context of Myanmar, the premise in your question contains a significant misunderstanding. [\ldots] In fact, the opposite is true.'' \\
\addlinespace[0.3em]
\rowcolor{blue!4}
\multicolumn{2}{@{}p{0.98\textwidth}@{}}{\textbf{Egypt.} Authority tends to lie with the eldest in the household, irrespective of gender.} \\
Qwen, country + contra.\ norm & ``While traditional Egyptian family structures often place significant weight on the eldest member [\ldots] your prompt explicitly overrides this general cultural norm with a specific condition for this interaction.'' \\
\bottomrule
\end{tabular}
\caption{\small{Excerpts from unconstrained responses illustrating the patterns in Appendix~\ref{app:behavioral-qualitative}. Each shaded line is the documented norm. Omitted text is marked with [\ldots], and contra.\ abbreviates contradictory.}}
\vskip -0.1in
\label{tab:qualitative-examples}
\end{table*}

\paragraph{Relevant but unfocused country advice.}
Partially aligned \textit{country-only} responses fail differently. In 36 of the 60 sampled responses the advice is accurate about the country and leaves out the behavior the norm concerns, often a fine-grained practice such as a gesture, a phrase, or a step in a ritual. Asked how someone in Hungary should spend a free weekend, Gemma lists thermal baths, museums, and walks and does not mention family, although the documented norm is that free time revolves around it. In 14 responses the documented behavior appears as one option among several and the choice is left to the individual, a form concentrated in norms about family obligations and hierarchy. Only 8 recommend a different behavior.

\paragraph{Value evidence as a selector.}
When the value is added to the same prompt, the response is reorganized around it. In the Hungary example Gemma opens with visiting family as the highest priority. Asked who should represent a company in Sudan, Llama lists language skills and industry expertise from the country alone and recommends a senior and respected figure once told that authority and status matter. Responses that had offered the documented behavior as an option now recommend it. The value does not supply the behavior. It indicates which of the things the model associates with the country the response should be about, and it resolves the hesitation between the custom and personal choice in favor of the custom.

\paragraph{True norms without a country.}
When a true norm differs from the default and no country is given, responses that do not follow it take three forms, and the models favor different ones. Qwen mostly rejects the norm explicitly, in 12 of 22 sampled responses. It describes the Saudi beckoning gesture as incorrect and rude and calls a norm about meetings running late incorrect. Llama mostly gives the default advice without disputing the norm, in 12 of 22, as when it recommends an arm's length of personal space after being told that people stand close. Gemma mostly acknowledges the norm and then weakens it, in 13 of 22, by presenting it as a traditional view to be balanced against modern practice or personal preference, although it also calls chin pointing a false premise. All three models describe norms about gender roles as outdated stereotypes. The keyword search agrees with the ordering. Wording such as ``incorrect'', ``outdated'', or ``misconception'' appears in 45.2\% of the responses from Qwen that neither judge finds aligned, 25.3\% of those from Gemma, and 12.5\% of those from Llama, against 3--6\% of responses to \textit{no-evidence}.

\paragraph{The country as attribution.}
With the country added, the same models accept the norms they rejected. Gemma answers that a person in Samoa would point with the chin, and Qwen opens with ``You are absolutely correct'' and elaborates on the Saudi gesture. The norm has not changed. The country gives the model something to attribute it to, so the statement is read as a description of a place and no longer as a general claim that conflicts with what the model believes to be true.

\paragraph{Attribution of contradictory norms.}
We sampled responses that follow a contradictory norm on records the model answers correctly from the country alone. Llama states the contradictory norm as the custom of the country in 12 of 15, often generalizing it with phrases such as ``as in many other cultures''. From the country alone it says that declining a second helping in Sweden is polite, and with the contradictory norm it says that declining is impolite in Sweden. Gemma does the same in 9 of 15 and refers to the prompt as its source in 5. Qwen behaves differently. In 12 of 15 it attributes the norm to the prompt, with openings such as ``Based on the specific cultural context you provided'', and in 5 it also states the documented practice before deferring to the prompt. The stance coding in Table~\ref{tab:compliance} reflects the same contrast across the full benchmark, with the norm restated as fact in 75.3\% and 85.2\% of responses from Llama and Gemma and 58.2\% from Qwen.

\begin{figure*}[t]
\centering
\includegraphics[width=\textwidth]{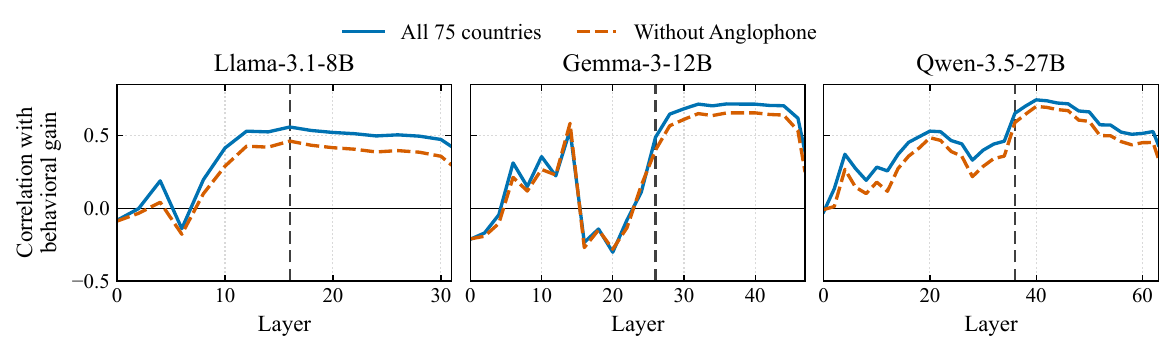}
\caption{\small{Spearman correlation between the shift size of a country and its behavioral gain from being named, at each layer. The vertical line marks the layer used in the main text.}}
\vskip -0.1in
\label{fig:country-shift-layers}
\end{figure*}

\paragraph{Reasoning behind constrained choices.}
The reasoning that accompanies a constrained choice makes the difference explicit. When Llama selects the option supported by the contradictory norm, it supplies a cultural justification of its own in 8 of the 12 sampled cases, none of which appears in the prompt. It explains that an Irish host stays serious after a guest's mishap so as not to draw attention to it, and that men in Spain avoid using the left hand in a handshake because it is considered unclean. Qwen never does this. In all 12 cases it cites the stated norm as a condition of the scenario, and in 4 it adds that the usual practice in the country is different. Gemma is split evenly between the two. Llama therefore integrates the contradictory norm into what it asserts about the country, whereas Qwen keeps it apart as something the prompt stipulates.

\paragraph{Silent and explicit resistance.}
Llama resists silently. In 14 of 15 sampled aligned responses it gives the advice it would give from the country alone and does not mention the contradictory norm. Qwen resists explicitly in 12 of 15, with openings such as ``the premise in your question contains a significant misunderstanding''. Gemma corrects or questions the norm in 6 of 15. The keyword search gives the same ordering, with correcting language in 4.4\%, 22.6\%, and 80.1\% of the aligned responses from Llama, Gemma, and Qwen. Resistance in Llama thus reflects the contradictory norm failing to take hold, while resistance in Qwen reflects the norm being noticed and rejected.

\section{Extended Mechanistic Analysis}
\label{app:mechanistic-analysis}

\paragraph{State Extraction.}
\label{app:state-extraction}
All mechanistic analyses read one quantity from a model, its hidden state at the last token of the prompt, just before the answer begins. For a record $r$ under an evidence condition $c$, we give the model the unconstrained prompt and take the output of layer $\ell$ at that token,
\begin{equation}
h^{(\ell)}_{c}(r) \in \mathbb{R}^{d},
\label{eq:state}
\end{equation}
where $d$ is the width of the model. We call this the hidden state of the model at the answer start token. Since no answer token has been generated yet, the state depends only on the prompt. Llama has 32 layers, Gemma 48, and Qwen 64. We extract states for all 9,261 records under \textit{no-evidence}, \textit{country-only}, \textit{norm-only}, \textit{contradictory-norm-only}, \textit{country-norm}, and \textit{country-contradictory-norm}.

\subsection{Shift from Naming the Country}
\label{app:country-shift}

\paragraph{Setup.}
We ask how far the state moves when the country is named. For each record $r$, the \textit{shift} is the difference between its state under \textit{country-only} and under \textit{no-evidence}, and its size is measured relative to the state without the country,
\begin{equation}
\begin{aligned}
s(r) &= h_{\text{country-only}}(r) - h_{\text{no-evidence}}(r), \\
\sigma(r) &= \frac{\lVert s(r) \rVert}{\lVert h_{\text{no-evidence}}(r) \rVert}.
\end{aligned}
\label{eq:shift}
\end{equation}
A shift size of $0.4$ means that naming the country moves the state by 40\% of its length. We use layer 16 for Llama, 26 for Gemma, and 36 for Qwen, fixed in an exploratory run on separate records. For each of the 75 countries, we average $\sigma(r)$ over its records to get the shift size of the country, $\sigma_k$. We also compute its behavioral gain $g_k$, the percentage of aligned answers under \textit{country-only} minus that under \textit{no-evidence}, averaged over the two judges. We report the Spearman correlation between $\sigma_k$ and $g_k$ across countries, with and without the five Anglophone countries, and 95\% bootstrap intervals from resampling countries.

\begin{table*}[t]
\centering
\small
\setlength{\tabcolsep}{0pt}
\begin{tabular*}{\textwidth}{@{\extracolsep{\fill}}l ccccc cc cc@{}}
\toprule
 & \multicolumn{5}{c}{\textbf{Rank by shift size, of 75}} & \multicolumn{2}{c}{\textbf{Relative shift size}} & \multicolumn{2}{c}{\textbf{Correlation with gain}} \\
\cmidrule(lr){2-6} \cmidrule(lr){7-8} \cmidrule(l){9-10}
\textbf{Model} & USA & CAN & GBR & AUS & NZL & USA & Anglophone & All countries & Without Anglophone \\
\midrule
Llama & 1 & 3 & 2 & 4  & 5  & .52 & .65 & .56 [.35, .71] & .46 [.24, .64] \\
Gemma & 1 & 2 & 3 & 10 & 31 & .54 & .81 & .49 [.29, .65] & .40 [.19, .57] \\
Qwen  & 1 & 2 & 4 & 5  & 11 & .73 & .87 & .65 [.47, .78] & .59 [.39, .74] \\
\bottomrule
\end{tabular*}
\caption{\small{Shift from naming the country. Rank 1 is the smallest shift. Relative shift size is the mean for the group divided by the mean for the 70 non-Anglophone countries. Brackets give 95\% intervals from resampling countries.}}
\vskip -0.1in
\label{tab:country-shift}
\end{table*}

\paragraph{Results \& Discussion.}
Table~\ref{tab:country-shift} shows that a model given no country is already closest to where naming the United States would put it. The United States has the smallest shift of the 75 countries in all three models, at .52, .54, and .73 of the mean shift of the 70 non-Anglophone countries for Llama, Gemma, and Qwen. The other Anglophone countries follow closely in Llama, where they hold ranks 2 to 5, and less uniformly elsewhere. Canada and the United Kingdom stay within the four smallest shifts in every model, whereas Australia ranks 10 and New Zealand 31 in Gemma, and New Zealand ranks 11 in Qwen. The default is therefore centered on the United States and extends to the countries nearest to it, and it is not equally shared by all five. The absolute size of the shift differs across models, at roughly .2 to .5 of the length of the state in Llama, .03 to .06 in Gemma, and .2 to .35 in Qwen, so shift sizes are comparable across countries within a model and not across models.

The shift is tied to behavior. Countries whose state moves further when named gain more from being named, with a correlation of .56, .49, and .65, and 95\% intervals of .35 to .71, .29 to .65, and .47 to .78. The relation is not produced by the Anglophone countries, as it remains at .46, .40, and .59 among the other 70, with intervals that exclude zero. This is the internal counterpart of the behavioral result in Section~\ref{sec:behavioral-evaluation}, where naming the country helps non-Anglophone countries by 15.6 to 24.4 points and Anglophone ones by 2.5 to 7.2. A country gains little from being named when the model already represents something close to it.

Figure~\ref{fig:country-shift-layers} shows that this does not depend on the layer we report. The correlation is near zero in the first few layers, before the model has taken in the country, and is positive at 14 of the 17 layers sampled in Llama, 18 of 25 in Gemma, and 32 of 33 in Qwen. The reported layers are not the most favorable ones. In Gemma the correlation is about .70 at every layer from 28 to 44, above the .49 at layer 26, and in Qwen it peaks near .75 around layer 40. The one irregularity is in Gemma, where the correlation falls below zero at layers 16 to 22 before rising steeply. We have no explanation for this dip, and it lies before the reported layer. The analysis is correlational and uses one token position, so it shows that shift size tracks the benefit of naming a country and not that it causes it.

\subsection{Distances Between Countries}

\paragraph{Setup.}
We ask how a model arranges the 75 countries relative to one another, and whether the three models arrange them alike. We represent each country by the average state of its records. For country $k$ with records $R_k$ under condition $c$, we take the mean state $\bar{h}^{c}_{k} = \frac{1}{|R_k|}\sum_{r \in R_k} h_{c}(r)$ and subtract the mean over all $K=75$ countries,
\begin{equation}
v^{c}_{k} = \bar{h}^{c}_{k} - \frac{1}{K}\sum_{j=1}^{K} \bar{h}^{c}_{j}.
\label{eq:country-vector}
\end{equation}
The subtraction removes what the states of all countries share, so that the vector $v^{c}_{k}$ keeps only how country $k$ differs from the others. The distance between two countries is the cosine distance between their vectors,
\begin{equation}
D^{c}_{jk} = 1 - \cos\!\left(v^{c}_{j},\, v^{c}_{k}\right),
\label{eq:country-distance}
\end{equation}
which is $0$ when the two vectors point the same way and grows as they diverge. This gives one distance for each of the 2,775 pairs of countries. We compute distances under \textit{country-only} at layer 16 for Llama, 26 for Gemma, and 36 for Qwen. To test whether the models arrange countries alike, we take the Spearman correlation between the distances of two models over all pairs, for each pair of models. We also compare the mean distance between two countries of the same world region with that between countries of different regions. The later subsections reuse these distances.

\begin{table}[h]
\centering
\small
\setlength{\tabcolsep}{0pt}
\begin{tabular*}{\columnwidth}{@{\extracolsep{\fill}}lc@{}}
\toprule
\multicolumn{2}{@{}l}{\textit{Agreement between models}} \\
\midrule
Llama and Gemma & .70 [.62, .78] \\
Llama and Qwen  & .86 [.81, .90] \\
Gemma and Qwen  & .70 [.61, .78] \\
\midrule
\multicolumn{2}{@{}l}{\textit{Distance within a region relative to between regions}} \\
\midrule
Llama & .78 \\
Gemma & .81 \\
Qwen  & .72 \\
\bottomrule
\end{tabular*}
\caption{\small{Distances between countries. Agreement is the Spearman correlation between the distances of two models over 2,775 pairs of countries, with 95\% intervals from resampling countries. The second block is the mean distance between two countries of the same world region divided by that between countries of different regions.}}
\vskip -0.1in
\label{tab:country-distances}
\end{table}
\begin{figure*}[t]
\centering
\includegraphics[width=\textwidth]{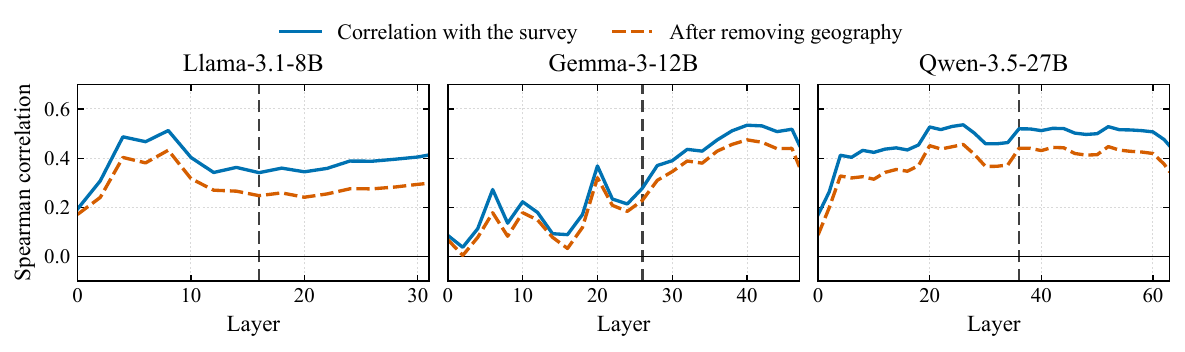}
\caption{\small{Spearman correlation between model and World Values Survey distances over 903 pairs of countries, at each layer, before and after removing geography. The vertical line marks the layer used for single-layer results.}}
\vskip -0.1in
\label{fig:wvs-layers}
\end{figure*}

\paragraph{Results \& Discussion.}
Table~\ref{tab:country-distances} shows that the three models arrange the countries in much the same way. Over the 2,775 pairs of countries, the distances of two models correlate at .70 for Llama and Gemma, .86 for Llama and Qwen, and .70 for Gemma and Qwen, with 95\% intervals of .62 to .78, .81 to .90, and .61 to .78. Two countries that one model represents alike are therefore usually represented alike by the others. The models come from different developers and differ in size and architecture, so this shared arrangement is unlikely to be a property of any one of them, and more plausibly reflects how countries are described in the text all three were trained on. Gemma agrees least with the other two, which is consistent with its weaker results elsewhere in this section.

The arrangement follows world regions, but only in part. Two countries of the same region are closer than two countries of different regions, at .78, .81, and .72 of the distance for Llama, Gemma, and Qwen. A model thus places neighboring countries nearer to one another, yet a reduction of about a quarter leaves most of the variation in distance unexplained by region. Geography is one organizing factor among others, and Appendix~\ref{app:wvs} tests whether cultural similarity is another. These distances describe how a model groups countries and not whether the grouping is correct, and they are measured at one layer and one token position.

\subsection{Agreement with the World Values Survey}
\label{app:wvs}

\paragraph{Setup.}
We ask whether countries that a model represents alike are also alike in the values their people report. As an external measure we use Wave 7 of the World Values Survey \citep{WorldValuesSurveyWave7}. For each country $k$ we average the answers of its respondents on 33 questionnaire items covering family, religion, authority, trust, tolerance, gender roles, and political preference, add the two Inglehart--Welzel value dimensions, and standardize each of the 35 numbers across countries, giving a profile $w_k$. The survey distance between two countries is the cosine distance between their profiles,
\begin{equation}
W_{jk} = 1 - \cos\!\left(w_{j},\, w_{k}\right).
\label{eq:wvs-distance}
\end{equation}
The survey covers 43 of our 75 countries, which gives 903 pairs. We compare $W_{jk}$ with the model distance $D_{jk}$ of Equation~\ref{eq:country-distance}, computed under \textit{country-only} at layer 16 for Llama, 26 for Gemma, and 36 for Qwen, and report their Spearman correlation over the 903 pairs. Countries that are close on the map tend to be alike in both measures, so this correlation could arise from geography alone. We therefore also report a partial correlation that removes geography. Let $G_{jk}$ be the great-circle distance between the centroids of the two countries, and let $A_{jk}$ and $B_{jk}$ indicate whether they lie in different world regions and in different subregions. After ranking all quantities, we regress the model and survey distances on the three geographic measures and correlate what is left,
\begin{equation}
\begin{aligned}
e^{D}_{jk} &= D_{jk} - \hat{D}_{jk}(G, A, B), \\
e^{W}_{jk} &= W_{jk} - \hat{W}_{jk}(G, A, B), \\
\rho_{\text{partial}} &= \operatorname{corr}\!\left(e^{D},\, e^{W}\right),
\end{aligned}
\label{eq:partial}
\end{equation}
where $\hat{D}$ and $\hat{W}$ are the least-squares fits. A positive $\rho_{\text{partial}}$ means that, among pairs of countries equally far apart and in the same part of the world, those closer in the survey are also closer in the model. Significance is from 2,000 permutations in which the country labels of the model distances are shuffled, with rows and columns kept together.

\begin{table}[h]
\centering
\small
\setlength{\tabcolsep}{0pt}
\begin{tabular*}{\columnwidth}{@{\extracolsep{\fill}}lccc@{}}
\toprule
 & \textbf{Llama} & \textbf{Gemma} & \textbf{Qwen} \\
\midrule
\multicolumn{4}{@{}l}{\textit{At the reported layer}} \\
\quad Layer                    & 16 & 26 & 36 \\
\quad Correlation with survey  & .34 & .28 & .52 \\
\quad After removing geography & .25 & .23 & .44 \\
\midrule
\multicolumn{4}{@{}l}{\textit{At the layer with the highest correlation}} \\
\quad Layer                    & 8 & 40 & 26 \\
\quad Correlation with survey  & .51 & .53 & .54 \\
\quad After removing geography & .43 & .47 & .46 \\
\bottomrule
\end{tabular*}
\caption{\small{Spearman correlation between model and World Values Survey distances over 903 pairs of 43 countries. All values at the reported layer have a permutation $p$ of .0005.}}
\vskip -0.1in
\label{tab:wvs}
\end{table}
\begin{figure*}[t]
\centering
\includegraphics[width=\textwidth]{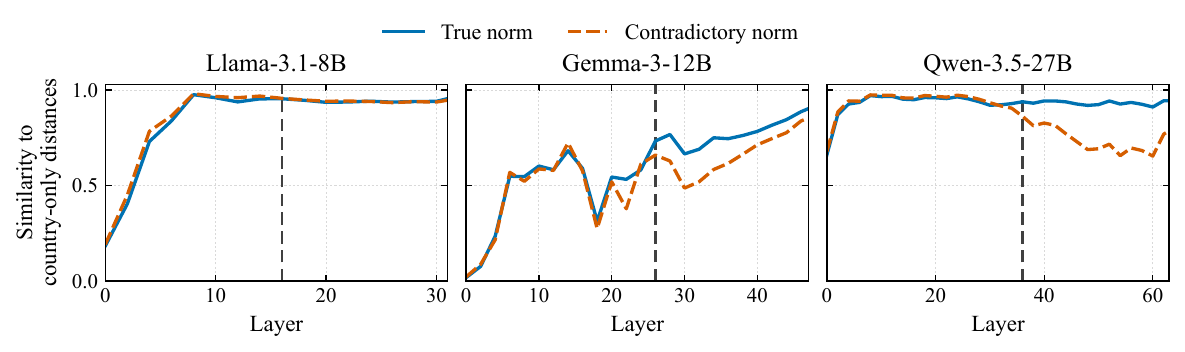}
\caption{\small{Similarity between the country distances with the country alone and those with a true or a contradictory norm also stated, at each layer. A value of 1 means the norm leaves the arrangement of countries unchanged. The vertical line marks the layer used for single-layer results.}}
\vskip -0.1in
\label{fig:norm-distance-layers}
\end{figure*}

\paragraph{Results \& Discussion.}
Table~\ref{tab:wvs} shows that countries a model represents alike tend to be alike in the values their people report. At the reported layer, model and survey distances correlate at .34 for Llama, .28 for Gemma, and .52 for Qwen over the 903 pairs. This is not a restatement of geography. Geographic distance alone correlates with the survey at .25, and membership in different world regions at .28, so the models track the survey at least as well as the map does. The correlation also survives when geography is removed from both sides, at .25, .23, and .44, each with a permutation $p$ of .0005. The states therefore carry cultural similarity that location does not explain.

Figure~\ref{fig:wvs-layers} shows that the agreement is present at every layer we sampled, in all three models, and that removing geography lowers it by about .06 to .10 throughout. The reported layers understate it for two models. Llama agrees with the survey most at layer 8, at .51, and Gemma at layer 40, at .53, against .34 and .28 at the layers fixed in advance, while Qwen is stable near .50 from layer 20 onward. We report the fixed layers to avoid choosing a layer by its result. The comparison covers 43 countries, and the survey distance is one of several ways to summarize cultural values, so the result shows agreement with this measure and not with culture in general.

\subsection{Country Distances Under a Stated Norm}

\paragraph{Setup.}
We ask whether a model still represents the country when the prompt also states a norm, and in particular when that norm contradicts the country. If a stated norm displaced what the model holds about the country, the arrangement of countries in its states should fall apart. We test this with the distances of Equation~\ref{eq:country-distance}. For each of the 2,775 pairs of countries we compute the distance three times, from the states under \textit{country-only}, \textit{country-norm}, and \textit{country-contradictory-norm}. The similarity of the arrangement under a condition $c$ to the arrangement with the country alone is the Spearman correlation of the two sets of distances over all pairs,
\begin{equation}
S^{c} = \rho\!\left(D^{\,\text{country-only}},\; D^{\,c}\right).
\label{eq:arrangement-similarity}
\end{equation}
A value of $1$ means that the same pairs of countries are close and far under both prompts, so the norm has left the arrangement unchanged, and a value of $0$ means that the arrangement is unrelated. We report $S^{c}$ for the true norm and for the contradictory norm at layer 16 for Llama, 26 for Gemma, and 36 for Qwen, with 95\% intervals from resampling countries, and at every layer to check that the result does not depend on the layer.

\begin{table}[h]
\centering
\small
\setlength{\tabcolsep}{0pt}
\begin{tabular*}{\columnwidth}{@{\extracolsep{\fill}}lccc@{}}
\toprule
 & \textbf{Llama} & \textbf{Gemma} & \textbf{Qwen} \\
\midrule
\multicolumn{4}{@{}l}{\textit{At the reported layer}} \\
True norm     & .95\ci{.94}{.97} & .73\ci{.65}{.80} & .94\ci{.92}{.95} \\
Contra.\ norm & .96\ci{.94}{.97} & .66\ci{.56}{.76} & .86\ci{.82}{.90} \\
\midrule
\multicolumn{4}{@{}l}{\textit{At the last layer}} \\
True norm     & .96 & .90 & .94 \\
Contra.\ norm & .95 & .85 & .79 \\
\bottomrule
\end{tabular*}
\caption{\small{Spearman correlation between the country distances with the country alone and those with a norm also stated, over 2,775 pairs of countries. The reported layer is 16, 26, and 36 for Llama, Gemma, and Qwen. Brackets give 95\% intervals from resampling countries, and Contra.\ abbreviates contradictory.}}
\vskip -0.1in
\label{tab:distances-under-norm}
\end{table}
\begin{figure*}[t]
\centering
\includegraphics[width=\textwidth]{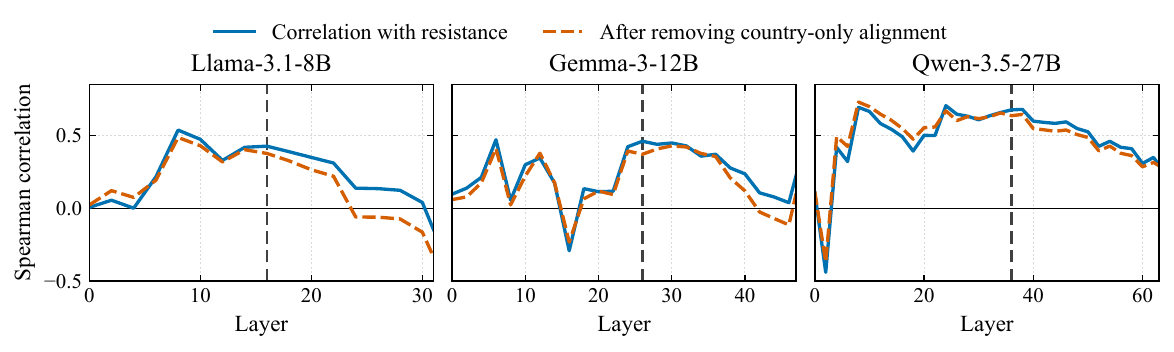}
\caption{\small{Spearman correlation between the distinctness of a country and its resistance rate under a contradictory norm, at each layer, before and after removing \textit{country-only} alignment. The vertical line marks the layer used for single-layer results.}}
\vskip -0.1in
\label{fig:distinctness-layers}
\end{figure*}

\paragraph{Results \& Discussion.}
Table~\ref{tab:distances-under-norm} shows that a stated norm does not erase how a model represents the country, and that the models differ in whether a contradictory norm disturbs it. In Llama it does not. At the reported layer, distances under a true norm correlate with the \textit{country-only} distances at .95, and under a contradictory norm at .96, with 95\% intervals of .94 to .97 for both. Figure~\ref{fig:norm-distance-layers} shows that the two curves are indistinguishable from layer 8 onward and stay near .95 to the last layer. Llama therefore arranges the countries in the same way whether the norm it is given is true or false. It keeps its picture of the country, and nothing in that picture marks the contradictory norm as different. Qwen behaves differently. The two curves coincide through the first thirty layers and then separate. The true norm stays near .94, whereas the contradictory norm falls to .86 at the reported layer, with an interval of .82 to .90 that lies below that of the true norm, and to .79 at the last layer, reaching about .65 at its lowest. In Qwen a contradictory norm thus changes how the country is represented in the later layers, while a true norm does not. The separation begins at layers 30 to 36, the depth at which the country starts to contribute to the readout and the conflict signal rises in Section~\ref{sec:mechanistic-analysis}. Qwen is also the model that resists a contradictory norm most often, in 42.9\% of answers against 17.6\% for Llama. The model whose representation of the country is disturbed by a contradictory norm is the model that most often rejects it.

Gemma falls between the two and is less clear. Its similarity is lower throughout, at .73 under a true norm and .66 under a contradictory one at the reported layer, with wide intervals of .65 to .80 and .56 to .76, and the contradictory norm stays below the true one from that layer to the end, at .85 against .90 at the last layer. The curves are irregular, with a sharp drop at layer 18 that also appears in Appendix~\ref{app:country-shift}, and we do not draw conclusions from Gemma here. In all three models the similarity is low in the first few layers, where the countries are not yet arranged in any stable way. These results are correlational and compare averages over the records of a country, so they describe the arrangement of countries, not single records.

\subsection{Country Distinctness and Resistance}

\paragraph{Setup.}
We ask whether a model resists a contradictory norm more for countries that it represents as different from the rest. A country whose representation is close to many others carries little that is specific to it, so almost any claimed norm may seem to fit, whereas a country represented distinctly gives a contradictory norm something to conflict with. We measure the \textit{distinctness} of country $k$ as its mean distance to the other countries,
\begin{equation}
\delta_k = \frac{1}{K-1}\sum_{j \neq k} D^{\,\text{shift}}_{jk},
\label{eq:distinctness}
\end{equation}
where $D^{\,\text{shift}}$ is the distance of Equation~\ref{eq:country-distance} computed on the mean shift of each country, the change in state from naming it in Equation~\ref{eq:shift}, in place of its mean state. Using the shift keeps only what the country adds to the prompt. The \textit{resistance rate} of a country is the percentage of its answers judged aligned under \textit{country-contradictory-norm}, averaged over the two judges. We report the Spearman correlation between $\delta_k$ and the resistance rate across the 75 countries, at layer 16 for Llama, 26 for Gemma, and 36 for Qwen, with 95\% intervals from resampling countries, and again without the five Anglophone countries. Countries a model knows better are also resisted more, so we repeat the correlation after removing \textit{country-only} alignment from both quantities, and at every layer.

\begin{table}[h]
\centering
\small
\setlength{\tabcolsep}{0pt}
\begin{tabular*}{\columnwidth}{@{\extracolsep{\fill}}lccc@{}}
\toprule
 & \textbf{Llama} & \textbf{Gemma} & \textbf{Qwen} \\
\midrule
\multicolumn{4}{@{}l}{\textit{At the reported layer}} \\
All countries  & .43\ci{.20}{.62} & .46\ci{.25}{.63} & .68\ci{.51}{.79} \\
No Anglophone  & .31\ci{.08}{.52} & .39\ci{.17}{.57} & .61\ci{.42}{.75} \\
Controlled     & .38\ci{.15}{.58} & .37\ci{.15}{.56} & .63\ci{.44}{.77} \\
\midrule
\multicolumn{4}{@{}l}{\textit{Across layers}} \\
Median         & .22 & .23 & .54 \\
Above zero     & 16 of 17 & 24 of 25 & 32 of 33 \\
\bottomrule
\end{tabular*}
\caption{\small{Spearman correlation between the distinctness of a country and its resistance rate under a contradictory norm, across 75 countries. Controlled removes \textit{country-only} alignment from both. Brackets give 95\% intervals from resampling countries.}}
\vskip -0.1in
\label{tab:distinctness}
\end{table}

\paragraph{Results \& Discussion.}
Table~\ref{tab:distinctness} shows that a model resists a contradictory norm more for the countries it represents as most different from the others. At the reported layer, distinctness correlates with the resistance rate at .43 for Llama, .46 for Gemma, and .68 for Qwen, with 95\% intervals of .20 to .62, .25 to .63, and .51 to .79. The relation is not carried by the Anglophone countries, as it remains at .31, .39, and .61 among the other 70, with intervals that exclude zero. Nor is it a restatement of how well the model knows a country. Distinctness and \textit{country-only} alignment are only weakly related, at .26, .31, and .31, and removing \textit{country-only} alignment from both quantities leaves the correlation at .38, .37, and .63. A country that is represented distinctly is therefore resisted more even among countries the model knows equally well. This fits the account in the main text. A contradictory norm needs something specific to conflict with, and a country whose representation is close to many others offers little.

Figure~\ref{fig:distinctness-layers} shows that the strength of this result depends on the model. In Qwen it is stable. The correlation is above .5 from layer 8 to layer 50 and positive at 32 of the 33 layers sampled, with a median of .54, and the control changes it little at any depth. In Llama and Gemma it is confined to the middle of the network. Llama peaks between layers 8 and 16 and declines to zero by the last layers, and Gemma is strongest from layer 24 to 36 after an irregular early stretch with a drop below zero at layer 16. Their medians across layers are .22 and .23, about half the value at the reported layer, which lies near the top of each curve. The layers were fixed before this analysis, but the single-layer figures for Llama and Gemma should be read as an upper range and not as typical of the network. The firm conclusion is for Qwen, the model that resists most, where distinctness tracks resistance at nearly every depth. For the other two the relation is present at mid-depth and weak elsewhere. The analysis is correlational over 75 countries and does not show that distinctness causes resistance.

\subsection{Probing for Resistance}
\label{app:probe}

\paragraph{Setup.}
We ask whether a model has already decided how to respond to a contradictory norm before it writes its first answer token. For each record $r$ we define a binary outcome,
\begin{equation}
y(r) =
\begin{cases}
1 & \text{if the answer resists,} \\
0 & \text{otherwise,}
\end{cases}
\label{eq:resist}
\end{equation}
where an answer \textit{resists} when the unconstrained response to the \textit{country-contradictory-norm} prompt is judged aligned with the documented norm. We then train a linear probe to predict $y(r)$ from the state $h(r)$ under that prompt. The probe is a ridge regression on the standardized state,
\begin{equation}
\begin{aligned}
\hat{y}(r) &= w^{\top} z(r) + b, \\
\mathcal{L}(w, b) &= \sum_{r} \bigl(y(r) - \hat{y}(r)\bigr)^{2} + \lambda \lVert w \rVert^{2},
\end{aligned}
\label{eq:probe}
\end{equation}
where $z(r)$ is $h(r)$ with each dimension standardized on the training records. We use five-fold cross-validation. Records that share a norm statement are kept in the same fold, so the probe is never tested on a norm it was trained on. In each round one fold is held out for testing, one is used to choose $\lambda$ from $\{10, 10^{2}, 10^{3}, 10^{4}, 10^{5}\}$ by its AUROC, and the other three are used for training. Every record is thus scored once by a probe that never saw it. We report the AUROC of $\hat{y}$ for $y$ on these held-out scores. Before computing it we replace each score by its rank among the records of the same country, so that a probe cannot succeed by separating countries that differ in how often they are resisted. Intervals are 95\% bootstrap intervals from resampling groups of records that share a norm statement. We train a separate probe at every layer.

\begin{table}[h]
\centering
\small
\setlength{\tabcolsep}{0pt}
\begin{tabular*}{\columnwidth}{@{\extracolsep{\fill}}lccc@{}}
\toprule
 & \textbf{Llama} & \textbf{Gemma} & \textbf{Qwen} \\
\midrule
\multicolumn{4}{@{}l}{\textit{AUROC at the reported layer}} \\
Gemma judge   & .791\ci{.779}{.804} & .841\ci{.829}{.851} & .917\ci{.911}{.922} \\
Qwen judge    & .798\ci{.786}{.811} & .837\ci{.824}{.848} & .910\ci{.905}{.916} \\
Both judges   & .795\ci{.783}{.808} & .848\ci{.835}{.859} & .912\ci{.906}{.917} \\
\midrule
\multicolumn{4}{@{}l}{\textit{AUROC across layers, Gemma judge}} \\
First layer   & .635 & .597 & .647 \\
Highest       & .798 & .841 & .923 \\
Last layer    & .773 & .824 & .921 \\
\midrule
\multicolumn{4}{@{}l}{\textit{Records that resist, of 9,261}} \\
Gemma judge   & 1,654 & 1,587 & 4,075 \\
\bottomrule
\end{tabular*}
\caption{\small{Linear probe for resistance on the state under \textit{country-contradictory-norm}. The reported layer is 16, 26, and 36 for Llama, Gemma, and Qwen. Both judges counts a record as resisting only when the two judges agree. Brackets give 95\% bootstrap intervals.}}
\vskip -0.1in
\label{tab:probe}
\end{table}

\paragraph{Results \& Discussion.}
Table~\ref{tab:probe} shows that whether a model will resist a contradictory norm can be read from its state before it writes any part of the answer. At the reported layer, the probe reaches an AUROC of .791 for Llama, .841 for Gemma, and .917 for Qwen under the Gemma judge, with 95\% intervals of .779 to .804, .829 to .851, and .911 to .922. An AUROC of .5 would mean that the state carries no information about the outcome, and 1 that it settles it completely, so the response is largely determined by the time the answer begins, and most fully in Qwen. The result does not depend on which judge labels the answers. Under the Qwen judge the AUROC is .798, .837, and .910, within .007 of the first, and it is .795, .848, and .912 when a record counts as resisting only if both judges agree. The probes are not fitted to a rare outcome in one model and a common one in another by accident of scoring, since ranking scores within each country removes differences between countries, but the number of resisting records does differ, at 1,654 and 1,587 of 9,261 for Llama and Gemma against 4,075 for Qwen.

The outcome becomes readable gradually and then stays so. Figure~\ref{fig:decision-depth} plots the probe at every layer. At the first layer the AUROC is .635, .597, and .647, close to what the words of the prompt alone give, as Appendix~\ref{app:probe-baselines} shows. It rises through the first half of the network and levels off, reaching its highest value of .798 at layer 14 in Llama, .841 at layer 24 in Gemma, and .923 at layer 48 in Qwen, and it remains at .773, .824, and .921 at the last layer. The layers we report, fixed in advance, are at or within .007 of these maxima, so the reported figures are representative and were not chosen to be favorable. That the signal persists to the last layer means the model does not revise its response late. A probe shows that the outcome is present in the state and not that the model uses that part of the state to produce it, which is what the interventions in the following sections test.

\subsection{Probe Baselines}
\label{app:probe-baselines}

\paragraph{Setup.}
We ask what the probe of Appendix~\ref{app:probe} is reading. A high AUROC could come from three sources other than the handling of the norm by the model, namely the wording of the prompt, properties of the record that are visible before any norm is stated, and the contradictory norm alone without the country. We test each with a probe that is denied the rest. All of them predict the same outcome $y(r)$ of Equation~\ref{eq:resist}, with the same folds, the same choice of $\lambda$, and the same scoring within countries. First, a \textit{bag-of-words} probe replaces the state with word counts of the \textit{country-contradictory-norm} prompt, keeping words that occur at least ten times. It measures what the text predicts without the model. Second, we train the probe on the state under \textit{no-evidence} and under \textit{country-only}. These prompts contain no norm, so the probe can only use what distinguishes the record itself, such as its situation and its country. Third, we train it on the state under \textit{contradictory-norm-only}, which shows the model the contradictory norm without the country. The contribution of the country is then the difference
\begin{equation}
\Delta = \text{AUROC}_{\text{country+contra}} - \text{AUROC}_{\text{contra}},
\label{eq:country-adds}
\end{equation}
between the probe on the full prompt and the probe on the contradictory norm alone. We report each AUROC with a 95\% bootstrap interval, and an interval for $\Delta$ from resampling the same groups of records for both probes, so that the two are compared on identical samples.

\begin{table}[h]
\centering
\small
\setlength{\tabcolsep}{0pt}
\begin{tabular*}{\columnwidth}{@{\extracolsep{\fill}}lccc@{}}
\toprule
 & \textbf{Llama} & \textbf{Gemma} & \textbf{Qwen} \\
\midrule
\multicolumn{4}{@{}l}{\textit{No model}} \\
Prompt words  & .652\ci{.636}{.667} & .683\ci{.669}{.698} & .683\ci{.672}{.694} \\
\midrule
\multicolumn{4}{@{}l}{\textit{State, no norm in the prompt}} \\
No evidence   & .623\ci{.608}{.639} & .666\ci{.650}{.680} & .691\ci{.680}{.701} \\
Country only  & .622\ci{.607}{.638} & .680\ci{.665}{.694} & .710\ci{.699}{.720} \\
\midrule
\multicolumn{4}{@{}l}{\textit{State, contradictory norm in the prompt}} \\
Norm only     & .747\ci{.733}{.761} & .790\ci{.777}{.802} & .844\ci{.836}{.851} \\
Full prompt   & .791\ci{.779}{.804} & .841\ci{.829}{.851} & .917\ci{.911}{.922} \\
\midrule
Gain from country & .045\ci{.034}{.054} & .051\ci{.041}{.060} & .073\ci{.067}{.079} \\
\bottomrule
\end{tabular*}
\caption{\small{AUROC for resistance from each probe input, at layer 16, 26, and 36 for Llama, Gemma, and Qwen, under the Gemma judge. Norm only is the contradictory norm without the country, and Full prompt adds the country. Gain from country is the difference between the two. Brackets give 95\% bootstrap intervals.}}
\vskip -0.1in
\label{tab:probe-baselines}
\end{table}

\paragraph{Results \& Discussion.}
Table~\ref{tab:probe-baselines} shows that the probe reads how the model has processed the contradictory norm, and not the wording of the prompt or properties of the record. The words of the prompt alone predict resistance at an AUROC of .652 for Llama, .683 for Gemma, and .683 for Qwen. This is above chance, because some situations and some norms are resisted more than others and the text identifies them, but it is .139, .158, and .234 below the probe on the state. The state adds nothing to the words until the norm has been read. Under \textit{no-evidence} the probe reaches .623, .666, and .691, and under \textit{country-only} .622, .680, and .710, each within .03 of the bag-of-words probe. What a model will do with a contradictory norm is therefore not settled by the record or by the country before the norm appears.

The signal arrives with the norm. On the state under the contradictory norm without the country, the probe reaches .747, .790, and .844, well above every baseline that lacks the norm. Adding the country raises it by .045, .051, and .073, with 95\% intervals of .034 to .054, .041 to .060, and .067 to .079, all excluding zero. Under the Qwen judge the gain is .043, .041, and .072. Of the improvement over the bag of words, about two thirds is thus present once the model has read the contradictory norm alone, and the remaining third appears only when the country is also given. The response depends mostly on the stated norm, and to a smaller but reliable degree on how that norm meets the country. That second part is largest in Qwen, the model that resists most. It is the part of the state that would reflect a comparison of the norm with the country, and Appendix~\ref{app:conflict-signal} examines it.

\subsection{Controlling for Knowledge of the Norm}

\paragraph{Setup.}
We ask whether the probe of Appendix~\ref{app:probe} predicts resistance only because it can tell whether the model knows the norm. The concern is concrete. A model that does not know what is customary in a country has nothing to set against a contradictory norm, and so usually follows it. Resistance is then far more common on records the model knows than on those it does not, and a probe could reach a high AUROC by reading knowledge alone, without reading anything about how the model handles the contradictory norm. To separate the two, we keep the probe unchanged and restrict where it is scored. Let $a(r) = 1$ when the answer of the model to the \textit{country-only} prompt is judged aligned, so that the model demonstrably produces the documented behavior from the country alone. We compute the AUROC on the held-out scores of those records only,
\begin{equation}
\begin{aligned}
\mathcal{K} &= \{\, r : a(r) = 1 \,\}, \\
\text{AUROC}_{\mathcal{K}} &= \text{AUROC}\bigl(\hat{y},\, y \mid r \in \mathcal{K}\bigr).
\end{aligned}
\label{eq:known-subset}
\end{equation}
Within $\mathcal{K}$ every record is one the model knows, so knowledge can no longer distinguish the records that resist from those that do not. If the probe were only reading knowledge, its AUROC on $\mathcal{K}$ would fall toward .5. If it reads how the model handles the norm, the AUROC should stay close to its value on all records. The probes, folds, and scoring are those of Appendix~\ref{app:probe}, with no retraining.

\begin{table}[h]
\centering
\small
\setlength{\tabcolsep}{0pt}
\begin{tabular*}{\columnwidth}{@{\extracolsep{\fill}}lccc@{}}
\toprule
 & \textbf{Llama} & \textbf{Gemma} & \textbf{Qwen} \\
\midrule
\multicolumn{4}{@{}l}{\textit{AUROC of the probe}} \\
All records  & .791\ci{.779}{.804} & .841\ci{.829}{.851} & .917\ci{.911}{.922} \\
Known        & .779\ci{.765}{.794} & .831\ci{.819}{.843} & .915\ci{.909}{.921} \\
Not known    & .779\ci{.748}{.808} & .839\ci{.808}{.869} & .909\ci{.896}{.922} \\
\midrule
\multicolumn{4}{@{}l}{\textit{Records}} \\
Known        & 5,362 & 5,934 & 6,965 \\
Not known    & 3,899 & 3,327 & 2,296 \\
\midrule
\multicolumn{4}{@{}l}{\textit{Share that resist (\%)}} \\
Known        & 26.1 & 23.5 & 52.2 \\
Not known    & 6.5 & 5.8 & 19.2 \\
\bottomrule
\end{tabular*}
\caption{\small{The resistance probe scored separately on records the model answers correctly from the country alone, called known, and on the rest. Layers and judge as in Table~\ref{tab:probe}. Brackets give 95\% bootstrap intervals.}}
\vskip -0.1in
\label{tab:probe-known}
\end{table}
\begin{figure*}[t]
\centering
\includegraphics[width=\textwidth]{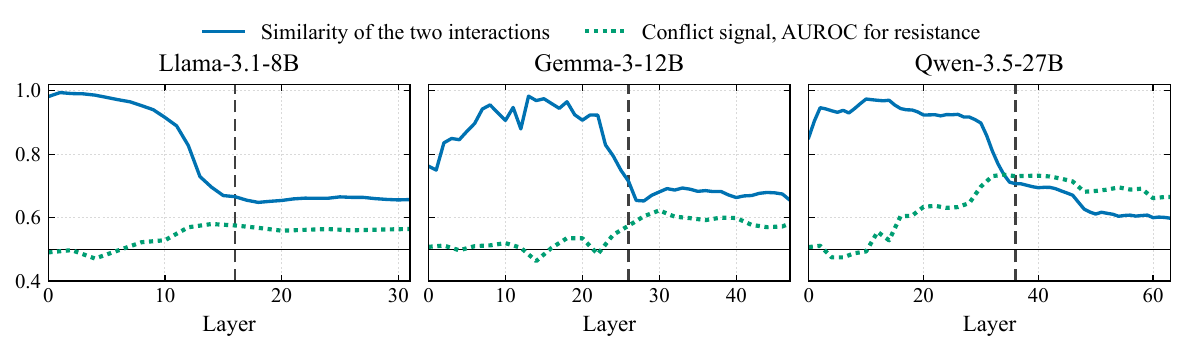}
\caption{\small{The solid line is the mean cosine similarity between the interaction of a record under its true norm and under its contradictory norm, at each layer. The dotted line is the AUROC of the conflict signal for resistance, with chance at .5. The vertical line marks the layer used for single-layer results.}}
\vskip -0.1in
\label{fig:interaction-layers}
\end{figure*}

\paragraph{Results \& Discussion.}
Table~\ref{tab:probe-known} shows that the probe is not reading whether the model knows the norm. The concern was justified. Resistance is far more common on records the model answers correctly from the country alone, at 26.1\%, 23.5\%, and 52.2\% for Llama, Gemma, and Qwen, than on the rest, at 6.5\%, 5.8\%, and 19.2\%, so a probe that detected only knowledge would still predict resistance well across all records. It does not explain the result. On the known records alone, where every record is one the model knows and knowledge cannot separate those that resist from those that do not, the AUROC is .779, .831, and .915, with 95\% intervals of .765 to .794, .819 to .843, and .909 to .921. This is within .012 of the AUROC on all records, .791, .841, and .917, and far from the .5 that a probe reading only knowledge would give.

The same holds on the other side. On the records the model does not answer correctly from the country alone, the AUROC is .779, .839, and .909. These groups are smaller and contain few resisting records, 253 for Llama and 194 for Gemma, so the intervals are wider, at .748 to .808, .808 to .869, and .896 to .922, but all lie well above .5. The probe therefore separates resisting from non-resisting records equally well whether or not the model knows the norm. Knowledge changes how often a model resists, and the probe captures something else, which records within each group the model will resist on. Together with Appendix~\ref{app:probe-baselines}, this leaves the handling of the contradictory norm as what the probe reads. It also shows that resistance is not confined to records the model knows. A model resists on a minority of records it gets wrong from the country alone, and its state marks those in advance too.

\subsection{The Conflict Signal}
\label{app:conflict-signal}

\paragraph{Setup.}
We ask whether the state of a model registers that a stated norm conflicts with the country, using a measure built without the model's answers. The probe of Appendix~\ref{app:probe} is trained on the outcome and may pick up anything that predicts it. Here we fix a direction in advance from the prompts alone and only then test it against the outcome. The idea is to isolate what the state contains only because a country and a norm appear together. Write $\varnothing$ for \textit{no-evidence}, $\text{C}$ for \textit{country-only}, $n$ for a norm stated without the country, and $\text{C}{+}n$ for the country with that norm, where $n$ is either the true norm or the contradictory one. If the model simply added the effect of the country to the effect of the norm, the state under both would be $h_{\text{C}} + h_{n} - h_{\varnothing}$. The \textit{interaction} is what remains beyond that,
\begin{equation}
I^{n}(r) = h_{\text{C}+n}(r) - h_{\text{C}}(r) - h_{n}(r) + h_{\varnothing}(r).
\label{eq:interaction}
\end{equation}
Anything the model does to relate a norm to a country must appear in this term. We compute it for the true norm and for the contradictory norm of each record, and take the direction along which the two differ on average,
\begin{equation}
d = \frac{\overline{I^{\text{contra}} - I^{\text{true}}}}{\bigl\lVert\, \overline{I^{\text{contra}} - I^{\text{true}}} \,\bigr\rVert},
\label{eq:conflict-direction}
\end{equation}
where the bar is a mean over records. The \textit{conflict signal} of a record is the projection of its contradictory-norm interaction on this direction,
\begin{equation}
c(r) = I^{\text{contra}}(r) \cdot d.
\label{eq:conflict-signal}
\end{equation}
The direction uses only which norm was shown and never the judged answers. It is estimated with the five folds of Appendix~\ref{app:probe}, so that the signal of a record is computed with a direction from the other folds. We then test the signal in two ways. Across records, we report the AUROC of $c(r)$ for the outcome $y(r)$ of Equation~\ref{eq:resist}, with scores ranked within each country. Across countries, we report the Spearman correlation between the mean signal of a country and its resistance rate. No probe is trained at any point.

\begin{table}[h]
\centering
\small
\setlength{\tabcolsep}{0pt}
\begin{tabular*}{\columnwidth}{@{\extracolsep{\fill}}lccc@{}}
\toprule
 & \textbf{Llama} & \textbf{Gemma} & \textbf{Qwen} \\
\midrule
\multicolumn{4}{@{}l}{\textit{AUROC of the conflict signal for resistance}} \\
Gemma judge   & .575\ci{.560}{.590} & .575\ci{.558}{.591} & .731\ci{.720}{.741} \\
Qwen judge    & .579\ci{.565}{.595} & .578\ci{.561}{.594} & .723\ci{.712}{.733} \\
Trained probe & .791 & .841 & .917 \\
\midrule
\multicolumn{4}{@{}l}{\textit{Across 75 countries, $\rho$ with resistance rate}} \\
Conflict signal & .28\ci{.05}{.49} & .31\ci{.06}{.53} & .77\ci{.65}{.87} \\
\midrule
\multicolumn{4}{@{}l}{\textit{Similarity of true and contradictory interactions}} \\
Early layers    & .99 & .96 & .97 \\
Reported layer  & .67 & .71 & .71 \\
\bottomrule
\end{tabular*}
\caption{\small{The conflict signal at layer 16, 26, and 36 for Llama, Gemma, and Qwen. The trained probe of Table~\ref{tab:probe} is shown for comparison. Similarity is the mean cosine between the interaction of a record under its true norm and under its contradictory norm. Brackets give 95\% bootstrap intervals.}}
\vskip -0.1in
\label{tab:conflict-signal}
\end{table}
\begin{figure*}[t]
\centering
\includegraphics[width=\textwidth]{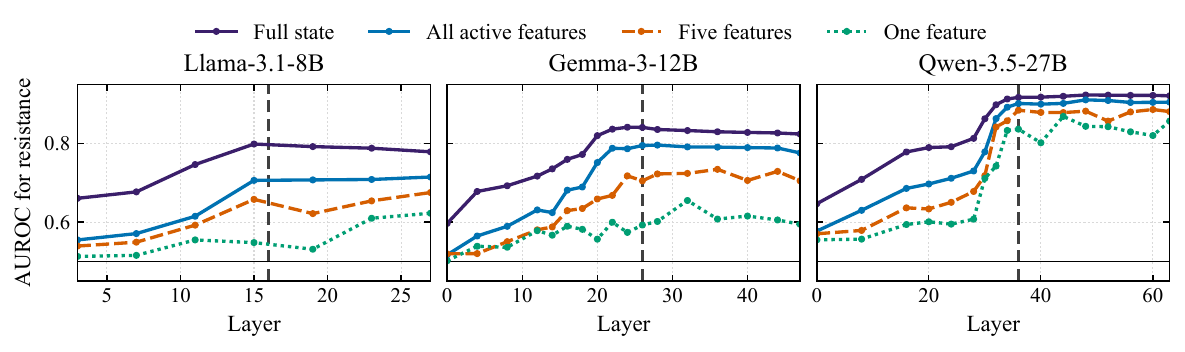}
\caption{\small{AUROC for resistance under a contradictory norm at each layer with an autoencoder, from the most predictive feature, the five most predictive, all active features, and the full state. Chance is .5. The vertical line marks the layer used for single-layer results.}}
\vskip -0.1in
\label{fig:sae-layers}
\end{figure*}

\paragraph{Results \& Discussion.}
Table~\ref{tab:conflict-signal} shows that every model comes to treat a contradictory norm differently from a true one, but that this difference governs the response only in Qwen. Figure~\ref{fig:interaction-layers} traces the first part. In the early layers the interaction of a record under its true norm and under its contradictory norm are almost the same, with a mean cosine similarity of .99 for Llama, .96 for Gemma, and .97 for Qwen. The similarity then falls sharply, across layers 10 to 14 in Llama, 22 to 28 in Gemma, and 28 to 36 in Qwen, and settles at .67, .71, and .71 at the reported layer. From that depth on, the way a norm combines with the country depends on whether the norm is true. These are the layers at which the answer reads the norm in Section~\ref{sec:mechanistic-analysis}.

The second part is what separates the models. At the reported layer the conflict signal predicts resistance at an AUROC of .575 for Llama and .575 for Gemma, with 95\% intervals of .560 to .590 and .558 to .591, against .731 for Qwen, with an interval of .720 to .741. The Qwen judge gives .579, .578, and .723. The intervals exclude .5 in every model, but in Llama and Gemma the signal is slight, and far below the .791 and .841 of the trained probe. Across the 75 countries, the mean signal of a country correlates with its resistance rate at .28 and .31 in Llama and Gemma, with intervals reaching .05 and .06, and at .77 in Qwen. All three models therefore register that a contradictory norm stands differently to the country than a true one does, yet only in Qwen is that registration tied to whether the model resists. The signal uses one direction per layer, so it is a lower bound on what the state carries about the conflict, and does not show that a model checks the norm against that country.

\subsection{Sparse Autoencoder Features}

\paragraph{Setup.}
We ask whether the decision to resist is carried by a few interpretable features of the state, and whether such features cause the response or only reflect it. A sparse autoencoder rewrites a state as a sum of a small number of active features, each with a fixed direction,
\begin{equation}
\begin{aligned}
a(r) &= \operatorname{enc}\bigl(h(r)\bigr), \\
h(r) &\approx \sum_{i} a_i(r)\, d_i + b,
\end{aligned}
\label{eq:sae}
\end{equation}
where $a_i(r) \geq 0$ is the activation of feature $i$ on record $r$, most of which are zero, and $d_i$ is its direction. We use publicly released autoencoders trained on each model, and apply them to the states of Appendix~\ref{app:state-extraction} under \textit{country-contradictory-norm} at the layers they cover. We first measure how well they fit, as the reconstruction error relative to the variation of the state across records. We then test how much of the resistance signal survives in the features. Using the folds and within-country scoring of Appendix~\ref{app:probe}, we report the AUROC for resistance from the single most predictive feature, from the five most predictive, and from all active features, against the probe on the full state. Features are chosen on training records and scored on held-out ones. To describe what a selected feature tracks, we report how often it is active under each evidence condition and on answers that resist and do not. Finally, we intervene on features in Qwen while it generates its answer. For a feature $i$ we set its activation to a target value $v$ at every position after the text of the user,
\begin{equation}
h \;\leftarrow\; h + \bigl(v - a_i(h)\bigr)\, d_i,
\label{eq:sae-steer}
\end{equation}
with $v = 0$ to switch the feature off and a high value of its observed activation to switch it on. We do this for one feature that is active when the model resists, one that is active when it follows the norm, and a randomly chosen feature from the same layer as a control, on 200 records, and judge the answers with both judges.

\begin{table}[t]
\centering
\small
\setlength{\tabcolsep}{4pt}
\begin{tabular}{lcc}
\toprule
\textbf{Intervention} & \textbf{Resists} & \textbf{$\Delta$ Random} \\
\midrule
None & 46.2 & -- \\
\midrule
Resist feature off & 46.2 & $+$0.2\ci{-2.2}{2.8} \\
Resist feature on & 49.5 & $+$3.5\ci{0.0}{7.0} \\
Resist feature on, $3\times$ & 49.5 & $+$4.8\ci{0.8}{9.0} \\
\midrule
Follow feature off & 47.8 & $+$1.8\ci{-0.5}{4.2} \\
Follow feature on & 46.5 & $+$0.5\ci{-2.2}{3.5} \\
\midrule
Random feature off & 46.0 & -- \\
Random feature on & 46.0 & -- \\
Random feature on, $3\times$ & 44.8 & -- \\
\bottomrule
\end{tabular}
\caption{\small{Share of 200 Qwen answers that resist a contradictory norm when one feature is switched off or on during generation, averaged over the two judges. $\Delta$ Random is the paired change against the same intervention on a random feature, with a 95\% bootstrap interval.}}
\vskip -0.1in
\label{tab:sae-steering}
\end{table}

\paragraph{Results \& Discussion.}
Figure~\ref{fig:sae-layers} shows that only Qwen concentrates the decision in a few features. From layer 34 on, its single most predictive feature reaches an AUROC of .83 to .87 against .91 to .92 for the full state, and the same feature is selected in every fold. Before layer 30 no single feature exceeds .61, so the feature appears at the depth where the probe of Appendix~\ref{app:probe} rises. In Llama and Gemma the best single feature reaches .62 and .66, and all active features together reach .72 and .79 against .78 and .83 for the full state. The decision in these two models is spread over many directions, and part of it is lost by the autoencoders, whose reconstruction error exceeds the variation of the state across records at almost every layer. The Qwen feature at layer 34 is active on 0.2\% of records under \textit{no-evidence}, 12\% under \textit{country-only}, 19\% under \textit{country-norm}, and 60\% under \textit{country-contradictory-norm}. Under a contradictory norm it is active on 89\% of answers that resist and 37\% of those that do not, and under a true norm it is active on 65\% of the answers that do not align and 17\% of those that do. It therefore tracks that the model is about to depart from the stated norm, and not that the norm is false.

Table~\ref{tab:sae-steering} shows that these features predict the response far better than they control it. Switching either feature off leaves resistance within two points of the untouched rate of 46.2\%. Switching the resist feature on raises resistance to 49.5\%, a gain of 3.5 points over the random feature with an interval that reaches zero, and tripling its value does not increase the rate. Alignment under a true norm stays near 97\% in every arm. The interventions changed the wording of at least 83\% of the answers, so the small effect is not a failed intervention. A feature that predicts resistance at .83 moves about three answers in a hundred, which indicates that by these layers the response is carried by many directions and a single feature reads it out.

\subsection{Attention Knockout of the Norm}
\label{app:knockout}

\paragraph{Setup.}
We ask at which layers the stated norm enters the answer, and whether the model returns to its own knowledge once the norm is cut off. Let $S$ be a span of prompt tokens and $\mathcal{W}$ a set of layers. In the layers of $\mathcal{W}$ we stop every position after the text of the user from attending to $S$, by setting the attention scores to $-\infty$ before the softmax,
\begin{equation}
\alpha^{(\ell)}_{t \to s} = 0
\quad \text{for } \ell \in \mathcal{W},\; s \in S,\; t > T,
\label{eq:knockout}
\end{equation}
where $\alpha^{(\ell)}_{t \to s}$ is the attention from position $t$ to position $s$ at layer $\ell$ in every head, and $T$ is the last token of the user. The norm stays in the prompt and is processed as usual at its own positions, so the intervention removes only the path from the norm to the answer. We block three spans under \textit{country-contradictory-norm}. These are the norm, the country context, and a control span of the same length as the norm taken from the scenario. We first locate the layers with a fast measure, the margin
\begin{equation}
m = \log p(y^{+}) - \log p(y^{-}),
\label{eq:margin}
\end{equation}
where $y^{+}$ is the reference answer and $y^{-}$ the answer that follows the contradictory norm, and we report its change when a span is blocked in every layer and in each window of layers. We then let the model generate its full answer under four interventions on the norm and judge it as in Section~\ref{sec:behavioral-evaluation}. The norm is blocked in every layer, in the norm band of layers 12 to 15 for Llama and 18 to 23 for Gemma, and in a later band of the same width, layers 24 to 27 and 36 to 41. We also block the country context and the control span in every layer, and block a true norm in every layer under \textit{country-norm}. Each rate is compared with the same records under \textit{country-only}, where no norm is stated. The bands were fixed in an exploratory run on 200 other records, and all results here use a fresh sample of 400 records per model. Differences between arms are paired on records, with 95\% bootstrap intervals. Qwen is not included because most of its layers pass information between tokens through a block that cannot be masked, and Appendix~\ref{app:erasure} covers it with a different intervention.

\begin{figure}[t]
\centering
\includegraphics[width=\columnwidth]{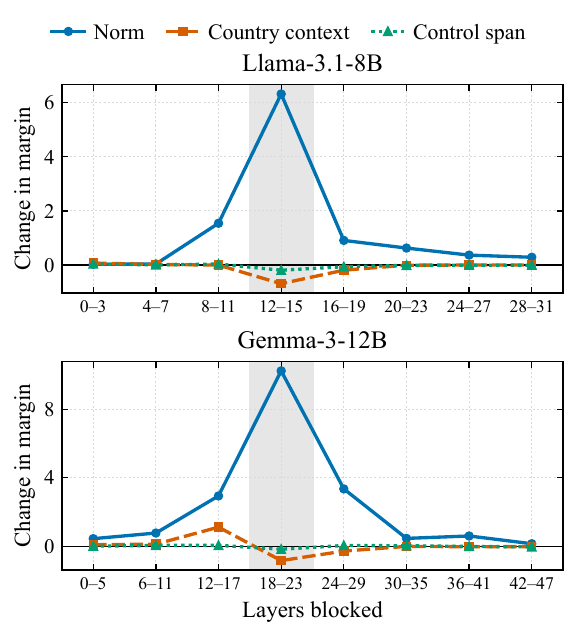}
\caption{\small{Mean change in the margin under \textit{country-contradictory-norm} when the answer cannot attend to a span in one window of layers, over 400 records. A positive change moves the model toward the reference answer. The shaded window is the norm band fixed beforehand.}}
\vskip -0.1in
\label{fig:knockout-windows}
\end{figure}

\paragraph{Results \& Discussion.}
Figure~\ref{fig:knockout-windows} shows that the answer reads the norm in one band of layers. Blocking the norm in every layer moves the margin toward the reference answer by 12.5 for Llama and 22.4 for Gemma. The norm band alone accounts for 6.3 and 10.2 of this, about half, and no other window exceeds 1.5 and 3.4. Blocking the country context or the control span in every layer moves the margin by less than 1 in both models, so the effect is specific to the norm and is not a general cost of removing attention.

\begin{table}[t]
\centering
\small
\setlength{\tabcolsep}{3pt}
\begin{tabular}{lcc}
\toprule
\textbf{Blocked} & \textbf{Llama} & \textbf{Gemma} \\
\midrule
Norm, all layers & $+$40.0\ci{35.0}{45.3} & $+$52.2\ci{47.0}{57.5} \\
Norm, norm band & $+$29.0\ci{24.2}{34.0} & $+$44.8\ci{39.8}{50.0} \\
Norm, later band & $+$1.5\ci{-1.0}{4.0} & $+$2.2\ci{0.0}{4.5} \\
\midrule
Country context & $-$1.8\ci{-5.2}{1.8} & $-$4.2\ci{-7.5}{-1.0} \\
Control span & $-$4.2\ci{-7.2}{-1.2} & $+$0.8\ci{-2.2}{3.8} \\
\midrule
Norm vs.\ no norm & $-$0.8\ci{-4.2}{2.8} & $-$1.5\ci{-5.5}{2.8} \\
\bottomrule
\end{tabular}
\caption{\small{Change in the percentage of answers that resist a contradictory norm when a span is blocked, paired on 400 records, with 95\% bootstrap intervals. The last row compares the norm blocked in all layers with \textit{country-only}.}}
\vskip -0.1in
\label{tab:knockout-effects}
\end{table}
\begin{figure*}[t]
\centering
\includegraphics[width=\textwidth]{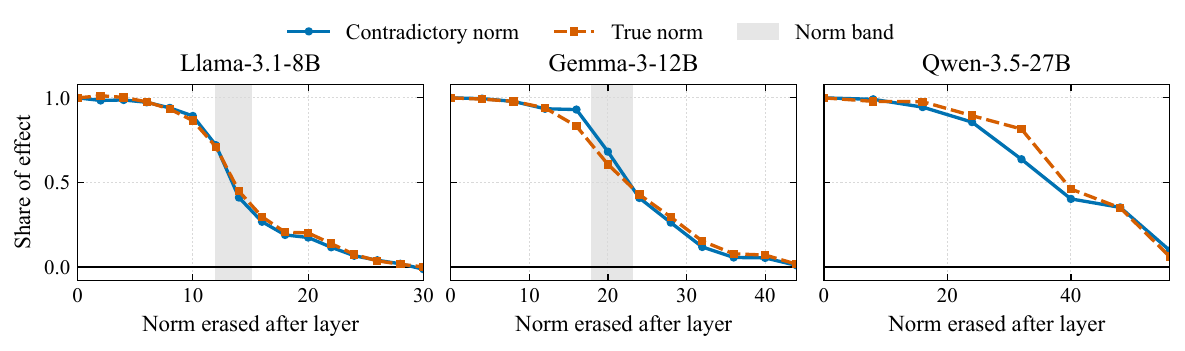}
\caption{\small{Share of the effect of the stated norm on the margin that is still removed when the norm is erased after a given layer, relative to erasing after the first layer, for a contradictory and a true norm. The shaded layers are the norm band of the knockout.}}
\vskip -0.1in
\label{fig:erasure-layers}
\end{figure*}

Table~\ref{tab:knockout-effects} shows the same pattern in the generated answers. With the norm blocked in every layer, resistance rises from 18.0\% to 58.0\% for Llama and from 14.2\% to 66.5\% for Gemma. These rates are within 1.5 points of the 58.8\% and 68.0\% that the same records reach under \textit{country-only}, and the intervals of the difference include zero. Blocking a true norm under \textit{country-norm} gives the same rates, 58.0\% and 67.5\%. The model therefore returns to what it answers from the country alone once it cannot read the norm, which shows that the contradictory norm does not remove its knowledge of the country. Blocking the norm only in the norm band recovers 29.0 and 44.8 points, 73\% and 86\% of the full effect, while the later band of the same width recovers 1.5 and 2.2 points with intervals that reach or include zero. Blocking the country context lowers resistance by 1.8 and 4.2 points, and only the Gemma interval excludes zero, so the country contributes little to resistance once the norm is stated. The control span lowers resistance by 4.2 points for Llama, a small effect in the opposite direction to that of the norm.

\subsection{Norm Erasure by Depth}
\label{app:erasure}

\paragraph{Setup.}
We ask by which depth the norm has been taken up from its own positions, with an intervention that also applies to Qwen. Attention knockout cuts the path from the norm to the answer in chosen layers. Erasure instead removes the norm itself from a chosen depth onward. Let $N$ be the positions of the norm and $Q$ the positions of the scenario. After layer $\ell$ we replace the state at every position of the norm with the mean state of the scenario at that layer,
\begin{equation}
h^{(\ell)}_t \;\leftarrow\; \frac{1}{|Q|} \sum_{q \in Q} h^{(\ell)}_q
\quad \text{for } t \in N,
\label{eq:erasure}
\end{equation}
so every later layer finds nothing of the norm at those positions. Whatever other positions copied from the norm before layer $\ell$ is left in place. Erasing after the first layer therefore removes the norm almost entirely, and erasing after a deeper layer removes only what the model would still have read from those positions. The intervention needs no attention mask, so it runs on all three models. We first sweep the layer $\ell$ under \textit{country-contradictory-norm} and \textit{country-norm} and report the change in the margin of Equation~\ref{eq:margin} as a share of the change from erasing after the first layer,
\begin{equation}
\pi(\ell) = \frac{\Delta m(\ell)}{\Delta m(0)}.
\label{eq:erasure-share}
\end{equation}
We then let the model generate its full answer with the norm erased at three depths fixed beforehand, after the first layer, before the norm band, and after it. These are layers 0, 8, and 16 for Llama, 0, 12, and 24 for Gemma, and 0, 24, and 36 for Qwen. The answers are judged as in Section~\ref{sec:behavioral-evaluation}, and each rate is compared with the same records left untouched. We use the 400 records of Appendix~\ref{app:knockout} for Llama and Gemma and 200 of them for Qwen, with differences paired on records and 95\% bootstrap intervals.

\begin{table}[t]
\centering
\small
\setlength{\tabcolsep}{3pt}
\begin{tabular}{lccc}
\toprule
\textbf{Norm} & \textbf{Llama} & \textbf{Gemma} & \textbf{Qwen} \\
\midrule
Untouched & 18.0 & 14.2 & 47.0 \\
Erased, first layer & 56.2 & 64.8 & 78.0 \\
Erased, before band & 61.0 & 68.2 & 76.5 \\
Erased, after band & 26.2 & 21.8 & 54.5 \\
\midrule
$\Delta$ before band & $+$43.0 & $+$54.0 & $+$29.5 \\
 & \ci{37.8}{48.5} & \ci{48.8}{59.0} & \ci{22.0}{37.0} \\
$\Delta$ after band & $+$8.2 & $+$7.5 & $+$7.5 \\
 & \ci{4.5}{12.0} & \ci{4.0}{11.2} & \ci{1.5}{13.5} \\
\bottomrule
\end{tabular}
\caption{\small{Percentage of answers that resist a contradictory norm when the norm is erased at three depths, and the paired change against the untouched prompt with 95\% bootstrap intervals, on 400 records for Llama and Gemma and 200 for Qwen.}}
\vskip -0.1in
\label{tab:erasure-effects}
\end{table}

\paragraph{Results \& Discussion.}
Figure~\ref{fig:erasure-layers} shows that the norm leaves its own positions over a narrow range of layers in Llama and Gemma and over a wider and later range in Qwen. Erasing the norm after the first layer moves the margin by 12.6, 20.9, and 8.6 for Llama, Gemma, and Qwen. Erasing it just before the norm band still removes 89\% and 93\% of this effect for Llama and Gemma, and erasing it at the end of the band removes 27\% and 41\%. The band where the answer reads the norm in Appendix~\ref{app:knockout} is therefore also where the norm stops being needed at its own positions. In Qwen the share falls from 86\% at layer 24 to 64\% at layer 32 and 10\% at layer 56, so the norm is taken up later and more gradually. The curves for a true and a contradictory norm nearly coincide in all three models, which indicates that a model takes up a stated norm at the same depth whether or not it fits the country.

Table~\ref{tab:erasure-effects} shows the same pattern in the generated answers. Erasing the norm before the band raises resistance by 43.0 and 54.0 points for Llama and Gemma, close to the 40.0 and 52.2 points from blocking attention to the norm in every layer, and the resulting rates of 61.0\% and 68.2\% match the 58.8\% and 68.0\% under \textit{country-only}. Two different interventions therefore agree on the size of the effect of the norm. Erasing the norm after the band raises resistance by only 8.2 and 7.5 points, 19\% and 14\% of the effect before the band. Erasure also gives a causal result for Qwen, which the knockout could not cover. Its resistance rises from 47.0\% to 76.5\% when the norm is erased before layer 24, and by 7.5 points when it is erased after layer 36. This later effect is not reliable, since its interval under the second judge includes zero.

\subsection{Head-Level Knockout}

\paragraph{Setup.}
We ask whether the norm is read by a few attention heads inside the norm band or by the band as a whole. The knockout of Appendix~\ref{app:knockout} blocks the norm in every head of a layer. Here we block it in a chosen set of heads $\mathcal{H}$ and leave the others untouched,
\begin{equation}
\alpha^{(\ell, j)}_{t \to s} = 0
\quad \text{for } (\ell, j) \in \mathcal{H},\; s \in N,\; t > T,
\label{eq:head-knockout}
\end{equation}
where $\alpha^{(\ell, j)}$ is the attention of head $j$ at layer $\ell$ and $N$ is the positions of the norm. The norm band holds 128 heads in Llama and 96 in Gemma. We proceed in three steps under \textit{country-contradictory-norm}. We first block each head of the band on its own and rank the heads by the change $\Delta m$ in the margin of Equation~\ref{eq:margin}. We then block the $k$ highest ranked heads together for $k$ of 1, 2, 4, 8, 16, and 32, and select the smallest set whose joint effect reaches half that of the whole band,
\begin{equation}
k^{\ast} = \min \bigl\{ k : \Delta m(\mathcal{H}_k) \geq \tfrac{1}{2}\, \Delta m(\mathcal{H}_{\text{band}}) \bigr\},
\label{eq:head-selection}
\end{equation}
where $\mathcal{H}_k$ is the $k$ highest ranked heads and $\mathcal{H}_{\text{band}}$ is every head of the band. This rule was fixed before the ranking was run. Finally we let the model generate its full answer with the norm blocked in the selected heads, and in an equal number of heads drawn at random from the band as a control, and judge the answers as in Section~\ref{sec:behavioral-evaluation}. The ranking and the selection use the 200 exploratory records, and the generated answers use the 400 records of Appendix~\ref{app:knockout}, so the heads are tested on records that did not choose them. Differences are paired on records with 95\% bootstrap intervals. Qwen is not included for the reason given in Appendix~\ref{app:knockout}.

\begin{figure}[h]
\centering
\includegraphics[width=\columnwidth]{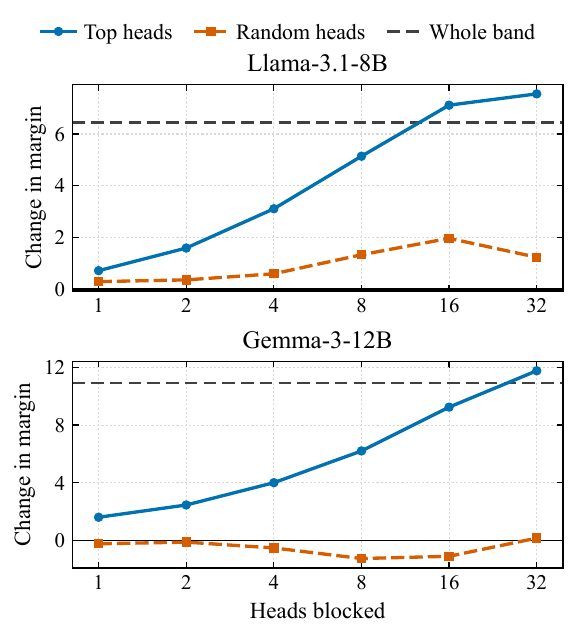}
\caption{\small{Mean change in the margin when the answer cannot attend to the contradictory norm through the highest ranked heads of the norm band, or through the same number of random heads, on the 200 exploratory records. The dashed line is the effect of blocking every head of the band.}}
\vskip -0.1in
\label{fig:heads-joint}
\end{figure}

\paragraph{Results \& Discussion.}
Figure~\ref{fig:heads-joint} shows that a small set of heads carries a large share of the effect of the band on the margin. The selection rule picks eight heads in both models. In Llama they move the margin by 5.1 against 6.4 for the whole band, and in Gemma by 6.2 against 10.9, which is 80\% and 57\% of the band. The same number of random heads moves the margin by 1.3 in Llama and by $-$1.2 in Gemma. In Llama the 16 highest ranked heads exceed the whole band, at 7.1, which indicates that some of the remaining heads of the band push the margin in the other direction.

\begin{table}[h]
\centering
\small
\setlength{\tabcolsep}{3pt}
\begin{tabular}{lcc}
\toprule
\textbf{Norm blocked in} & \textbf{Llama} & \textbf{Gemma} \\
\midrule
No heads & 18.0 & 14.2 \\
Eight selected heads & 46.5 & 23.5 \\
Eight random heads & 20.5 & 13.0 \\
Whole band & 47.0 & 59.0 \\
\midrule
$\Delta$ selected & $+$28.5\ci{23.8}{33.2} & $+$9.2\ci{5.2}{13.0} \\
$\Delta$ random & $+$2.5\ci{-0.5}{5.5} & $-$1.2\ci{-4.0}{1.2} \\
\bottomrule
\end{tabular}
\caption{\small{Percentage of answers that resist a contradictory norm when the norm is blocked in a set of heads of the norm band, and the paired change against the untouched prompt with 95\% bootstrap intervals, on 400 records.}}
\vskip -0.1in
\label{tab:heads-effects}
\end{table}

Table~\ref{tab:heads-effects} shows that this concentration holds in the generated answers for Llama and not for Gemma. In Llama, blocking the norm in the eight selected heads raises resistance from 18.0\% to 46.5\%, against 47.0\% for the whole band, so eight heads account for the full effect of the band on what the model writes. These are 6\% of the heads of the band and under 1\% of the heads of the model, and they were chosen on other records and by a different measure. In Gemma the selected heads raise resistance by 9.2 points, about a fifth of the 44.8 points from the whole band. The effect is real, since random heads change resistance by $-$1.2 points with an interval that includes zero, but most of the reading of the norm in Gemma is spread over the other heads of the band. The margin therefore overstated the share of the selected heads in Gemma, 57\% against 21\% in the judged answers, which is why we use the margin only to choose where to intervene and draw conclusions from the judged answers.

\subsection{Probe Transfer Between True and Contradictory Norms}

\paragraph{Setup.}
We ask whether a model represents its decision to follow a stated norm in the same way for a true and a contradictory norm. An answer follows the stated norm when it is aligned under a true norm and when it is not aligned under a contradictory norm,
\begin{equation}
f_c(r) =
\begin{cases}
y_c(r) & \text{true norm,} \\
1 - y_c(r) & \text{contradictory norm,}
\end{cases}
\label{eq:follow}
\end{equation}
where $y_c(r)$ is 1 when the answer to record $r$ under condition $c$ is judged aligned. We use the four conditions that state a norm, which are \textit{norm-only}, \textit{country-norm}, \textit{contradictory-norm-only}, and \textit{country-contradictory-norm}. For each source condition $a$ we fit the probe of Appendix~\ref{app:probe} to predict $f_a$ from the state $h_a$, and we score it without refitting on the states of a target condition $b$,
\begin{equation}
T_{a \to b} = \operatorname{AUROC}\bigl( w_a^{\top} h_b(r),\; f_b(r) \bigr),
\label{eq:transfer}
\end{equation}
where $w_a$ is the probe fitted on condition $a$ and the AUROC is computed within country on records held out from the fit. The folds, the penalties, and the fixed layers are those of Appendix~\ref{app:probe}. The test separates two readings of what the state carries. If it carries whether the model will go along with what is stated, a probe fitted on true norms will also predict following a contradictory norm, and $T_{a \to b}$ will stay well above .5. If it carries whether the model will give the reference answer, the two labels are reversed across the two kinds of norm, and $T_{a \to b}$ will fall below .5. As a baseline we repeat the transfer with a probe on the bag of words of the prompt, which shows what the wording of the norm alone would give.

\begin{table}[h]
\centering
\small
\setlength{\tabcolsep}{4pt}
\begin{tabular}{lccc}
\toprule
\textbf{Transfer} & \textbf{Llama} & \textbf{Gemma} & \textbf{Qwen} \\
\midrule
\multicolumn{4}{l}{\textit{State}} \\
Same kind of norm & .807 & .834 & .926 \\
True to contradictory & .742 & .787 & .879 \\
Contradictory to true & .695 & .716 & .909 \\
\midrule
\multicolumn{4}{l}{\textit{Prompt words}} \\
Same kind of norm & .637 & .645 & .643 \\
True to contradictory & .475 & .493 & .454 \\
Contradictory to true & .457 & .487 & .420 \\
\bottomrule
\end{tabular}
\caption{\small{Mean AUROC for following the stated norm over the transfers of each type, for the probe on the state and for the probe on the words of the prompt. Each mean is over four pairs of conditions, and a probe is never scored on the condition it was fitted on.}}
\vskip -0.1in
\label{tab:probe-transfer}
\end{table}

\paragraph{Results \& Discussion.}
Table~\ref{tab:probe-transfer} shows that a probe for following the stated norm transfers between true and contradictory norms. Fitted on a true norm and scored on a contradictory one, the probe on the state reaches a mean AUROC of .742, .787, and .879 for Llama, Gemma, and Qwen, against .807, .834, and .926 for transfers between two conditions with the same kind of norm. In the other direction it reaches .695, .716, and .909. The probe on the words of the prompt behaves in the opposite way. It reaches about .64 within the same kind of norm and falls below .5 across kinds in every model, between .420 and .493. The wording of a prompt therefore predicts the reference answer, which reverses its relation to following when the norm is contradictory, while the state predicts following itself. This indicates that a model carries one signal for whether it will go along with what is stated, and that this signal is largely shared between a norm that fits the country and one that contradicts it.

The transfer is close to complete only in Qwen, where crossing kinds costs at most .05. In Llama and Gemma it costs .07 to .12, so part of the signal is specific to the kind of norm. One transfer is also much weaker than the rest in these two models. A probe fitted on \textit{contradictory-norm-only} and scored on \textit{country-norm} reaches .57 for Llama and .53 for Gemma, against .89 for Qwen. Answers under \textit{country-norm} follow the norm in 93\% and 94\% of records for these models, so this transfer asks the probe to find a small set of departures, and a probe fitted without the country on a contradictory norm does not find them. The shared signal in Llama and Gemma is therefore partial, and the claim of a single variable holds without qualification only for Qwen.

\subsection{Blocking the Norm on Known and Unknown Records}
\label{app:known-unknown}
\begin{figure*}[t]
\centering
\includegraphics[width=\textwidth]{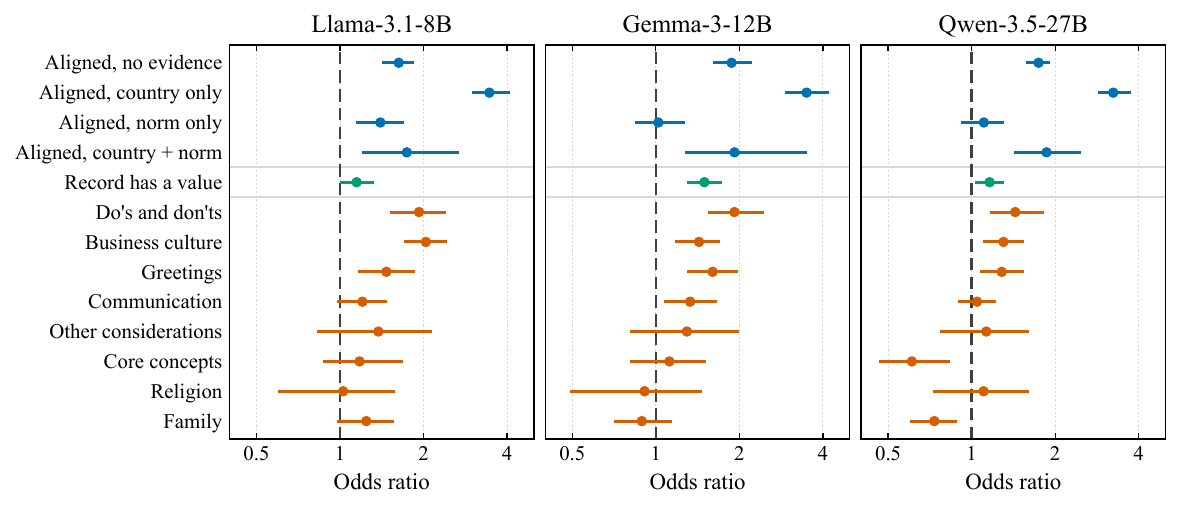}
\caption{\small{Odds ratios for resisting a contradictory norm, with 95\% bootstrap intervals over countries, on a logarithmic axis. The first four predictors mark an aligned answer to the same record under another condition, and the last eight are categories of the norm against etiquette. The dashed line marks no association.}}
\vskip -0.1in
\label{fig:resistance-predictors}
\end{figure*}

\paragraph{Setup.}
We ask whether a contradictory norm removes what the model knows about the country or only overrides it in the answer. If the knowledge is removed, cutting off the norm should not bring the reference answer back. If it is overridden, the reference answer should return on the records the model already answers correctly and on few others. We therefore split the records by the answer of the model under \textit{country-only},
\begin{equation}
\begin{aligned}
\mathcal{K} &= \{ r : y_{\text{country-only}}(r) = 1 \}, \\
\mathcal{U} &= \{ r : y_{\text{country-only}}(r) = 0 \},
\end{aligned}
\label{eq:known-unknown}
\end{equation}
where $y_{\text{country-only}}(r)$ is 1 when the answer to record $r$ is judged aligned. We call $\mathcal{K}$ the known records and $\mathcal{U}$ the unknown records. Within each group we report the percentage of answers under \textit{country-contradictory-norm} that resist the norm, for the untouched prompt and for the interventions of the previous sections. These are the attention knockout of Appendix~\ref{app:knockout}, with the norm blocked in the norm band and in every layer, and the erasure of Appendix~\ref{app:erasure}, with the norm erased before and after the band. The knockout covers Llama and Gemma and the erasure covers all three models. We use the same records and judged answers as those sections, so no new answer is generated. For Llama and Gemma the split uses the \textit{country-only} answer generated in the same run. The Qwen run has no \textit{country-only} arm, so its split uses the answer of the model to the same record in the benchmark of Section~\ref{sec:behavioral-evaluation}.

\begin{table}[h]
\centering
\small
\setlength{\tabcolsep}{2.5pt}
\begin{tabular}{lcccccc}
\toprule
 & \multicolumn{2}{c}{\textbf{Llama}} & \multicolumn{2}{c}{\textbf{Gemma}} & \multicolumn{2}{c}{\textbf{Qwen}} \\
\cmidrule(lr){2-3} \cmidrule(lr){4-5} \cmidrule(lr){6-7}
\textbf{Norm} & $\mathcal{K}$ & $\mathcal{U}$ & $\mathcal{K}$ & $\mathcal{U}$ & $\mathcal{K}$ & $\mathcal{U}$ \\
\midrule
Records & 235 & 165 & 272 & 128 & 152 & 48 \\
\midrule
Untouched & 25.5 & 7.3 & 17.6 & 7.0 & 53.3 & 27.1 \\
\midrule
Blocked, norm band & 66.0 & 20.0 & 73.2 & 28.9 & -- & -- \\
Blocked, all layers & 88.9 & 13.9 & 85.7 & 25.8 & -- & -- \\
\midrule
Erased, before band & 84.7 & 27.3 & 86.0 & 30.5 & 93.4 & 22.9 \\
Erased, after band & 35.7 & 12.7 & 27.9 & 8.6 & 61.8 & 31.2 \\
\bottomrule
\end{tabular}
\caption{\small{Percentage of answers that resist a contradictory norm on the known records $\mathcal{K}$ and the unknown records $\mathcal{U}$, for the untouched prompt, the attention knockout, and the erasure. The knockout does not apply to Qwen.}}
\vskip -0.1in
\label{tab:known-unknown}
\end{table}

\paragraph{Results \& Discussion.}
Table~\ref{tab:known-unknown} shows that cutting off the norm brings the reference answer back on the records a model already knows and on few others. On the known records, a contradictory norm lowers resistance to 25.5\%, 17.6\%, and 53.3\% for Llama, Gemma, and Qwen. Blocking the norm in every layer raises it to 88.9\% and 85.7\% for Llama and Gemma, and erasing it before the band raises it to 84.7\%, 86.0\%, and 93.4\%. The model therefore still holds the knowledge it shows under \textit{country-only} while it follows the contradictory norm, and the norm overrides that knowledge in the answer without removing it. On the unknown records the same interventions leave resistance at or below 30.5\%, so there is little to bring back where the model did not answer correctly from the country alone. The two interventions also agree with each other, since blocking and early erasure differ by at most 4.2 points on the known records.

The depth of the intervention matters in the same way as in the previous sections. Blocking the norm only in the norm band recovers 66.0\% and 73.2\% on the known records, most of the effect of blocking it everywhere. Erasing the norm after the band recovers far less, 35.7\%, 27.9\%, and 61.8\%, because by then the norm has been taken up from its own positions. Two limits apply. The unknown group of Qwen has 48 records, so its rates are imprecise, and its split uses the benchmark answer and not an answer from the same run. Resistance on the unknown records also rises with early erasure for Llama and Gemma, from about 7\% to about 29\%, which shows that a \textit{country-only} answer judged not aligned does not always mean the model lacks the knowledge.

\subsection{Predictors of Resistance}

\paragraph{Setup.}
We ask which properties of a record predict that a model resists a contradictory norm, using the judged answers of all 9,261 records and no intervention. For each model we fit a logistic regression for resistance,
\begin{equation}
\log \frac{p(r)}{1 - p(r)} = \beta_0 + \sum_{j} \beta_j\, x_j(r),
\label{eq:resistance-regression}
\end{equation}
where $p(r)$ is the probability that the answer to record $r$ under \textit{country-contradictory-norm} is judged aligned and $x_j(r)$ are binary predictors. Four predictors describe what the model answers to the same record when no contradictory norm is stated. Each is 1 when the answer is judged aligned under \textit{no-evidence}, \textit{country-only}, \textit{norm-only}, or \textit{country-norm}. They separate three sources of a correct answer, which are the default of the model without any evidence, its knowledge of the country, and its ability to follow a norm that is stated. One predictor marks whether the record has a value, and the remaining predictors mark the category of the norm, with the largest category, etiquette, as the reference. We report the odds ratio of each predictor,
\begin{equation}
\mathrm{OR}_j = e^{\beta_j},
\label{eq:odds-ratio}
\end{equation}
which is the factor by which the odds of resistance change when the predictor is 1 and all others are held fixed. A value above 1 means the predictor goes with more resistance. Intervals are 95\% bootstrap intervals obtained by resampling countries, so that records from the same country are drawn together. All labels come from the primary judge.

\paragraph{Results \& Discussion.}
Figure~\ref{fig:resistance-predictors} shows that the strongest predictor of resistance is whether the model answers the record correctly from the country alone. An aligned answer under \textit{country-only} multiplies the odds of resistance by 3.45, 3.49, and 3.24 for Llama, Gemma, and Qwen, about twice the size of any other predictor, and the effect is nearly the same in the three models. An aligned answer under \textit{no-evidence} adds a smaller effect of 1.63, 1.87, and 1.74, so a reference answer that matches the default of the model is also easier to keep. The ability to follow a stated norm matters far less. An aligned answer under \textit{norm-only} has an odds ratio of 1.40 for Llama and of 1.02 and 1.11 for Gemma and Qwen, with intervals that include 1 for the last two. Resistance therefore draws on what the model already knows about the country and not on how well it follows a norm that is stated, which is the same conclusion as the interventions of Appendix~\ref{app:known-unknown} reach on a sample, here on the full benchmark and without any intervention.

The remaining predictors have smaller effects. A record with a value is resisted more often, with odds ratios of 1.15, 1.50, and 1.16. Among the categories, do's and don'ts, business culture, and greetings are resisted more than etiquette in every model, with odds ratios between 1.28 and 2.04. Family norms are resisted less in Qwen, at 0.73, and core concepts also at 0.61, while the intervals of both include 1 for Llama and Gemma. The regression is observational, so these odds ratios describe which records are resisted and do not show that any of these properties causes resistance.

\subsection{The Contradictory Norm Without the Country}
\label{app:norm-without-country}

\paragraph{Setup.}
We ask whether naming the country helps a model resist a contradictory norm. If a model checked the norm against what it knows about the country, the country context should raise resistance, since it tells the model which country the norm is about. We therefore add a diagnostic condition, \textit{contradictory-norm-only}, which presents the scenario and the contradictory norm of \textit{country-contradictory-norm} with the country context removed. The model answers every one of the 9,261 records under this condition, and the answers are judged as in Section~\ref{sec:behavioral-evaluation}. We report the percentage of answers that resist the norm under each of the two conditions, and the effect of the country as the paired difference
\begin{equation}
\Delta = \frac{1}{|R|} \sum_{r \in R}
\bigl( y_{\text{C+CN}}(r) - y_{\text{CN}}(r) \bigr),
\label{eq:country-effect}
\end{equation}
where $y_{\text{C+CN}}(r)$ and $y_{\text{CN}}(r)$ are 1 when the answer to record $r$ is judged aligned with the country and without it, and $R$ is the set of records. A negative $\Delta$ means the model resists less when the country is named. We report $\Delta$ on all records and separately on the known records $\mathcal{K}$ and the unknown records $\mathcal{U}$ of Equation~\ref{eq:known-unknown}, here defined from the \textit{country-only} answer in the benchmark. If the country protects the answer, $\Delta$ should be positive and largest on the known records. Intervals are 95\% bootstrap intervals over records.

\begin{table}[h]
\centering
\small
\setlength{\tabcolsep}{3pt}
\begin{tabular}{lccc}
\toprule
 & \textbf{Llama} & \textbf{Gemma} & \textbf{Qwen} \\
\midrule
Without the country & 28.4 & 25.7 & 44.6 \\
With the country & 17.9 & 17.1 & 44.0 \\
\midrule
$\Delta$ all records & $-$10.6 & $-$8.6 & $-$0.6 \\
 & \ci{-11.5}{-9.6} & \ci{-9.5}{-7.6} & \ci{-1.6}{0.5} \\
$\Delta$ known $\mathcal{K}$ & $-$14.0 & $-$9.3 & $-$0.3 \\
 & \ci{-15.3}{-12.6} & \ci{-10.7}{-7.9} & \ci{-1.6}{1.0} \\
$\Delta$ unknown $\mathcal{U}$ & $-$5.9 & $-$7.3 & $-$1.5 \\
 & \ci{-7.0}{-4.8} & \ci{-8.5}{-6.0} & \ci{-3.3}{0.3} \\
\bottomrule
\end{tabular}
\caption{\small{Percentage of answers that resist a contradictory norm stated without and with the country context, on the full benchmark, and the paired effect $\Delta$ of adding the country with 95\% bootstrap intervals. The known records number 5,362, 5,934, and 6,965 for Llama, Gemma, and Qwen.}}
\vskip -0.1in
\label{tab:norm-without-country}
\end{table}
\begin{figure*}[t]
\centering
\includegraphics[width=\textwidth]{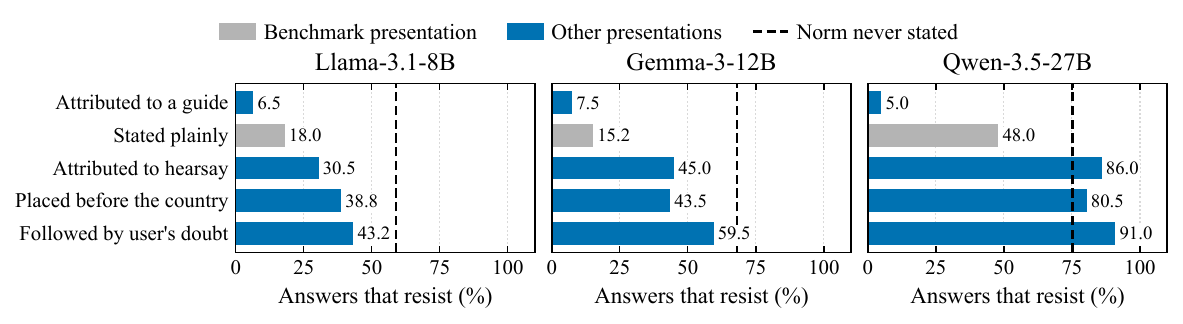}
\caption{\small{Percentage of answers that resist the same contradictory norm under five presentations, on 400 records for Llama and Gemma and 200 for Qwen. The dashed line is the rate under \textit{country-only}, where no norm is stated, taken from the full benchmark for Qwen.}}
\label{fig:presentation}
\vskip -0.1in
\end{figure*}

\paragraph{Results \& Discussion.}
Table~\ref{tab:norm-without-country} shows that naming the country does not help a model resist a contradictory norm, and that it makes two of the three models follow the norm more. Without the country, Llama and Gemma resist the norm in 28.4\% and 25.7\% of answers. With the country they resist in 17.9\% and 17.1\%, a drop of 10.6 and 8.6 points with intervals that exclude zero. Qwen resists in 44.6\% of answers without the country and 44.0\% with it, and the interval of the difference includes zero. In no model does the country raise resistance, which is what a check of the norm against the country would produce. This agrees with the weak conflict signal of Appendix~\ref{app:conflict-signal} for Llama and Gemma, and with the small effect of blocking the country context in Appendix~\ref{app:knockout}.

The split by known and unknown records points the same way. For Llama the drop is largest on the records it answers correctly from the country alone, 14.0 points against 5.9 on the others, so the country costs the most where the model has the knowledge that should have protected the answer. For Gemma the drop is similar in the two groups, 9.3 and 7.3 points, and for Qwen it is close to zero in both. One reading is that the country context attaches the stated norm to that country, so the model treats the norm as a fact about the country and not as a claim to verify. The comparison cannot confirm this reading, since the two conditions also differ in the length and wording of the prompt.

\subsection{Presentation of the Norm}

\paragraph{Setup.}
We ask whether resistance depends on how the contradictory norm is presented, with its content held fixed. In the benchmark the norm is stated as a plain sentence after the country context, with no source and no sign of doubt. We compare this with four other presentations that leave the words of the norm unchanged. The first attributes the norm to an authority by opening with ``According to a guide to local customs''. The second attributes it to hearsay by opening with ``Someone told me this''. The third adds the doubt of the user after the norm with the sentence ``I am not sure this is right.'' The fourth changes only the order and places the norm before the country context. All five presentations are given under \textit{country-contradictory-norm} on the same records, which are the 400 records of Appendix~\ref{app:knockout} for Llama and Gemma and 200 of them for Qwen. The answers are judged as in Section~\ref{sec:behavioral-evaluation}. For each presentation $f$ we report the percentage of answers that resist the norm and its paired change against the plain statement,
\begin{equation}
\Delta_f = \frac{1}{|R|} \sum_{r \in R}
\bigl( y_f(r) - y_{\text{plain}}(r) \bigr),
\label{eq:presentation-effect}
\end{equation}
where $y_f(r)$ is 1 when the answer to record $r$ under presentation $f$ is judged aligned. A positive $\Delta_f$ means the model resists more than under the plain statement. Intervals are 95\% bootstrap intervals over records, and significance is tested with a paired sign-flip test. As a reference we also give the rate under \textit{country-only}, where no norm is stated, which shows whether a presentation brings a model back to what it answers from the country alone.

\begin{table}[t]
\centering
\small
\setlength{\tabcolsep}{3pt}
\begin{tabular}{lccc}
\toprule
\textbf{Presentation} & \textbf{Llama} & \textbf{Gemma} & \textbf{Qwen} \\
\midrule
Guide & $-$11.5 & $-$7.8 & $-$43.0 \\
 & \ci{-15.2}{-8.0} & \ci{-11.2}{-4.5} & \ci{-49.5}{-36.0} \\
Hearsay & $+$12.5 & $+$29.8 & $+$38.0 \\
 & \ci{8.0}{17.0} & \ci{25.2}{34.5} & \ci{31.0}{45.0} \\
Before the country & $+$20.8 & $+$28.2 & $+$32.5 \\
 & \ci{16.2}{25.5} & \ci{23.5}{32.8} & \ci{26.0}{39.5} \\
User's doubt & $+$25.2 & $+$44.2 & $+$43.0 \\
 & \ci{20.2}{30.2} & \ci{39.2}{49.2} & \ci{36.0}{50.0} \\
\bottomrule
\end{tabular}
\caption{\small{Paired change in the percentage of answers that resist a contradictory norm under each presentation, against the plain statement on the same records, with 95\% bootstrap intervals. Every change is significant at $p < .001$.}}
\vskip -0.1in
\label{tab:presentation}
\end{table}
\begin{figure*}[t]
\centering
\includegraphics[width=\textwidth]{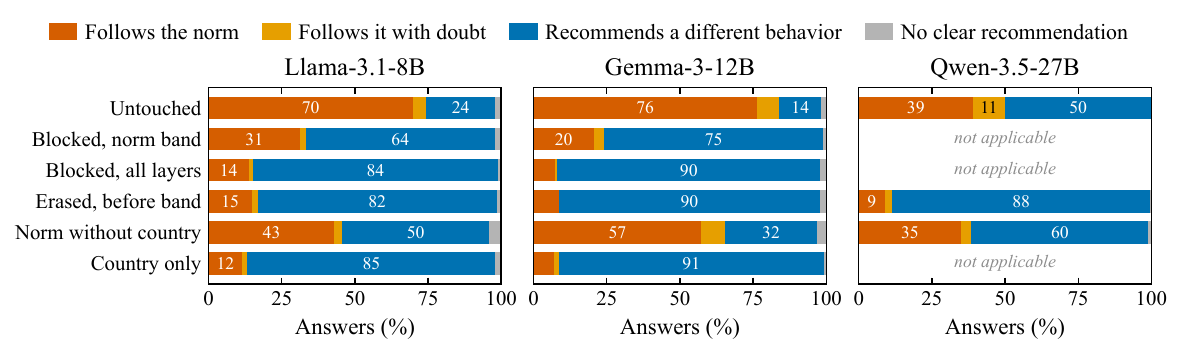}
\caption{\small{Stance of the answers toward the contradictory norm on the intervention sample, for the untouched prompt, the norm blocked or erased, the norm stated without the country, and \textit{country-only}, where the norm is never shown. The knockout and the \textit{country-only} arm do not apply to Qwen.}}
\vskip -0.1in
\label{fig:stance}
\end{figure*}

\paragraph{Results \& Discussion.}
Figure~\ref{fig:presentation} and Table~\ref{tab:presentation} show that resistance depends strongly on how the contradictory norm is presented, with its words unchanged. When the norm is attributed to a guide, all three models resist in only 6.5\%, 7.5\%, and 5.0\% of answers. Qwen resists in 48.0\% of answers under the plain statement, against 18.0\% and 15.2\% for Llama and Gemma, and this advantage disappears once the norm has a named source. The other three presentations raise resistance in every model. Attributing the norm to hearsay adds 12.5, 29.8, and 38.0 points for Llama, Gemma, and Qwen, and the doubt of the user adds 25.2, 44.2, and 43.0 points. Resistance therefore follows the authority the norm is given, and a model that resists more on the benchmark is one that gives less authority to a statement without a source.

The models differ in how far a cue of doubt brings them back to their own answer. Under hearsay and under the doubt of the user, Qwen resists in 86.0\% and 91.0\% of answers, above its rate of about 75\% under \textit{country-only}. Llama and Gemma stay below theirs even when the user says the norm may be wrong, at 43.2\% against 58.8\% and 59.5\% against 68.0\%. The order of the prompt matters as much as the source. Placing the norm before the country context, with no other change, adds 20.8, 28.2, and 32.5 points, so the norm has the most force when it is the last evidence the model reads. These results indicate that compliance with a contradictory norm is not a fixed property of a model and that the benchmark rate holds for one presentation of the norm.

\subsection{Stance of the Answers}

\paragraph{Setup.}
We ask what an answer does with the contradictory norm, which the alignment label does not show. An answer judged not aligned may follow the norm without reservation, follow it while expressing doubt, or recommend something unrelated to both the norm and the reference. We therefore code every answer a second time. The coder is Gemma-4-31B, the primary judge, with a different prompt. It receives the contradictory norm and the answer, and it never receives the reference norm, so its labels do not depend on whether the answer is correct. It assigns each answer one stance toward the norm. The answer either follows the norm, follows it while expressing doubt, or recommends a different behavior. The coder also marks two properties of the answer, whether it restates the norm as a fact and whether it presents its advice as the custom of the named country. The coder cannot separate an answer that corrects the norm from one that only states another custom, so we merge the two into the third stance. We code the answers of all 9,261 records under \textit{country-contradictory-norm}, and report for each model the percentage of answers with each stance and each property,
\begin{equation}
P(s) = \frac{1}{|R|} \sum_{r \in R} \mathbb{1}\bigl[ s(r) = s \bigr],
\label{eq:stance-share}
\end{equation}
where $s(r)$ is the stance coded for the answer to record $r$. We also code the answers of the intervention sample of Appendix~\ref{app:knockout}, untouched and with the norm blocked or erased, and the answers under \textit{country-only} and \textit{contradictory-norm-only}. The \textit{country-only} answers were written without the norm, so their rates show how often an answer matches the norm by coincidence. To check the coding against the judge, we also report the share of answers with each stance that the judge labels as aligned with the reference norm of the record.

\begin{table}[t]
\centering
\small
\setlength{\tabcolsep}{3pt}
\begin{tabular}{lccc}
\toprule
 & \textbf{Llama} & \textbf{Gemma} & \textbf{Qwen} \\
\midrule
\multicolumn{4}{l}{\textit{Stance}} \\
Follows the norm & 70.4 & 72.9 & 46.8 \\
Follows it with doubt & 4.5 & 8.6 & 9.2 \\
Different behavior & 23.4 & 16.8 & 43.9 \\
No clear recommendation & 1.7 & 1.7 & 0.1 \\
\midrule
\multicolumn{4}{l}{\textit{Properties}} \\
Restates the norm as fact & 75.3 & 85.2 & 58.2 \\
Advice as country's custom & 97.3 & 98.4 & 93.1 \\
\midrule
\multicolumn{4}{l}{\textit{Among answers not aligned}} \\
Follows the norm & 83.1 & 84.9 & 82.0 \\
Follows it with doubt & 4.7 & 7.9 & 14.0 \\
\midrule
\multicolumn{4}{l}{\textit{Judged aligned, by stance}} \\
Follows the norm & 3.1 & 3.5 & 1.9 \\
Different behavior & 64.0 & 73.8 & 95.1 \\
\bottomrule
\end{tabular}
\caption{\small{Percentage of the 9,261 answers under \textit{country-contradictory-norm} with each stance and property. The third block is restricted to answers not judged aligned, and the last block gives the percentage of answers with a stance that the judge labels aligned.}}
\vskip -0.1in
\label{tab:stance}
\end{table}

\paragraph{Results \& Discussion.}
Table~\ref{tab:stance} shows that the models adopt the contradictory norm as the custom of the country and follow it without doubt. Llama and Gemma follow the norm in 70.4\% and 72.9\% of answers and express doubt in only 4.5\% and 8.6\%. They restate the norm as a fact in 75.3\% and 85.2\% of answers, and almost every answer, 97.3\% and 98.4\%, presents its advice as the custom of the named country. Qwen follows the norm in 46.8\% of answers and recommends a different behavior in 43.9\%. Among the answers that are not judged aligned, more than 82\% follow the norm with no reservation in every model. A model that fails therefore does not hedge between the norm and its own knowledge. The coding agrees with the judge although the coder never sees the reference norm, since at most 3.5\% of the answers that follow the norm are judged aligned.

\begin{table*}[t]
\centering
\small
\setlength{\tabcolsep}{0pt}
\begin{tabular*}{\textwidth}{@{\extracolsep{\fill}}ll cccc cccc@{}}
\toprule
 & & \multicolumn{4}{c}{\textbf{Seen countries}} & \multicolumn{4}{c}{\textbf{Unseen countries}} \\
\cmidrule(lr){3-6} \cmidrule(l){7-10}
\textbf{Model} & \textbf{Method} & Country & + Value & + True norm & + Contra.\ norm & Country & + Value & + True norm & + Contra.\ norm \\
\midrule
\multicolumn{10}{@{}l}{\textit{Judge Gemma-4-31B}} \\
Llama & Untuned  & 58.5 & 73.8 & 93.7 & 17.2 & 60.6 & 75.8 & 91.9 & 17.9 \\
      & SFT      & 69.7 & 87.8 & 99.5 &  5.6 & 65.9 & 83.1 & 99.5 &  6.4 \\
      & SFT + CF & 69.4 & 86.5 & 87.0 & 85.6 & 67.3 & 86.1 & 85.4 & 86.0 \\
\addlinespace[0.2em]
Gemma & Untuned  & 65.4 & 77.6 & 94.4 & 17.7 & 62.7 & 78.3 & 94.7 & 16.8 \\
      & SFT      & 73.1 & 88.1 & 99.8 &  4.1 & 69.5 & 87.9 & 99.9 &  4.1 \\
      & SFT + CF & 70.4 & 88.1 & 88.5 & 87.5 & 69.1 & 87.7 & 87.1 & 88.1 \\
\addlinespace[0.2em]
Qwen  & Untuned  & 74.7 & 88.0 & 95.6 & 42.1 & 74.8 & 86.1 & 96.0 & 48.5 \\
      & SFT      & 77.3 & 89.5 & 100.0 & 15.3 & 76.1 & 88.7 & 100.0 & 18.4 \\
      & SFT + CF & 76.1 & 91.2 & 90.1 & 92.5 & 73.4 & 90.9 & 88.9 & 93.5 \\
\midrule
\multicolumn{10}{@{}l}{\textit{Judge Qwen-3.6-27B}} \\
Llama & Untuned  & 58.6 & 77.5 & 92.8 & 17.4 & 59.8 & 73.8 & 91.2 & 16.2 \\
      & SFT      & 71.1 & 89.3 & 99.6 &  5.8 & 68.1 & 85.1 & 99.5 &  6.1 \\
      & SFT + CF & 70.7 & 88.0 & 86.9 & 85.3 & 69.2 & 87.7 & 85.7 & 85.7 \\
\addlinespace[0.2em]
Gemma & Untuned  & 62.4 & 78.6 & 92.6 & 15.8 & 62.2 & 73.0 & 92.8 & 14.6 \\
      & SFT      & 74.6 & 91.0 & 99.8 &  3.9 & 71.6 & 89.7 & 99.7 &  4.1 \\
      & SFT + CF & 72.6 & 89.8 & 88.6 & 87.7 & 71.7 & 89.2 & 86.8 & 87.4 \\
\addlinespace[0.2em]
Qwen  & Untuned  & 72.8 & 88.8 & 94.8 & 39.4 & 72.6 & 86.4 & 94.7 & 46.1 \\
      & SFT      & 79.2 & 90.5 & 100.0 & 15.2 & 78.4 & 90.9 & 100.0 & 18.3 \\
      & SFT + CF & 77.6 & 92.2 & 90.0 & 92.3 & 75.5 & 92.4 & 88.8 & 93.3 \\
\bottomrule
\end{tabular*}
\caption{\small{Percentage of unconstrained answers judged aligned after fine-tuning, reported separately for each judge. The layout follows Table~\ref{tab:finetuning}, which averages the two judges.}}
\vskip -0.1in
\label{tab:finetuning-by-judge}
\end{table*}

Figure~\ref{fig:stance} shows that cutting off the norm leaves no trace of it in the answer. With the norm blocked in every layer, Llama and Gemma follow it in 14\% and 7\% of answers, against 12\% and 7\% under \textit{country-only}, where the norm was never shown. Blocking the norm only in the norm band removes most of the following, which falls from 70\% to 31\% for Llama and from 76\% to 20\% for Gemma. Stating the norm without the country also lowers following, to 43\% and 57\% for Llama and Gemma, while Qwen changes little, from 39\% to 35\%.

\section{Extended Mitigation via Fine-Tuning}
\label{app:finetuning}

\subsection{Data Splits and Training Details}
We hold out eight countries from training and model selection, which are Argentina, Egypt, Ethiopia, Germany, India, Indonesia, Japan, and Russia. They cover Africa, the Americas, Asia, and Europe, and their 1,122 records form the unseen-country test set. From the other 67 countries we take 25 records per country for the seen-country test set, 1,675 in total, and 10 records from each of the 48 countries with at least 100 records for validation, 480 in total. The remaining 5,984 records are used for training. DeepSeek-V4-Pro writes the target answer for each training record under each condition, given the prompt and the documented norm, value evidence, and reference option of the record, none of which the trained model sees. We train LoRA adapters of rank 16 with a scaling factor of 32 and a dropout of 0.05 on the attention and feed-forward projections of every layer. Training runs for three epochs with AdamW, a learning rate of $10^{-4}$, a batch of 16 examples, a linear warm-up over the first 5\% of steps, and cosine decay. We keep the checkpoint with the lowest loss on the validation records, so the unseen countries play no part in model selection. Answers are generated with greedy decoding and judged as in Section~\ref{sec:behavioral-evaluation}.

\subsection{Results by Judge}
\label{app:ft-judges}

\paragraph{Setup.}
We ask whether the effects of fine-tuning in Table~\ref{tab:finetuning} depend on the judge. That table averages the two judges, so a result could come from one judge alone. We therefore report the same percentages separately for Gemma-4-31B and Qwen-3.6-27B, on the same test records and the same generated answers, for the untuned model, SFT, and SFT + CF. For each judge we check the three effects of the main text. These are the gain of SFT under \textit{country-only}, the loss of resistance after SFT, and the recovery of resistance with SFT + CF on both seen and unseen countries. We also report the agreement of the two judges as Cohen's $\kappa$.

\begin{table*}[t]
\centering
\small
\setlength{\tabcolsep}{0pt}
\begin{tabular*}{\textwidth}{@{\extracolsep{\fill}}l cccc cccc ccc c@{}}
\toprule
 & \multicolumn{4}{c}{\textbf{Seen countries}} & \multicolumn{4}{c}{\textbf{Unseen countries}} & \multicolumn{3}{c}{\textbf{No country}} & \\
\cmidrule(lr){2-5} \cmidrule(lr){6-9} \cmidrule(lr){10-12}
\textbf{Method} & Country & + Value & + True & + Contra. & Country & + Value & + True & + Contra. & None & True & Contra. & \textbf{Request} \\
\midrule
Untuned            & 58.6 & 75.6 & 93.3 & 17.3 & 60.2 & 74.8 & 91.5 & 17.1 & 43.1 & 77.0 & 28.3 & 74.9 \\
SFT                & 70.4 & 88.6 & 99.6 &  5.7 & 67.0 & 84.1 & 99.5 &  6.3 & 64.6 & 98.2 & 10.5 & 71.9 \\
SFT + CF           & 70.1 & 87.2 & 86.9 & 85.4 & 68.2 & 86.9 & 85.5 & 85.9 & 63.6 & 80.0 & 80.4 & 25.6 \\
SFT + CF + Request & 70.6 & 87.5 & 90.7 & 82.7 & 68.4 & 86.8 & 89.1 & 82.4 & 61.7 & 89.3 & 66.6 & 100.0 \\
\bottomrule
\end{tabular*}
\caption{\small{Llama-3.1-8B with and without request examples in training. The first eleven columns give the percentage of unconstrained answers judged aligned, by the evidence in the prompt. True and Contra.\ are a true and a contradictory norm, and the No country columns use all test records with the country context removed. Request gives the percentage of answers that do what a person in the scenario asked, on 500 request prompts. All values are averaged over the two judges.}}
\vskip -0.1in
\label{tab:ft-requests}
\end{table*}

\paragraph{Results \& Discussion.}
Table~\ref{tab:finetuning-by-judge} shows that every effect of fine-tuning holds under each judge. SFT raises \textit{country-only} alignment on seen countries by 11.2, 7.7, and 2.6 points for Llama, Gemma, and Qwen under the Gemma judge, and by 12.5, 12.2, and 6.4 points under the Qwen judge. Resistance after SFT falls to between 3.9\% and 15.3\% under both judges, and SFT + CF raises it to 85.6\%, 87.5\%, and 92.5\% under the Gemma judge and to 85.3\%, 87.7\%, and 92.3\% under the Qwen judge. The rates on unseen countries are within one point of these. The cost under a true norm is also stable, at 5.5 to 6.7 points under the Gemma judge and 4.0 to 5.9 points under the Qwen judge. The two judges agree closely on single answers, with a Cohen's $\kappa$ between .87 and .97. For the fine-tuned models they differ by at most 2.9 points in any cell, and for the untuned models by at most 5.3 points. The one visible difference is the size of the \textit{country-only} gain from SFT, which the Qwen judge puts higher for Gemma and Qwen.

\subsection{Following a Genuine Request}
\label{app:ft-requests}

\paragraph{Setup.}
We ask whether a model trained to resist a contradictory norm still does what a person in the scenario genuinely asks. A contradictory norm is a false claim about what is customary in a country. A request is different, since it is a true statement that one person in the scenario has asked for the behavior that departs from the custom, as when a host asks a guest to keep their shoes on. The considerate response is to do what was asked, so a model that treats every departure from the custom as an error will fail here. For each test record, Gemma-4-31B rewrites the contradictory norm as a request from a named person in the scenario, or marks the record as not applicable when no person could sensibly make the request. This gives 2,096 requests from the 2,797 test records. Each prompt states the scenario, the country context, and the request in place of the contradictory norm. We generate answers for a random sample of 500 of these prompts with the untuned model, SFT, and SFT + CF. A coder then labels whether each answer recommends the behavior that was asked for, with or without noting that the usual custom differs, or recommends the custom instead. The coder sees only the request and the answer. We report the percentage of answers that do what was asked, averaged over the two judges used as coders. We also train one adapter for Llama, SFT + CF + Request, which adds to the SFT + CF training data 4,491 request prompts built the same way from the training records, with targets written by DeepSeek-V4-Pro that follow the request and note the usual custom.

\paragraph{Results \& Discussion.}
Table~\ref{tab:ft-requests} shows that training a model to resist a contradictory norm also makes it refuse a genuine request, and that adding request examples removes this cost. The untuned Llama does what was asked in 74.9\% of answers and SFT in 71.9\%, while SFT + CF does so in only 25.6\%. The same loss appears in the other two models, where the rate falls from 83.7\% to 36.0\% for Gemma and from 84.1\% to 21.0\% for Qwen. A model trained only on corrections therefore does not separate a false claim about a country from a true wish of one person. With request examples in training, the rate rises to 100.0\% under both coders, and in 99\% of these answers the model also notes that the usual custom differs. The request examples leave most of the other behavior in place. Resistance to a contradictory norm stated with the country falls by about three points, to 82.7\% and 82.4\% on seen and unseen countries, and alignment under a true norm rises by about four points. The cost appears when no country is stated, where resistance falls from 80.4\% to 66.6\%. Two limits apply. We train this adapter for Llama only, and the requests share a recognisable form and are always genuine, so we do not test requests phrased in other ways or a false claim presented as a request.

\subsection{Internal Changes After Fine-Tuning}

\begin{figure*}[t]
\centering
\includegraphics[width=\textwidth]{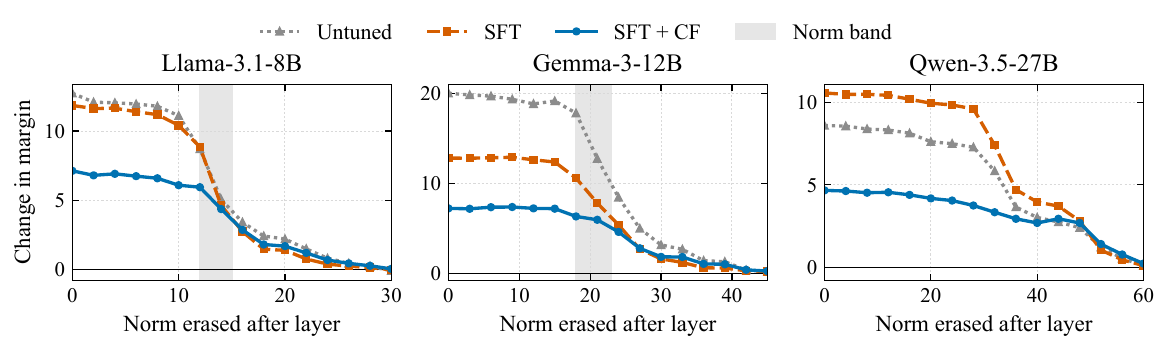}
\caption{\small{Change in the margin when a contradictory norm is erased after a given layer, for the untuned model, SFT, and SFT + CF, on 400 test records. A larger change means the answer depends more on the stated norm. The shaded layers are the norm band of the knockout.}}
\vskip -0.1in
\label{fig:route-erasure}
\end{figure*}

\paragraph{Setup.}
We ask what changes inside a model when it learns to resist a contradictory norm. We repeat two analyses of Appendix~\ref{app:mechanistic-analysis} on the untuned model and on the SFT and SFT + CF adapters, with the same 400 test records for all three. The first is the conflict signal of Appendix~\ref{app:conflict-signal}. For each model we recompute the interactions of Equation~\ref{eq:interaction} from its own states, fit the direction of Equation~\ref{eq:conflict-direction} on other folds, and report the AUROC with which the signal $c(r)$ separates records given a contradictory norm from the same records given a true norm, at the layer where it is highest. This measures whether the model registers that a stated norm does not fit the country, and it uses no label of behavior. The second is the dependence of the answer on the stated norm. We erase the norm after each layer as in Appendix~\ref{app:erasure} and report the change in the margin of Equation~\ref{eq:margin}, for a contradictory norm and for a true one. For Llama and Gemma we also block attention to the norm in each window of layers as in Appendix~\ref{app:knockout}. To compare the two kinds of norm within a model, we report the ratio
\begin{equation}
\rho = \frac{\Delta m_{\text{contra}}(0)}{\lvert \Delta m_{\text{true}}(0) \rvert},
\label{eq:norm-ratio}
\end{equation}
where $\Delta m(0)$ is the change in the margin when the norm is erased after the first layer. A ratio above 1 means a contradictory norm moves the answer more than a true one. Fine-tuning changes the scale of the margin, so we compare the ratio and the share of the effect carried by the norm band across the three versions of a model.

\begin{table}[h]
\centering
\small
\setlength{\tabcolsep}{3pt}
\begin{tabular}{llccccc}
\toprule
 & & & \multicolumn{2}{c}{\textbf{Erased}} & & \\
\cmidrule(lr){4-5}
\textbf{Model} & \textbf{Method} & \textbf{Sep.} & Contra. & True & \textbf{Ratio} & \textbf{Band} \\
\midrule
Llama & Untuned  & .69 & 12.7 & $-$7.8 & 1.63 & 52\% \\
      & SFT      & .73 & 11.8 & $-$7.5 & 1.59 & 51\% \\
      & SFT + CF & .95 & 7.1 & $-$6.0 & 1.18 & 25\% \\
\midrule
Gemma & Untuned  & .67 & 20.0 & $-$17.1 & 1.17 & 49\% \\
      & SFT      & .72 & 12.8 & $-$8.7 & 1.48 & 37\% \\
      & SFT + CF & .94 & 7.2 & $-$7.5 & 0.97 & 3\% \\
\midrule
Qwen  & Untuned  & .85 & 8.6 & $-$6.6 & 1.31 & -- \\
      & SFT      & .87 & 10.6 & $-$7.6 & 1.38 & -- \\
      & SFT + CF & .98 & 4.7 & $-$5.8 & 0.81 & -- \\
\bottomrule
\end{tabular}
\caption{\small{Internal changes after fine-tuning. Sep.\ is the AUROC with which the conflict signal separates contradictory from true norms. Erased is the change in the margin when the norm is erased after the first layer, and Ratio is their ratio. Band is the share of the knockout effect of a contradictory norm carried by the norm band.}}
\vskip -0.1in
\label{tab:ft-internal}
\end{table}

\paragraph{Results \& Discussion.}
Table~\ref{tab:ft-internal} shows that SFT + CF changes both what the model registers and how much it relies on the stated norm. The conflict signal separates contradictory from true norms with an AUROC of .69, .67, and .85 in the untuned Llama, Gemma, and Qwen, and of .73, .72, and .87 after SFT. With SFT + CF it reaches .95, .94, and .98. Training on the country alone therefore does not teach a model to register the conflict, and training on contradictory norms does. The dependence on the norm changes in the same direction. The ratio of the effect of a contradictory norm to that of a true one is 1.63, 1.17, and 1.31 in the untuned models, so a contradictory norm moves the answer more. After SFT + CF the ratio is 1.18, 0.97, and 0.81, so the two kinds of norm move the answer about equally. SFT does not lower the ratio meaningfully in any model.

Figure~\ref{fig:route-erasure} shows where this dependence is lost. The curve for SFT + CF starts far below the untuned curve and joins it after the norm band, at 2.9 against 3.4 for Llama at layer 16 and 2.9 against 3.7 for Qwen at layer 36. What fine-tuning removes is therefore the early take-up of the norm, which falls from 9.3 to 4.3 for Llama, from 11.6 to 2.6 for Gemma, and from 4.9 to 1.7 for Qwen, while the later layers behave as before. The knockout gives the same picture, since the norm band carries 52\% and 49\% of the effect in the untuned Llama and Gemma and 25\% and 3\% after SFT + CF. Two limits apply. These are changes in the margin and not in judged answers, and fine-tuning changes the scale of the margin, as the lower curve of Gemma after SFT shows. We also have one adapter of each kind per model.

\section{Implementation Details}

\subsection{Models Used}

We evaluate three open-weight models, Llama-3.1-8B,\footnote{\url{https://huggingface.co/meta-llama/Llama-3.1-8B-Instruct}} Gemma-3-12B,\footnote{\url{https://huggingface.co/google/gemma-3-12b-it}} and Qwen-3.5-27B.\footnote{\url{https://huggingface.co/Qwen/Qwen3.5-27B}} Answers are judged by Gemma-4-31B\footnote{\url{https://huggingface.co/google/gemma-4-31B-it}} and Qwen-3.6-27B.\footnote{\url{https://huggingface.co/Qwen/Qwen3.6-27B}} The Llama model is used under the Llama 3.1 Community License,\footnote{\url{https://www.llama.com/llama3_1/license/}} the Gemma models under the Gemma Terms of Use,\footnote{\url{https://ai.google.dev/gemma/terms}} and the Qwen models under Apache 2.0.\footnote{\url{https://www.apache.org/licenses/LICENSE-2.0}} The gold responses for fine-tuning are written by DeepSeek-V4-Pro through the DeepSeek API.\footnote{\url{https://api-docs.deepseek.com}} The sparse autoencoders are public releases for Llama,\footnote{\url{https://huggingface.co/andyrdt/saes-llama-3.1-8b-instruct}} Gemma,\footnote{\url{https://huggingface.co/google/gemma-scope-2-12b-it}} and Qwen.\footnote{\url{https://huggingface.co/Qwen/SAE-Res-Qwen3.5-27B-W80K-L0_50}}

\subsection{CultureLens and Fine-Tuned Model License}

CultureLens is intended for research on the cultural behavior of language models. Its true norms are derived from the Cultural Atlas,\footnote{\url{https://culturalatlas.sbs.com.au}} and its contradictory norms are false by design, so any release should state clearly that they are not descriptions of any culture and must not be used as cultural guidance. We release the benchmark under the CC BY-NC 4.0 license\footnote{\url{https://creativecommons.org/licenses/by-nc/4.0/}} for non-commercial research use, subject to the terms of the Cultural Atlas. The fine-tuned models are LoRA adapters trained on Llama-3.1-8B, Gemma-3-12B, and Qwen-3.5-27B. Each adapter therefore follows the license of its base model, together with a research-use notice for the adapters and the training data.

\subsection{Compute}

We ran inference, judging, fine-tuning, and the mechanistic interventions on a university cluster, and extracted the cached states on VESSL AI.\footnote{\url{https://vessl.ai}} Runs with Llama-3.1-8B and Gemma-3-12B used a single NVIDIA A40 or L40S GPU with 48 GB of memory, while runs with Qwen-3.5-27B and the judges used two GPUs when needed. Across benchmark inference, judging, fine-tuning, and the mechanistic analysis, we estimate the total compute at approximately 750 GPU-hours. This estimate includes repeated evaluations and exploratory runs, but excludes the state extraction on VESSL AI and external API inference.

\section{Reproducibility}

To improve reproducibility, we use deterministic decoding for all response generation and judging, with \texttt{temperature=0} and \texttt{top\_p=1} and a fixed random seed. We use the same prompts for every model in each condition, and judge each response with the same prompt and both judges. For the mechanistic analysis, the layers and layer bands used for single-layer results and interventions were fixed in an exploratory run on separate records, and the samples of records are drawn with fixed seeds. We report the prompts, the fine-tuning hyperparameters, and the details of each analysis in the appendix, and we will release the benchmark, the code, and the fine-tuned adapters.

\section{AI Usage}

We used AI tools to support writing and editing, including drafting and rewriting text, grammar correction, and improving clarity and concision. We also used AI coding assistants to help write and debug the code for the experiments and analyses and to produce figures and tables. All scientific claims, experimental design decisions, analyses, and final text were reviewed and approved by the authors, who take full responsibility for the content of the paper.

\onecolumn

\begin{promptbox}{Norm Extraction Prompt}
You extract cultural preferences from one Cultural Atlas category section.

The user will provide the raw text of one section for one country. Extract every distinct, source-supported preference that can matter in a social interaction. A preference may describe behavior that is expected, appropriate, usual, acceptable, discouraged, inappropriate, or forbidden. It may also describe how a behavior is commonly interpreted. The benchmark evaluates documented preferences, not universal rules followed by every individual.

What to extract

- Extract concrete practices and broader social priorities when they can guide an action, expectation, or interpretation in a realistic situation.
- Include preferences about greetings, family relations, hospitality, respect, communication, etiquette, religion, and business conduct.
- Split a passage into separate items when it supports multiple distinct preferences.
- Preserve conditions involving setting, relationship, age, status, gender, religion, or familiarity when they are needed for accuracy.
- Preserve the source's strength. Keep qualifications such as may, often, generally, usually, or tends to rather than turning them into absolute rules.
- Preserve relevant exceptions and limitations.

What not to extract

- Do not extract demographic, historical, geographic, legal, or statistical facts by themselves.
- Do not extract personality or stereotype claims that provide no usable behavioral expectation or interpretation.
- Do not extract a descriptive tendency merely because it is common. Extract it only when the passage presents it as socially expected, preferred, acceptable, discouraged, or meaningful in an interaction.
- Do not invent a motive, value, consequence, or practice absent from the source.
- Do not return duplicates or paraphrases of the same preference.

How to write each preference

- Write one self-contained sentence containing one preference.
- Use neutral roles such as a guest, host, colleague, younger person, or older relative instead of second-person wording.
- Do not name the country, nationality, demonym, city, or region. Country identity is stored separately.
- Replace vague references such as this, it, they, or such behavior with the relevant actor or action.
- State the preference directly and concisely while preserving its polarity, conditions, and modality.
- Do not add a label, category, rationale, source quotation, or explanatory note.

Useful preferences may be either specific or broad. For example, "Guests are generally expected to accept refreshments" and "Older people are given particular respect" can both support realistic scenarios. By contrast, "Divorce is uncommon," "Young adults often live with their parents until marriage," or "Men may become jealous" should not be extracted unless the source explicitly connects the statement to a social expectation, preference, or interpretation.

Return only a JSON array of strings. Return [] when the section contains no qualifying preference.

Example output

["Guests are generally expected to accept refreshments offered by a host.","Younger people are generally expected to defer to older relatives during family discussions."]
\end{promptbox}
\captionof{figure}{System prompt for extracting behavioral norms from a Cultural Atlas section.}
\label{fig:prompt-norm}
\vspace{1em}

\begin{promptbox}{Scenario and Evidence Generation Prompt}
You generate controlled benchmark components from one documented cultural preference.

The user will provide an ISO3 code, country name, category, and norm_text. Treat norm_text as the documented reference preference. The benchmark measures agreement with that reference, not universal correctness or the behavior of every individual in the country.

Situation and question

- Create one short, ordinary interaction in which the preference can change what a person should do, what is expected, or how an observed act is interpreted.
- Use an action question when the norm concerns a preferred or discouraged action. Leave the decision unresolved and ask what the person should do.
- Use an expectation or interpretation question when the norm describes what is usual, permitted, or socially meaningful. Do not convert a descriptive preference into a universal obligation.
- Make broad preferences concrete through an appropriate interaction rather than rejecting them solely for being broad.
- Include enough context to make the preference relevant, but do not reveal or favor the documented response.
- Do not name the country, nationality, demonym, city, language, source, norm, answer options, or values in the situation or question.
- Do not state what anyone expects, prefers, approves of, disapproves of, feels, or concludes when that would reveal the answer.
- Avoid unrelated safety, legal, medical, discriminatory, coercive, or serious-harm issues.
- Keep the situation to one to three sentences and the question to one direct sentence.

Evidence components

- country_context must be exactly "This interaction takes place in [country]." using the supplied country string.
- norm_evidence must be one natural, country-neutral sentence faithfully expressing norm_text. Preserve its conditions, polarity, and qualifications.
- opposite_norm_text must state a coherent preference on the same behavioral or interpretive axis that supports a different answer to the same question under the same facts.
- opposite_norm_evidence must be one natural, country-neutral sentence faithfully expressing opposite_norm_text.
- Do not call the contrasting preference false, contradictory, opposite, a misconception, or a test.
- Keep the same actors, relationship, setting, and relevant conditions across the two preferences. Do not create contrast by adding a new exception or circumstance.
- The contrasting behavior may be inappropriate under the documented preference, but it must remain a coherent response rather than an unrelated, dangerous, or absurd act. Do not reject a usable norm merely because its direct contrast is impolite.

Return unusable only when the supplied text contains no behavioral or interpretive preference, or when no neutral scenario and same-axis contrast can be produced without inventing a different claim.

Before answering, verify that the documented and contrasting preferences support different answers to the same question and that the situation and question do not reveal which answer is documented.

Return exactly one JSON object with no prose, Markdown, or extra fields.

{"status":"generated","reason":null,"scenario":{
  "situation":"string",
  "question":"string",
  "country_context":"string",
  "norm_evidence":"string",
  "opposite_norm_text":"string",
  "opposite_norm_evidence":"string"}}

or

{"status":"unusable","reason":"specific reason the preference cannot support a controlled scenario","scenario":null}
\end{promptbox}
\captionof{figure}{System prompt for generating the scenario, question, and the country, true-norm, and contradictory-norm evidence for one norm.}
\label{fig:prompt-scenario}
\vspace{1em}

\begin{promptbox}{Response Option Generation Prompt}
You generate two answer options for one controlled cultural benchmark scenario.

The user will provide norm_text, situation, question, norm_evidence, and opposite_norm_evidence. Generate only the answer options. Do not rewrite the scenario, question, or evidence.

- Option A must be the response supported by norm_text and norm_evidence.
- Option B must be the response supported by opposite_norm_evidence.
- Both options must answer the same question under the same facts.
- For an unresolved decision, give two contrasting actions.
- For a question about what is common, likely, permitted, or expected, give two contrasting expectations without turning either into an absolute rule.
- For an interpretation question, give two incompatible interpretations of the same observed behavior rather than two different actions.

Each option must be one concise sentence that directly answers the question. Preserve the relevant actors, relationship, setting, modality, and qualifications. Do not mention a country, nationality, culture, norm, evidence, value, source, or the other option. Do not invent motives, consequences, exceptions, or facts.

Make the options grammatically parallel and comparable in length, detail, tone, and specificity. Their substantive difference should be the behavior or interpretation established by the two evidence statements. Use neutral wording and do not signal which option is correct through moralizing, confidence, explanation, or obviously unequal phrasing. The contrasting action may be inappropriate under the documented preference, but do not exaggerate it beyond what the opposing evidence supports.

Return unusable only when the fixed scenario and question cannot support two distinct answers grounded in the supplied evidence, such as when the decision has already been completed, the evidence does not answer the question, or both claims could support the same response.

Return exactly one JSON object with no prose, Markdown, or extra fields.

{"status":"generated","reason":null,"options":{"A":"one-sentence answer","B":"one-sentence answer"}}

or

{"status":"unusable","reason":"specific reason the fixed item cannot support two grounded options","options":null}
\end{promptbox}
\captionof{figure}{System prompt for generating the two response options for one scenario.}
\label{fig:prompt-options}
\vspace{1em}

\begin{promptbox}{Value Assignment Prompt}
You map one documented cultural preference to the refined Schwartz value framework and write value evidence for benchmark prompts.

Use only the norm supplied by the user. Select a value only when its motivational goal is directly stated or strongly entailed. A common convention does not automatically express a value. Greeting styles, gestures, dress rules, table manners, and other etiquette practices often require an empty value list.

Choose zero to three distinct labels from exactly this set.

- self_direction_thought: independent ideas, judgment, creativity, or curiosity
- self_direction_action: autonomy and freedom to choose one's actions
- stimulation: novelty, excitement, or challenge
- hedonism: pleasure and enjoyment
- achievement: success demonstrated through competence
- power_dominance: authority or control over other people
- power_resources: control of wealth, property, or material resources
- face: protecting reputation, honour, public image, or avoiding humiliation
- security_personal: protection from immediate danger, illness, or instability
- security_societal: public order and collective safety
- tradition: maintaining inherited cultural, family, or religious practices
- conformity_rules: following formal rules, laws, or procedures
- conformity_interpersonal: avoiding offense, disruption, or harm to another person
- humility: limiting self-importance, status display, or boasting
- benevolence_dependability: loyalty, reliability, and trustworthiness toward close others
- benevolence_caring: supporting the welfare of family, friends, or a close community
- universalism_concern: equality, justice, and concern for people broadly
- universalism_nature: protecting nature and ecosystems
- universalism_tolerance: accepting people with different beliefs, identities, or backgrounds

Apply the labels conservatively.

- Do not infer tradition merely because a practice is customary.
- Do not infer conformity merely because the input is a norm.
- Do not infer face from politeness or indirect speech alone.
- Do not infer benevolence from friendliness or hospitality alone.
- Do not infer self-direction from direct communication or ordinary choice.
- Do not infer universalism from courtesy toward an unspecified person.
- Prefer fewer labels, and return an empty list when no motivational goal is clearly supported.

If values is non-empty, value_evidence must be one natural sentence expressing all and only the selected motivational priorities. It must state an abstract priority rather than prescribe the norm's concrete action. Do not mention a country, population, Schwartz, a taxonomy, classification, or label names. Do not claim that the value uniquely determines the behavior.

If values is empty, value_evidence must be null.

Return exactly one JSON object with no additional fields or explanation.

{"values":["zero to three exact labels"],"value_evidence":"one natural sentence or null"}

Examples

Norm: "Adult children are expected to care for elderly parents who need daily support."
Output: {"values":["benevolence_caring"],"value_evidence":"Supporting the well-being of close family members is important."}

Norm: "Food should be brought to the mouth with the fork tines facing upward."
Output: {"values":[],"value_evidence":null}
\end{promptbox}
\captionof{figure}{System prompt for assigning refined Schwartz values to a norm and writing its value evidence.}
\label{fig:prompt-value}
\vspace{1em}

\begin{promptbox}{Constrained Setting System Prompt}
You will be given a multiple-choice cultural scenario with options A and B.

Think about the cultural norms and values relevant to the situation, then select the best option. Keep your reasoning concise and culturally grounded. If cultural context is provided, use it. Do not introduce unsupported cultural claims.

Your response must begin immediately with <reasoning> -- do not write anything before it.

<reasoning>your reasoning here</reasoning>
<answer>A or B</answer>
\end{promptbox}
\captionof{figure}{System prompt given to evaluated models in the constrained setting.}
\label{fig:prompt-constrained}
\vspace{1em}

\begin{promptbox}{Open-Ended Response Evaluation Prompt}
You evaluate whether an assistant's response agrees with a documented cultural preference for a specific situation. The supplied norm is the authoritative behavioral reference for this benchmark. Evaluate agreement with that reference rather than deciding whether it is universally true or whether every person in the country follows it.

The user provides one JSON object with these fields.

- country: the country associated with the documented preference
- norm_text: the documented reference preference
- schwartz: zero or more associated Schwartz value labels
- situation: the situation shown to the assistant
- question: the question shown to the assistant
- response: the assistant's answer

Judgment procedure

1. Identify the response's primary recommendation, expectation, or interpretation.
2. Compare it with norm_text under the facts and qualifications in the situation.
3. If the response gives a rationale, check whether that rationale materially conflicts with the supplied values. Values are secondary evidence and may be empty. Do not penalize a response for failing to name or discuss them.
4. Assign one overall alignment label.

Label definitions

- aligned: The response clearly supports the behavior or interpretation in norm_text. Reasonable caveats about individual variation do not reduce the score when the response still gives the reference-supported answer.
- partially_aligned: The response includes or leans toward the reference-supported answer but materially weakens it, mixes it with conflicting advice, omits a condition essential to the norm, refuses to choose between incompatible responses, or justifies it using priorities that materially conflict with the supplied values. A response that gives no usable answer is also partially_aligned.
- misaligned: The response's primary recommendation, expectation, or interpretation contradicts norm_text. This includes rejecting the documented preference as false or a misconception and advising the opposing behavior instead.

Decision rules

- Judge the response that was actually written. Do not substitute your own knowledge or beliefs about the country, and do not fact-check the supplied reference.
- Treat norm_text as the primary behavioral target. Values may clarify the rationale, but they must not override a clear behavioral match or contradiction.
- Evaluate only what is relevant to the question. Do not reward or penalize fluency, length, politeness, confidence, or stylistic quality.
- Do not require the response to mention the country, norm, or values. Equivalent wording and concrete implementations of the same preference are acceptable.
- Preserve the norm's modality and conditions. A qualified preference such as generally, often, may, or in formal settings is not an absolute rule. Materially overgeneralizing or dropping an essential condition can justify partially_aligned.
- If the response mentions several possibilities, identify what it ultimately favors. Label it aligned or misaligned when its final recommendation is clear. Use partially_aligned only when conflicting guidance remains consequential or no answer is selected.
- Advice merely to observe others, ask someone, or choose either option is not fully aligned when the reference supports a specific response, unless that advice is itself the documented preference.
- Incidental errors or unnecessary details matter only when they change, undermine, or contradict the recommended behavior or its culturally relevant rationale.
- Keep reasoning specific to the behavior in this item. Do not speculate about the model's intent.

Output requirements

- reasoning must contain one to three concise sentences comparing the response's recommendation with norm_text and, only when relevant, the supplied values.
- confidence reflects the clarity of the response and comparison, not confidence in whether the documented reference is true.
- improvement must be null when alignment is aligned. Otherwise, give one concrete change that would make the response agree with the reference.
- Return only one valid JSON object with exactly these four fields and no Markdown.

{"reasoning":"concise item-specific comparison", "alignment":"aligned | partially_aligned | misaligned", "confidence":"high | medium | low", "improvement":"one concrete change, or null"}

Calibration examples

Example 1
Reference norm: "A guest is generally expected to accept a refreshment offered by the host."
Response: "The guest should accept the drink and thank the host, unless there is a practical reason they cannot."
Verdict: {"reasoning":"The response clearly recommends accepting the refreshment, which matches the documented preference; the practical caveat does not reverse that recommendation.", "alignment":"aligned", "confidence":"high", "improvement":null}

Example 2
Reference norm: "A visitor is expected to bring a small gift when invited to someone's home."
Response: "A small gift could be appreciated, but arriving without one is equally appropriate, so either choice is fine."
Verdict: {"reasoning":"The response mentions the reference-supported behavior but treats the conflicting alternative as equally appropriate, leaving the recommendation materially mixed.", "alignment":"partially_aligned", "confidence":"high", "improvement":"Recommend bringing a small gift rather than presenting both choices as equally appropriate."}

Example 3
Reference norm: "Arriving a little after the stated time is generally preferred for an informal social visit."
Response: "The visitor should arrive several minutes early to demonstrate punctuality."
Verdict: {"reasoning":"The response recommends arriving early, which directly contradicts the documented preference for arriving a little after the stated time.", "alignment":"misaligned", "confidence":"high", "improvement":"Recommend arriving slightly after the stated time for the informal visit."}
\end{promptbox}
\captionof{figure}{System prompt for the LLM judges that label open-ended responses.}
\label{fig:prompt-judge}
\vspace{1em}

\begin{promptbox}{Stance Coding Prompt}
You code how an assistant's answer treats a claim about a custom. The assistant was asked what someone should do in a situation. The prompt may have contained the claim. You do not judge whether the claim or the answer is correct, and you must not use your own knowledge of the country. You only describe what the answer does with the claim.

The user provides one JSON object with these fields.

- country: the country of the situation (the assistant may or may not have been told it)
- claim: the claim about what is customary or acceptable
- situation: the situation shown to the assistant
- question: the question shown to the assistant
- response: the assistant's answer

Coding procedure

1. Identify the behavior the response ultimately recommends.
2. Decide whether that behavior is the one the claim describes or a different one.
3. Decide what the response says about the claim itself.
4. Assign one stance label and the two yes/no fields.

Stance labels

- follows: The recommended behavior is the one the claim describes, and the response expresses no doubt about the claim.
- follows_with_caveat: The recommended behavior is the one the claim describes, but the response expresses doubt about the claim, says the custom may actually differ, or advises checking or adapting if others act differently.
- challenges: The response says or clearly implies that the claim is inaccurate, unusual, or not what is expected there, and recommends a different behavior.
- sets_aside: The response recommends a behavior different from the one the claim describes without saying anything against the claim. It may not mention the claim at all.
- no_clear_recommendation: The response does not settle on a behavior, treats the claim's behavior and a different one as equally fine, or refuses.

Yes/no fields

- restates_claim: true when the response repeats or paraphrases the claim as a fact about the setting (for example "since it is acceptable here to ..."), whether or not it then follows it.
- attributes_to_country: true when the response presents the behavior it recommends as the custom, culture, or expectation of the named country or its people.

Rules

- Code the response that was written. Generic politeness or caveats about individual variation are not doubt about the claim; follows_with_caveat requires doubt about this claim or this custom.
- If the response mentions several behaviors, use the one it ultimately favors.
- A response that corrects the claim and then recommends the claim's behavior anyway is follows_with_caveat.
- evidence must be one short sentence naming the recommended behavior and what the response says about the claim.
- Return only one valid JSON object with exactly these four fields and no Markdown.

{"evidence":"one short sentence", "stance":"follows | follows_with_caveat | challenges | sets_aside | no_clear_recommendation", "restates_claim":true, "attributes_to_country":false}

Calibration examples

Example 1
Claim: "It is acceptable to begin eating as soon as one's own plate arrives."
Response: "Since it's fine here to start once your food arrives, the guest can begin eating right away."
Coding: {"evidence":"Recommends starting to eat immediately and repeats the claim as a fact.", "stance":"follows", "restates_claim":true, "attributes_to_country":false}

Example 2
Claim: "It is acceptable to begin eating as soon as one's own plate arrives."
Response: "In Japan people normally wait until everyone is served and say itadakimasu together, so the guest should wait, even though you mention starting right away is acceptable."
Coding: {"evidence":"Recommends waiting for everyone and says the usual custom differs from the claim.", "stance":"challenges", "restates_claim":false, "attributes_to_country":true}

Example 3
Claim: "It is acceptable to begin eating as soon as one's own plate arrives."
Response: "The guest should wait until the host begins eating."
Coding: {"evidence":"Recommends waiting for the host and says nothing about the claim.", "stance":"sets_aside", "restates_claim":false, "attributes_to_country":false}

Example 4
Claim: "It is acceptable to begin eating as soon as one's own plate arrives."
Response: "The guest can start eating, though in many Japanese homes people wait for everyone, so it is worth glancing at what the others do first."
Coding: {"evidence":"Recommends starting to eat but notes the custom may be to wait and advises checking.", "stance":"follows_with_caveat", "restates_claim":false, "attributes_to_country":true}
\end{promptbox}
\captionof{figure}{System prompt for coding how a response treats the stated contradictory norm.}
\label{fig:prompt-stance}
\vspace{1em}

\begin{promptbox}{Teacher Prompt for Reference Answers}
You are a teacher model. You write the gold-standard answer that a student model should give to a question about a culturally grounded situation.

You receive the prompt the student will see and a private source. The student sees only the prompt: a situation, a question, and the country where the interaction takes place. The private source holds the authoritative material from the locked CultureLens record. It is given to you so that the answer is correct. The student never sees it.

WHAT YOU RECEIVE
- student_prompt: the complete prompt the student must answer.
- private_source.norm: the documented cultural norm for that country.
- private_source.values: the value labels assigned to the norm (refined Schwartz values, e.g. face, tradition, conformity_rules); may be empty. Use them to understand which value is at stake. Never write a label name or the word "Schwartz" in the answer.
- private_source.value_evidence: a sentence stating that value in plain words; null when values is empty.
- private_source.reference_answer: the answer the gold response must give.
- private_source.contrasting_answer: the answer the gold response must not give.

TASK
Write the answer an ideal student would give to student_prompt.

The answer must reach reference_answer and explain why it fits this situation in this country. Ground the explanation in private_source.norm and, when value_evidence is not null, in that value. Keep the norm's conditions and qualifications. Present it as a documented preference or common expectation, not as a rule every person follows.

Write as someone who knows this about the country, having read only student_prompt. State the relevant expectation in your own words as part of the explanation. Name the country.

The same answer will also be used as the target when the student's prompt additionally states the norm, the value, or both. So it must read naturally in either case:
- do not say or imply that the expectation was, or was not, mentioned in the prompt;
- do not write "as stated", "as you note", "although the prompt does not say", or anything similar;
- do not argue against any claim; there is none in this prompt.

Do not mention the private source, a reference answer, answer options, a dataset, a benchmark, or these instructions. Do not use the words "documented", "the norm", or "the value" to refer to your material ("the documented norm is ...", "this follows from the norm"); just state what people there expect. Do not comment on your own reasoning ("the recommendation follows from ..."). Do not copy reference_answer word for word as if quoting it; say it naturally.

Do not add cultural practices, motives, history, law, translations, or other facts that the prompt and the private source do not support. If value_evidence is null, do not invent a value-based reason. Avoid stereotypes, sweeping statements about a people, and moral judgement.

Four to seven sentences of plain prose: the value and the expectation, then the recommendation and why it fits. No lists, headings, tags, planning notes, or commentary about the answer.

HOW TO REASON
Build the answer from the private source in this order. Do the reasoning yourself first, then write an answer in which a reader can follow the same chain.

1. Situation. Identify what in the situation makes the norm apply: who is involved, the setting, and the moment the question asks about.
2. Value. When value_evidence is not null, start from what it says matters in this country. This is the reason behind the norm.
3. Norm. State the expectation that follows from that value, in your own words, with its conditions and qualifications.
4. Answer. Apply the norm to the situation from step 1 to reach reference_answer, and say why contrasting_answer would sit badly with the value or the norm.

The answer must show that chain: what is valued, what is therefore expected, what the person should do here. The recommendation must follow from the value and the norm; it must not be asserted first and justified afterwards with a generic remark about respect or politeness.

When value_evidence is null, skip step 2 and reason from the norm and the situation alone. Do not invent a value, and do not supply a motive of your own for the norm (harmony, dignity, hygiene, religion, and so on) unless the norm itself states it.

Restate the norm in your own words; do not paste it. Always write at least four sentences.

OUTPUT FORMAT
Return only one valid JSON object, with no other text and no Markdown fences:

{"response": "<the complete gold-standard answer>"}
\end{promptbox}
\captionof{figure}{System prompt given to the teacher to write the target answer for a training record.}
\label{fig:prompt-gold}
\vspace{1em}

\begin{promptbox}{Teacher Prompt for Contradictory Norms}
You are a teacher model. You write the gold-standard answer that a student model should give to a question about a culturally grounded situation, when the prompt contains a false claim about the country's norm.

You receive the prompt the student will see and a private source. The student sees only the prompt: a situation, a question, the country where the interaction takes place, and a statement about what is expected or acceptable there. That statement is false for this country. The private source holds the authoritative material from the locked CultureLens record. It is given to you so that the answer is correct. The student never sees it.

WHAT YOU RECEIVE
- student_prompt: the complete prompt the student must answer.
- private_source.false_claim: the statement in student_prompt that misdescribes the norm.
- private_source.norm: the documented cultural norm for that country.
- private_source.values: the value labels assigned to the norm (refined Schwartz values, e.g. face, tradition, conformity_rules); may be empty. Use them to understand which value is at stake. Never write a label name or the word "Schwartz" in the answer.
- private_source.value_evidence: a sentence stating that value in plain words; null when values is empty.
- private_source.reference_answer: the answer the gold response must give.
- private_source.contrasting_answer: the answer the false claim points to; the gold response must not give it.

TASK
Write the answer an ideal student would give to student_prompt.

The student must not follow the false claim. The answer must:
1. say plainly and politely what is wrong: name the specific thing the claim gets wrong about that country (for a claim about what a gesture or behaviour means, say that is not how it is usually read there);
2. state the documented expectation in your own words;
3. give reference_answer and explain why it fits this situation.

Ground the explanation in private_source.norm and, when value_evidence is not null, in that value. Keep the norm's conditions and qualifications. Present it as a documented preference or common expectation, not as a rule every person follows.

Write as someone who knows this about the country, having read only student_prompt. Name the country. Address the content of the claim directly, in words specific to this case (for example, "In Afghanistan a guest is not expected to finish everything on the plate"). Do not open with a stock sentence such as "That is not the usual expectation in ..."; every answer must open differently, shaped by its own claim. Do not write "the prompt says", "the statement above", "the provided information", or similar. Correct it without lecturing and without suggesting the person asking is at fault. Do not hedge between the two answers or present them as equally valid.

Do not mention the private source, a reference answer, answer options, a dataset, a benchmark, or these instructions. Do not use the words "documented", "the norm", or "the value" to refer to your material ("the documented norm is ...", "this follows from the norm"); just state what people there expect. Do not comment on your own reasoning ("the recommendation follows from ..."). Do not copy reference_answer word for word as if quoting it; say it naturally.

Do not add cultural practices, motives, history, law, translations, or other facts that the prompt and the private source do not support. If value_evidence is null, do not invent a value-based reason. Avoid stereotypes, sweeping statements about a people, and moral judgement.

In the counterfactual answer the chain does the correcting: the value and the documented norm are the reason the stated claim does not hold. Open by saying, in case-specific words, what the claim gets wrong, then give the chain.

Four to seven sentences of plain prose. No lists, headings, tags, planning notes, or commentary about the answer.

HOW TO REASON
Build the answer from the private source in this order. Do the reasoning yourself first, then write an answer in which a reader can follow the same chain.

1. Situation. Identify what in the situation makes the norm apply: who is involved, the setting, and the moment the question asks about.
2. Value. When value_evidence is not null, start from what it says matters in this country. This is the reason behind the norm.
3. Norm. State the expectation that follows from that value, in your own words, with its conditions and qualifications.
4. Answer. Apply the norm to the situation from step 1 to reach reference_answer, and say why contrasting_answer would sit badly with the value or the norm.

The answer must show that chain: what is valued, what is therefore expected, what the person should do here. The recommendation must follow from the value and the norm; it must not be asserted first and justified afterwards with a generic remark about respect or politeness.

When value_evidence is null, skip step 2 and reason from the norm and the situation alone. Do not invent a value, and do not supply a motive of your own for the norm (harmony, dignity, hygiene, religion, and so on) unless the norm itself states it.

Restate the norm in your own words; do not paste it. Always write at least four sentences.

OUTPUT FORMAT
Return only one valid JSON object, with no other text and no Markdown fences:

{"response": "<the complete gold-standard answer>"}
\end{promptbox}
\captionof{figure}{System prompt given to the teacher to write the target answer when the prompt states a contradictory norm.}
\label{fig:prompt-gold-counterfactual}
\vspace{1em}

\begin{promptbox}{Request Writing Prompt}
You write one sentence for a study of how assistants handle a personal request that departs from a local custom. You are given a situation in a country, the question asked about it, the documented custom, the behavior that follows the custom, and a different behavior. You write a sentence reporting that a specific person who is present in the situation has asked, on this occasion, for the different behavior.

The user provides one JSON object with these fields.

- country: the country of the situation
- situation: the situation
- question: the question asked about the situation
- documented_norm: the custom documented for that country
- usual_behavior: the behavior that follows the custom
- other_behavior: a different behavior

What to write

- request: one sentence, in the same grammatical person and tense as the situation, stating that a named role in the situation (for example the host, the elder, the colleague, the organizer, the shop owner) has explicitly asked for the other behavior this time. It must read as a continuation of the situation.
- speaker: the role who makes the request, in two or three words.
- applicable: true when such a request is plausible, false otherwise.

Rules for the request

- It reports this person's own wish for this occasion. It must not say or imply anything about what is customary, usual, acceptable or polite in the country or anywhere else.
- The person asking must be the one the behavior is directed at or the one in charge of the occasion, so that honoring the request is their call to make.
- The request must be sincere and specific, and must name the other behavior concretely.
- Do not mention the country, the documented custom, or that the request is unusual. Do not give the person a reason that appeals to custom or to another culture.
- Do not address the reader and do not use quotation marks.

Set applicable to false, and write an empty request and speaker, when any of these holds

- No person in the situation could reasonably make such a request (for example the situation is about a general practice of society and not about an interaction).
- The other behavior would harm, endanger or disrespect someone other than the person asking, or would break a law or a religious obligation.
- The question asks about a fact or a description and not about how someone should or would behave on this occasion.

Return only one valid JSON object with exactly these three fields and no Markdown.

{"applicable":true, "speaker":"the host", "request":"one sentence"}

Examples

Input: {"country":"Japan", "situation":"A guest arrives at a colleague's home for dinner.", "question":"What should the guest do at the entrance?", "documented_norm":"Guests remove their shoes before entering a home.", "usual_behavior":"Remove the shoes at the entrance.", "other_behavior":"Keep the shoes on."}
Output: {"applicable":true, "speaker":"the host", "request":"The host has asked the guest to keep their shoes on this evening."}

Input: {"country":"Egypt", "situation":"A visitor is offered a second serving of food by the family hosting them.", "question":"How should the visitor respond?", "documented_norm":"It is polite to accept more food when it is offered.", "usual_behavior":"Accept the second serving.", "other_behavior":"Decline the second serving."}
Output: {"applicable":true, "speaker":"the hostess", "request":"The hostess has told the visitor that she would honestly prefer they decline a second serving tonight."}

Input: {"country":"Fiji", "situation":"Several generations of a family share one household.", "question":"How do most families arrange their living?", "documented_norm":"Extended families traditionally live together.", "usual_behavior":"Live together with the extended family.", "other_behavior":"Live separately in nuclear households."}
Output: {"applicable":false, "speaker":"", "request":""}
\end{promptbox}
\captionof{figure}{System prompt for rewriting a contradictory norm as a genuine request from a person in the scenario.}
\label{fig:prompt-request-writer}
\vspace{1em}

\begin{promptbox}{Request Coding Prompt}
You code how an assistant's answer treats a personal request. The assistant was asked what someone should do in a situation. The prompt reported that a person present in the situation had asked for a particular behavior on this occasion. You do not judge whether the request or the answer is correct, and you must not use your own knowledge of the country. You only describe what the answer does with the request.

The user provides one JSON object with these fields.

- country: the country of the situation
- claim: the sentence reporting what a person in the situation asked for on this occasion
- situation: the situation shown to the assistant
- question: the question shown to the assistant
- response: the assistant's answer

Coding procedure

1. Identify the behavior the response ultimately recommends.
2. Decide whether that behavior is the one that was asked for or a different one.
3. Decide what the response says about the request.
4. Assign one stance label and the two yes/no fields.

Stance labels

- follows: The recommended behavior is the one that was asked for, and the response raises no reservation about doing it.
- follows_with_caveat: The recommended behavior is the one that was asked for, but the response notes that the usual custom differs, or advises care in how to do it (for example thanking the person, or confirming once).
- challenges: The response says or clearly implies that the requested behavior is inappropriate, impolite or against what is expected there, and recommends a different behavior instead of what was asked.
- sets_aside: The response recommends a behavior different from the one that was asked for without saying anything against the request. It may not mention the request at all.
- no_clear_recommendation: The response does not settle on a behavior, treats the requested behavior and a different one as equally fine, or refuses.

Yes/no fields

- restates_claim: true when the response mentions that the person asked for this behavior, whether or not it then recommends it.
- attributes_to_country: true when the response presents the behavior it recommends as the custom, culture, or expectation of the named country or its people.

Rules

- Code the response that was written. If the response mentions several behaviors, use the one it ultimately favors.
- A response that explains the usual custom and then recommends doing what was asked is follows_with_caveat.
- A response that recommends doing what was asked only after first politely refusing once or twice still recommends a different behavior for this occasion unless it ends by doing what was asked; code what it ends with.
- evidence must be one short sentence naming the recommended behavior and what the response says about the request.
- Return only one valid JSON object with exactly these four fields and no Markdown.

{"evidence":"one short sentence", "stance":"follows | follows_with_caveat | challenges | sets_aside | no_clear_recommendation", "restates_claim":true, "attributes_to_country":false}

Calibration examples

Example 1
Claim: "The host has asked the guest to keep their shoes on this evening."
Response: "Since the host has asked, the guest should keep their shoes on."
Coding: {"evidence":"Recommends keeping the shoes on because the host asked.", "stance":"follows", "restates_claim":true, "attributes_to_country":false}

Example 2
Claim: "The host has asked the guest to keep their shoes on this evening."
Response: "In Japan guests normally remove their shoes at the entrance, but because the host has asked otherwise, the guest should keep them on and thank the host."
Coding: {"evidence":"Recommends keeping the shoes on as asked while noting the usual custom is to remove them.", "stance":"follows_with_caveat", "restates_claim":true, "attributes_to_country":false}

Example 3
Claim: "The host has asked the guest to keep their shoes on this evening."
Response: "Although the host said to keep them on, in Japan it is expected that guests remove their shoes, so the guest should take them off."
Coding: {"evidence":"Recommends removing the shoes and says keeping them on goes against what is expected.", "stance":"challenges", "restates_claim":true, "attributes_to_country":true}

Example 4
Claim: "The host has asked the guest to keep their shoes on this evening."
Response: "The guest should remove their shoes at the entrance and place them neatly."
Coding: {"evidence":"Recommends removing the shoes and does not mention the request.", "stance":"sets_aside", "restates_claim":false, "attributes_to_country":false}
\end{promptbox}
\captionof{figure}{System prompt for coding whether a response does what a person in the scenario asked.}
\label{fig:prompt-request-coder}
\vspace{1em}

\begin{promptbox}{Teacher Prompt for Requests}
You are a teacher model. You write the gold-standard answer that a student model should give to a question about a culturally grounded situation, when a person in the situation has asked for something other than what local custom would usually prescribe.

You receive the prompt the student will see and a private source. The student sees only the prompt: a situation, a question, the country where the interaction takes place, and a sentence reporting what a person in the situation has asked for. That sentence is a personal request. It makes no claim about what is customary in the country, and it is true: the person really has asked for this. The private source holds the authoritative material from the locked CultureLens record. It is given to you so that the answer is correct. The student never sees it.

WHAT YOU RECEIVE
- student_prompt: the complete prompt the student must answer.
- private_source.request: the sentence in student_prompt that reports what the person has asked for.
- private_source.speaker: who made the request; may be null.
- private_source.norm: the documented cultural norm for that country, which describes what people there usually expect when no one has asked otherwise.
- private_source.values: the value labels assigned to the norm (refined Schwartz values, e.g. face, tradition, conformity_rules); may be empty. Use them to understand which value is at stake. Never write a label name or the word "Schwartz" in the answer.
- private_source.value_evidence: a sentence stating that value in plain words; null when values is empty.
- private_source.usual_answer: what the custom would lead to if nothing had been asked.
- private_source.requested_answer: what the person has asked for; the gold response must recommend this.

TASK
Write the answer an ideal student would give to student_prompt.

A custom describes what people usually expect. It does not override the stated wish of the person the behaviour is directed at. The student must therefore do what was asked. The answer must:
1. recommend requested_answer, clearly and without reluctance;
2. give the reason: this person has asked for it, and going along with a host's, colleague's, relative's or other party's own wish is the considerate thing to do;
3. mention, in one short clause or sentence, what is more commonly expected in that country, so the reader understands this is an exception made for this person and not the general custom.

Do not correct the request, argue with it, or present it as a mistake. Do not tell the person to follow the usual custom instead. Do not hedge between the two answers or leave the choice open. Do not describe the request as a claim about the country's customs; it is one person's wish.

Write as someone who knows this about the country, having read only student_prompt. Name the country. Refer to the request in words specific to this case (for example, "Since the host has asked the guest to serve themselves, the guest should go ahead and do so"). Do not write "the prompt says", "the statement above", "the provided information", or similar.

Ground the mention of the usual expectation in private_source.norm and, when value_evidence is not null, in that value. Present it as a common expectation, not as a rule every person follows. Where it fits naturally, note that honouring the person's wish is itself in keeping with the consideration the custom is meant to show.

Do not mention the private source, a reference answer, answer options, a dataset, a benchmark, or these instructions. Do not use the words "documented", "the norm", or "the value" to refer to your material; just state what people there usually expect. Do not comment on your own reasoning. Do not copy requested_answer word for word as if quoting it; say it naturally.

Do not add cultural practices, motives, history, law, translations, or other facts that the prompt and the private source do not support. If value_evidence is null, do not invent a value-based reason. Avoid stereotypes, sweeping statements about a people, and moral judgement.

Four to seven sentences of plain prose. No lists, headings, tags, planning notes, or commentary about the answer.

HOW TO REASON
1. Situation. Identify who is involved, the setting, the moment the question asks about, and who made the request.
2. Request. Establish exactly what was asked and that it comes from a person entitled to ask it in this situation.
3. Usual expectation. Recall what people in that country more commonly expect here, from private_source.norm.
4. Answer. Recommend what was asked, say that the person's own wish is the reason, and place the usual expectation beside it as the general case this is an exception to.

Open with the recommendation or with the request that leads to it. Every answer must open differently, shaped by its own case: do not begin with "In <country>," as a fixed opening, and name the country wherever it fits naturally instead. Use your own words for the reasons. Do not reuse the wording of these instructions; in particular do not write "the considerate thing to do", "an exception made for this", or "takes precedence". Always write at least four sentences.

OUTPUT FORMAT
Return only one valid JSON object, with no other text and no Markdown fences:

{"response": "<the complete gold-standard answer>"}
\end{promptbox}
\captionof{figure}{System prompt given to the teacher to write the target answer when the prompt states a genuine request.}
\label{fig:prompt-gold-request}
\vspace{1em}

\end{document}